\def\confName{CVPR}
\def\confYear{2027}

\def\paperTitle{LoopDVR: Looped Transformer for Efficient Dense Visual Recognition}

\def\authorBlock{
    Sirui Liang$^{1,2}$ \qquad Xuanlong Yu$^{2}$ \qquad Longfei Liu$^{2}$ \qquad
    Xiaodong Cun$^{1}$\thanks{Corresponding author.} \qquad Xi Shen$^{2}$ \\\\
    $^{1}$~GVC Lab, Great Bay University \qquad
    $^{2}$~Intellindust AI Lab \\[1em]
    \normalsize Project Page: \url{https://siruuui.github.io/LoopDVR/} \\
    \normalsize  Code: \url{https://github.com/Intellindust-AI-Lab/LoopDVR}
}

\newif\ifreview 
\newif\ifarxiv \newcommand{\arxiv}{\arxivtrue}
\newif\ifcamera 
\newif\ifrebuttal

\makeatletter
\DeclareRobustCommand\onedot{\futurelet\@let@token\@onedot}
\def\@onedot{\ifx\@let@token.\else.\null\fi\xspace}

\def\eg{\emph{e.g}\onedot} 
\def\ie{\emph{i.e}\onedot}

\makeatother

\makeatletter
\DeclareRobustCommand\onedot{\futurelet\@let@token\@onedot}
\def\@onedot{\ifx\@let@token.\else.\null\fi\xspace}

\def\eg{\emph{e.g}\onedot} 
\def\ie{\emph{i.e}\onedot}

\makeatother

\arxiv %

\pdfoutput=1
\documentclass[10pt,twocolumn,letterpaper]{article}
\PassOptionsToPackage{table,xcdraw}{xcolor}

\ifreview \usepackage[review]{cvpr} \fi
\ifarxiv \usepackage[pagenumbers]{cvpr} \fi
\ifrebuttal \usepackage[rebuttal]{cvpr} \fi
\ifcamera \usepackage{cvpr} \fi

\usepackage{microtype}
\usepackage{epsfig}
\usepackage{float}
\usepackage{placeins}
\usepackage{stfloats}
\usepackage{tabularx}
\usepackage{xstring}
\usepackage{multirow}
\usepackage[hang,flushmargin]{footmisc}
\usepackage{makecell}
\usepackage{listings}
\usepackage{algorithm}
\usepackage{xcolor}
\usepackage{algorithm}
\usepackage{listings}
\usepackage{xcolor}

\lstdefinestyle{pseudocode}{
    language=Python,
    basicstyle=\ttfamily\footnotesize,
    keywordstyle=\bfseries,
    commentstyle=\color{teal!75!black},
    emph={zip, norm, mlp, append},
    emphstyle=\color{magenta},
    columns=fullflexible,
    keepspaces=true,
    showstringspaces=false,
    frame=none,
    numbers=none,
    aboveskip=0pt,
    belowskip=0pt
}

\ifcamera \usepackage[accsupp]{axessibility} \fi

\newsavebox{\autotabularbox}
\NewDocumentEnvironment{autotabular}{m m +b}{%
  \sbox{\autotabularbox}{\begin{tabular}{#2}#3\end{tabular}}%
  \setlength{\tabcolsep}{\dimexpr\tabcolsep+(\linewidth-\wd\autotabularbox)/#1/2\relax}%
  \begin{tabular}{#2}#3\end{tabular}%
}{}

\newcommand{\parambase}[1]{%
    \makebox[2.4em][r]{#1}%
    \makebox[3.35em][l]{}%
}
\newcommand{\paramdrop}[2]{%
    \makebox[2.4em][r]{#1}%
    \makebox[3.35em][l]{%
        \hspace{0.18em}%
        \raisebox{-0.45ex}{%
            \textcolor{myred}{\scriptsize$\downarrow#2\%$}%
        }%
    }%
}

\ifarxiv  \fi

\newcommand{\R}[1]{{%
    \textbf{%
        \ifstrequal{#1}{1}{\textcolor{red}{R#1}}{%
        \ifstrequal{#1}{2}{\textcolor{blue}{R#1}}{%
        \ifstrequal{#1}{3}{\textcolor{magenta}{R#1}}{%
        \ifstrequal{#1}{4}{\textcolor{teal}{R#1}}{%
                           \textcolor{cyan}{R#1}%
        }}}}%
    }%
}}

\usepackage[draft]{changes}
\definechangesauthor[name={Xiaodong}, color=orange]{xd}

\usepackage{xr-hyper}

\makeatletter
\newcommand*{\addFileDependency}[1]{
  \typeout{(#1)}
  \@addtofilelist{#1}
  \IfFileExists{#1}{}{\typeout{No file #1.}}
}
\makeatother

\ifarxiv
\definecolor{cvprblue}{rgb}{0.21,0.49,0.74}
\definecolor{alizarin}{rgb}{0.82,0.1,0.26}
\usepackage[
    pagebackref,
    breaklinks,
    colorlinks,
    citecolor=blue,
    urlcolor=alizarin
]{hyperref}
\else
\definecolor{cvprblue}{rgb}{0.21,0.49,0.74}
\usepackage[
    pagebackref,
    breaklinks,
    colorlinks,
    allcolors=cvprblue
]{hyperref}
\fi

\definecolor{mygreen}{RGB}{0, 170, 110}
\definecolor{myred}{RGB}{210, 65, 70}
\usepackage{cuted}

\begin{document}
\title{\paperTitle}
\author{\authorBlock}
\maketitle
\begin{strip}
    \centering
    \begin{minipage}[t]{0.33\linewidth}
        \centering
        \includegraphics[width=\linewidth]{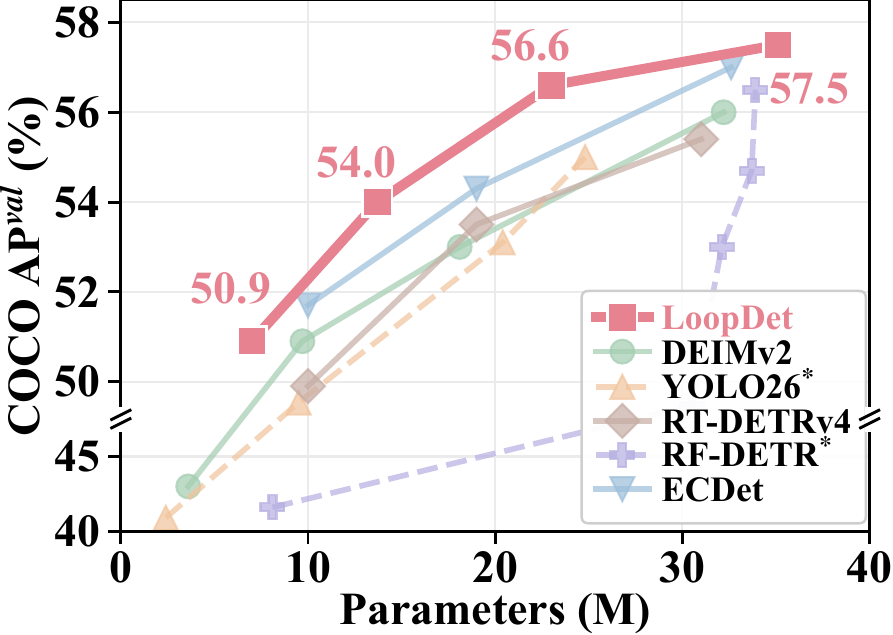}
        {\small\textbf{(a) Object Detection}\par}
    \end{minipage}%
    \hfill
    \begin{minipage}[t]{0.33\linewidth}
        \centering
        \includegraphics[width=\linewidth]{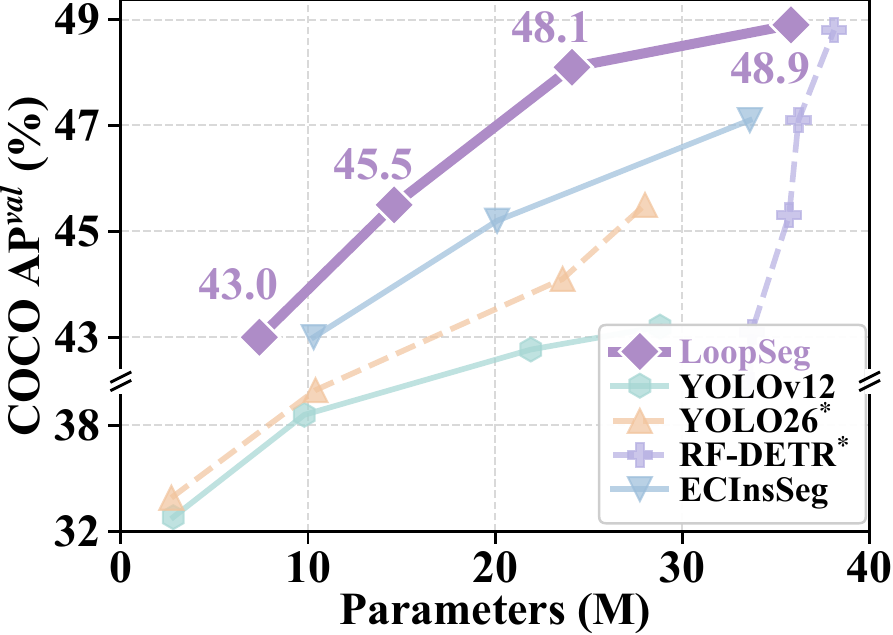}
        {\small\textbf{(b) Instance Segmentation}\par}
    \end{minipage}
    \hfill
    \begin{minipage}[t]{0.33\linewidth}
        \centering
        \includegraphics[width=\linewidth]{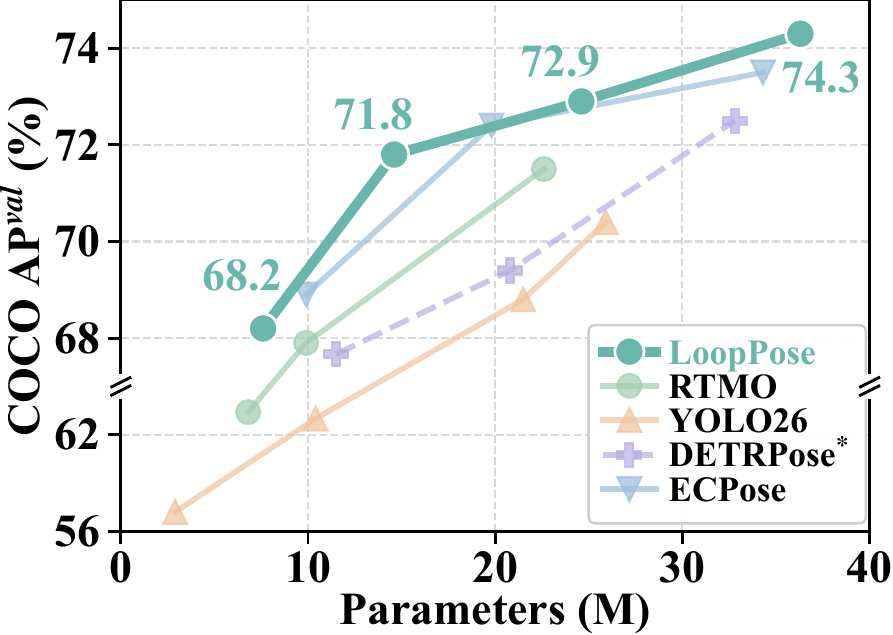}
        {\small\textbf{(c) Human Pose Estimation}\par}
    \end{minipage}
    \captionsetup{hypcap=false}
    \captionof{figure}{
        Parameter--accuracy trade-offs compared on the COCO~\cite{lin2014microsoft} dataset~(* indicates additional pretraining on extra data).
    }
    \captionsetup{hypcap=true}
    \label{fig:main_comparison}
\end{strip}

\begin{abstract}
We propose LoopDVR, a framework for converting pretrained Transformer-based dense visual recognition models into parameter-efficient looped networks. 
Rather than reducing the network width or depth directly, LoopDVR reuses Transformer blocks across layers in both the ViT backbone and Transformer decoder, reducing the number of parameters while retaining the original computational structure.
However, naively sharing parameters can disrupt the distinct transformations learned at different depths.
To address this challenge, we identify suitable sharing groups based on inter-layer representation dissimilarity and keep the final backbone block independent for multi-scale feature construction.
We then initialize each shared backbone block with the pretrained weights of its group's terminal layer to better preserve the learned representations.
An independent Loop Adapter at each layer further helps retain depth-specific transformations within shared blocks.
Subsequently, we distill visual representations into the looped backbone and integrate it with a looped decoder for task-specific training.
Experiments demonstrate that LoopDVR achieves the best accuracy--parameter trade-offs across dense recognition tasks. 
Compared with the reference models, it reduces parameters by approximately 28\%  while maintaining competitive accuracy.
Notably, LoopDet-S achieves 50.9 box AP with only 7.0M parameters, highlighting its effectiveness in compact models.
\end{abstract}

\section{Introduction}
\label{sec:intro}

Dense Visual Recognition (DVR), including object detection~\cite{girshick2015fastrcnn}, instance segmentation~\cite{he2017maskrcnn}, and human pose estimation~\cite{Openpose}, supports broad applications, \eg, autonomous driving~\cite{Eduardo2019driving}, robotics~\cite{labbe2020CosyPose} and augmented reality~\cite{chan2019everybody}.
Transformer-based architectures achieve strong performance via attention-based visual representations and iterative query refinement~\cite{carion2020detr,zhu2021deformable}.
But deploying these models on edge devices with limited model-size and GPU memory budgets remains challenging.
This motivates extensive research regarding parameter efficiency.

Existing approaches improve model parameter efficiency through compact architecture designs~\cite{chen2024lw,zhao2024detrs,robinson2025rf,lv2024rtdetrv2} or knowledge distillation to smaller students~\cite{huang2025deimv2,chang2023detrdistill,chen2024d3etr}.
Using shallower or narrower networks is a straightforward way to reduce parameter counts.
However, it also limits computational depth or representational capacity, and potentially compromises performance, particularly for compact models.
This raises a question: Can we reduce parameter counts while preserving the original computational structure, enabling parameter-efficient models to achieve performance comparable to full-sized models?
Recent studies on looped Vision Transformers, \eg, MiniViT~\cite{zhang2022minivit} and Raptor~\cite{jacobs2026block}, offer a promising solution. 
These methods reuse the same Transformer weights across multiple layers within each group. 
This strategy reduces parameter counts while retaining the original computational structure and maintaining competitive performance.
However, these classification-oriented methods perform less effectively on dense recognition tasks.
Similar ideas have also been explored in object detection: DEQDet~\cite{wang2023deq} and RecursiveDet~\cite{zhao2023recursivedet} reuse decoder layers for iterative query refinement. 
But both leave the backbone untouched, overlooking the substantial parameter overhead associated with this component.
Moreover, directly looping the same weights across layers can disrupt the hierarchical representations and depth-specific behaviors, leading to training instability and performance degradation~\cite{zhang2022minivit}.

To overcome these limitations, we propose \textbf{LoopDVR}, a pretrained-model-guided framework for converting dense recognition models into parameter-efficient looped networks.
It applies looped Transformers within both the ViT backbone and the query decoder.
To enable effective parameter sharing while preserving depth-specific behaviors, we investigate two key questions: 
\textbf{1) Where to Loop?} 
This concerns identifying layers suitable for parameter sharing while minimizing disruption to the hierarchical representations. 
We adopt a representation-based grouping strategy, which partitions consecutive layers with similar representations into the same sharing group. 
Furthermore, to better support multi-scale feature construction from the penultimate and final layers, we reserve the terminal block as an independent group. 
\textbf{2) How to Loop?} This focuses on enabling the shared network to preserve the depth-specific transformation of different layers.
To this end, we use structured initialization, initializing each shared backbone block with the pretrained weights of its group's terminal layer, which preserves the learned representations of the pretrained model.
We further introduce an independent lightweight Loop Adapter for each layer, which performs depth-specific transformation after the shared block. 
Subsequently, we distill visual representations into the looped backbone before integrating it with the looped decoder for downstream training.

We instantiate LoopDVR as LoopDet, LoopSeg, and LoopPose for object detection, instance segmentation, and human pose estimation, respectively. 
Across four model scales, LoopDVR reduces parameters by approximately 28\% while maintaining competitive accuracy. 
As shown in Fig.~\ref{fig:main_comparison}, our models achieve the best accuracy–parameter trade-offs.
Notably, LoopDet-S achieves 50.9 box AP with only 7.0M parameters, while LoopSeg-L improves mask AP by 1.0 over its non-looped reference model.

In summary, our main contributions are as follows:
\begin{itemize}
    \item We introduce \textit{LoopDVR}, a pretrained-model-guided conversion framework that combines representation-based grouping, group-end initialization, and feature distillation to enable effective parameter sharing.
    \item We design a unified looped architecture  by reusing Transformer blocks in both the ViT backbone and query decoder with Loop Adapters, achieving exceptional parameter efficiency while preserving computational depth.
    \item We demonstrate the effectiveness of LoopDVR on object detection, instance segmentation, and human pose estimation across four model scales, achieving the most favorable accuracy–parameter trade-offs.
\end{itemize}

\section{Related Work}
\label{sec:related}

\noindent\textbf{Transformer-based Dense Recognition.}
DETR~\cite{carion2020detr} established query-based set prediction for object detection, laying the foundation for a Transformer-based paradigm in dense visual recognition~\cite{liu2022dabdetr,zhang2023dino,cheng2021maskformer,mask2former,PETR}.
ViDT~\cite{song2022vidt} extended the role of Transformers to feature extraction by pairing a vision Transformer backbone with a deformable query decoder~\cite{zhu2021deformable}.
This paradigm extends to other dense visual recognition tasks, including instance segmentation with MaskDINO~\cite{maskdino} and human pose estimation with ED-Pose~\cite{ED-Pose} and GroupPose~\cite{GroupPose}.
Subsequent work tailored this architecture to real-time detection.
LW-DETR~\cite{chen2024lw} pairs a pretrained ViT backbone with a shallow decoder, while RF-DETR~\cite{robinson2025rf} builds a real-time detector based on a DINOv2~\cite{oquab2023dinov2} backbone.
DEIMv2~\cite{huang2025deimv2} distills DINOv3~\cite{simeoni2025dinov3} features into narrower ViT backbones for real-time detection.
EdgeCrafter~\cite{liu2026edgecrafter} brings these directions together in a compact framework for edge dense recognition. 
It distills features from a detection-specialized teacher into ECViT and reuses the resulting backbone for object detection, instance segmentation, and pose estimation.
These advances have made Transformer-based models effective across dense recognition tasks. 
Yet feature extraction and query refinement still rely on independently parameterized blocks, whose parameter cost remains a concern under tight memory budgets.

\vspace{0.5em}
\noindent\textbf{Looped Transformers in Computer Vision. }
Looped Transformers introduced a recurrent paradigm which reduces parameter cost by reusing blocks across depth~\cite{dehghani2018universal,Lan2020ALBERT:}, a principle later adopted in computer vision.
In Vision Transformers, SReT~\cite{shen2022sliced} recursively applies shared Transformer blocks, while MiniViT~\cite{zhang2022minivit} uses depth-specific transformations to preserve variation across layers that share weights.
Raptor~\cite{jacobs2026block} introduces the \emph{Block-Recurrent Hypothesis} and groups consecutive layers by the similarity of their representations in a pretrained ViT, approximating each group with a recurrent block.
Recent studies have extended looped Transformers to image super-resolution~\cite{zhou2024ristra}, multi-view 3D reconstruction~\cite{burzio2026dejaview}, visual tracking~\cite{peng2026looptrack}, image and video generation~\cite{chng2026looped,goyal2026elt}, and vision-language models~\cite{adaloop2026,qian2026loopvl}.
In object detection, RecursiveDet~\cite{zhao2023recursivedet} shares decoder parameters across box refinement stages, while DEQDet~\cite{wang2023deq,Wang2025deqds} formulates query refinement as a fixed-point iteration.
Together, these studies show that Transformers can build computational depth by repeatedly applying shared blocks. 
How to exploit this reuse across depth in both the ViT backbone and the query decoder of a compact dense recognition model remains underexplored.

\begin{figure*}[tp]
    \centering
    \includegraphics[width=\textwidth]{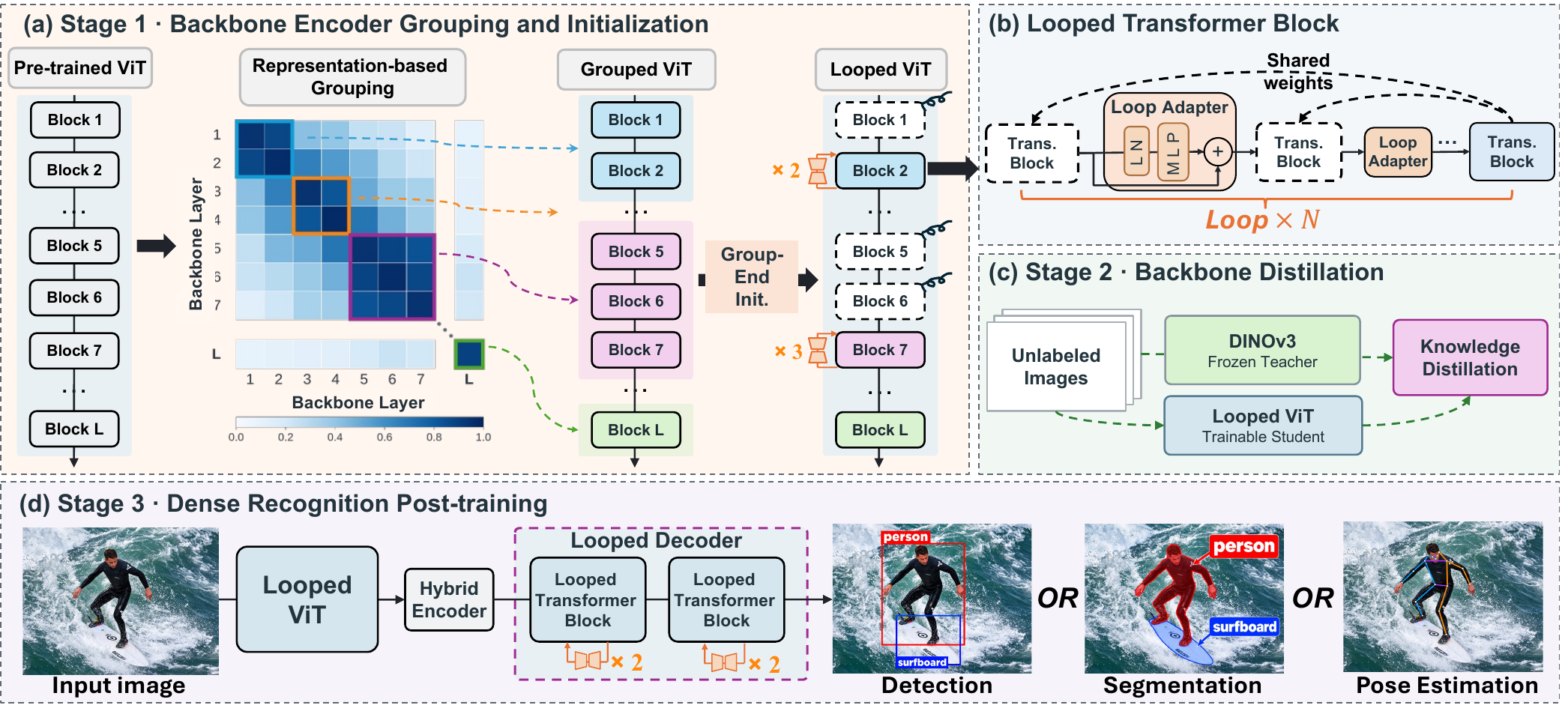}
    \vspace{-1.5em}
    \caption{\textbf{Overview of LoopDVR.}
    Given a dense recognition model, (a) Stage 1 groups the layers of the pre-trained ViT backbone based on inter-layer representation similarity and initializes each looped backbone block using the weights of its group’s terminal layer. 
    (b) Transformer block parameters are shared within each group, while each layer has a Loop Adapter to preserve depth-specific transformations.
    (c) Stage 2 distills the looped ViT using a DINOv3 teacher.
    (d) Stage 3 post-trains the resulting architecture for dense recognition, combining the looped ViT, the hybrid encoder and the looped decoder.
    } 
    \label{fig:main}
\end{figure*}

\section{Method}
\label{sec:method}

We present LoopDVR, a general framework that transforms pre-trained dense visual recognition models into more parameter-efficient looped networks. 
As illustrated in Fig.~\ref{fig:main}, LoopDVR reduces parameter cost by reusing Transformer blocks across multiple layers in both the backbone and decoder, while preserving the original computational depth.
We first review the common architecture of Transformer-based dense visual recognition models in Sec.~\ref{sec:preliminaries}.
We then describe the construction of the looped backbone in Sec.~\ref{sec:loop_construction}, addressing where to loop through representation-based grouping and how to loop through structured initialization and Loop Adapters.
Finally, we describe the adaptation to dense recognition tasks in Sec.~\ref{sec:training}, including backbone distillation, looped decoder construction, and task-specific training.

\subsection{Preliminaries: Transformer-based DVR }
\label{sec:preliminaries}

A Transformer-based dense visual recognition architecture consists of a ViT backbone, a hybrid encoder for multi-scale feature fusion, and a query-based Transformer decoder with task-specific prediction heads~\cite{huang2025deimv2,liu2026edgecrafter}.

The ViT backbone extracts visual representations by sequentially processing image patch tokens through $L$ Transformer blocks~\cite{dosovitskiy2020vit}.
Starting from the input patch embeddings $\mathbf X^{(0)}$, the $l$-th block $\mathcal B_l$ transforms the preceding token representations $\mathbf X^{(l-1)}$ into $\mathbf X^{(l)}$ as
\begin{equation}
    \mathbf X^{(l)}
    = \mathcal B_l(\mathbf X^{(l-1)}),
    \quad l=1,\ldots,L.
    \label{eq:standard_vit}
\end{equation}

Dense visual recognition benefits from multi-scale feature representations~\cite{zhu2021deformable}.
To this end, a hybrid encoder transforms the backbone representations into multi-scale features for the decoder~\cite{zhao2024detrs}.
The encoder uses attention to process features within each scale and convolutions to fuse features across scales.
Given the final two backbone representations, the encoded features are computed as~\cite{liu2026edgecrafter}
\begin{equation}
    \mathbf F_{\mathrm{enc}}
    = \mathcal E(\mathbf X^{(L-1)},\mathbf X^{(L)}),
    \label{eq:feature_encoder}
\end{equation}
where $\mathcal E$ denotes the hybrid encoder.
The resulting features $\mathbf F_{\mathrm{enc}}$ provide multi-scale visual context for the decoder.

Given the encoded features $\mathbf F_{\mathrm{enc}}$, the Transformer decoder iteratively refines object queries and their reference boxes through $T$ Transformer blocks~\cite{liu2022dabdetr,zhang2023dino}.
Each decoder block contains self-attention among object queries, deformable cross-attention to encoded features guided by reference boxes, and a feed-forward network~\cite{zhu2021deformable}.
At layer $t$, the query representations $\mathbf Q^{(t-1)}$ are updated by the decoder block $\mathcal D_t$ using reference boxes $\mathbf R^{(t-1)}$ and features $\mathbf F_{\mathrm{enc}}$:
\begin{equation}
    \mathbf Q^{(t)}
    = \mathcal D_t\left(
        \mathbf Q^{(t-1)},
        \mathbf R^{(t-1)},
        \mathbf F_{\mathrm{enc}}
    \right),
    \mathbf Q^{(t)} \in \mathbb R^{N_q \times C},
    \label{eq:standard_decoder}
\end{equation}
where $N_q$ denotes the number of object queries, and $C$ is the query feature dimension.
The updated queries are used to further refine reference boxes and generate task-specific predictions.

We instantiate LoopDVR based on a representative family of dense visual recognition models, EdgeCrafter (EC)~\cite{liu2026edgecrafter}, which is built on pretrained ECViT backbones.

\subsection{Looped Backbone Construction}
\label{sec:loop_construction}

As shown in Fig.~\ref{fig:main}~(a), given a pretrained ViT backbone, LoopDVR first converts it into a looped ViT. 
The main challenge in this conversion is that naive parameter sharing may disrupt the original hierarchical representations and depth-specific behaviors~\cite{zhang2022minivit}.
To address this challenge, we optimize the looped architecture from two perspectives, \ie, where to loop and how to loop.
Specifically, we determine which layers should share parameters based on representational dissimilarity, while using group-end initialization and Loop Adapters to make parameter sharing more effective.

\label{sec:partitioning}
\vspace{0.5em}
\noindent\textbf{Representation-Based Grouping.}
Directly sharing the weights across different layers disrupts the original hierarchical features.
Following Raptor~\cite{jacobs2026block}, we adopt representation-based grouping, which places adjacent layers with similar representations into the same sharing group.
All layers within a group reuse a Transformer block, while different groups maintain independent block parameters.

\begin{figure}
    \centering
    \begin{subfigure}[t]{0.49\linewidth}
        \centering
        \includegraphics[width=\linewidth]{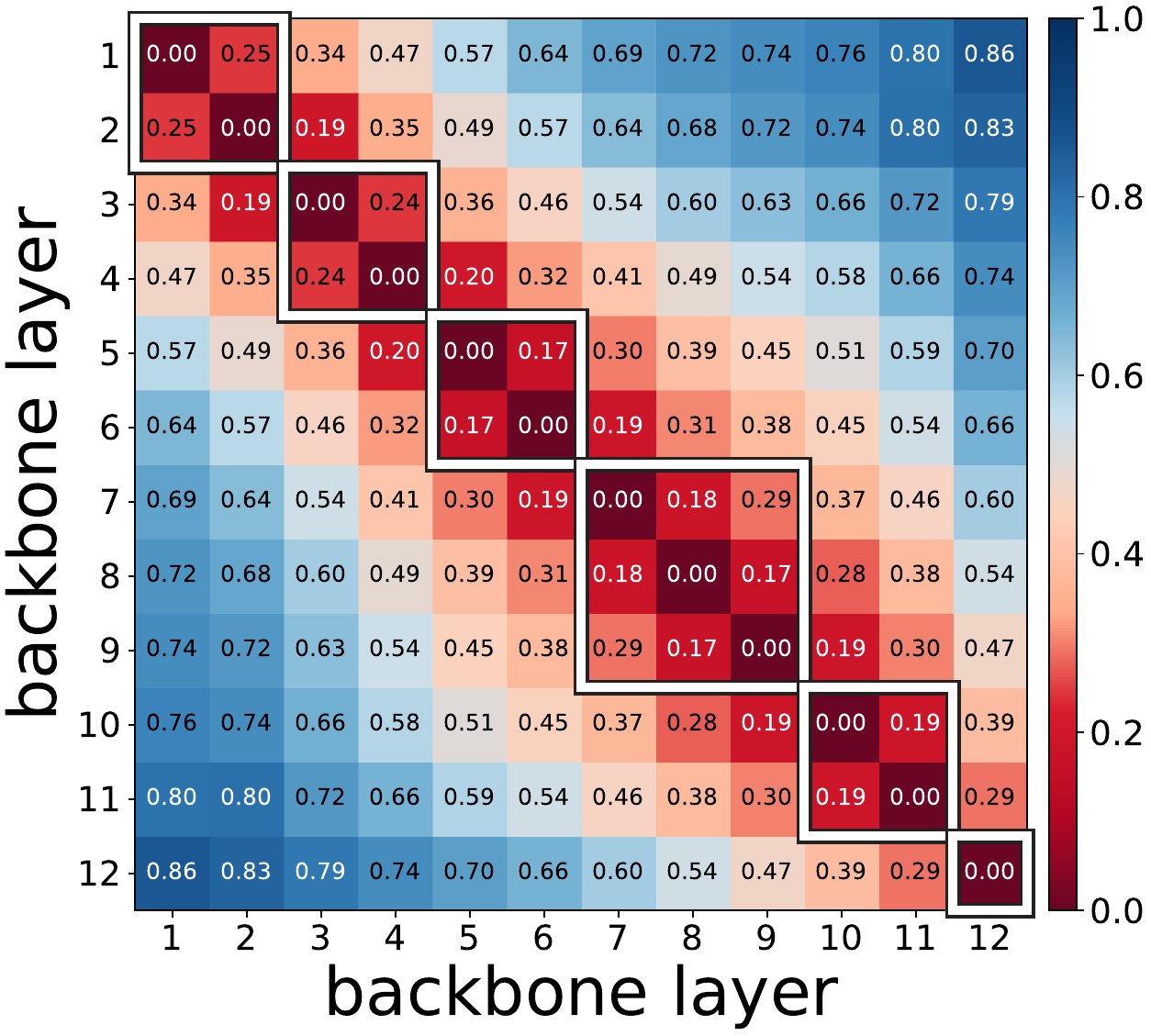}
        \caption{Backbone}
        \label{fig:maxcut_backbone}
    \end{subfigure}
    \hfill
    \begin{subfigure}[t]{0.49\linewidth}
        \centering
        \includegraphics[width=\linewidth]{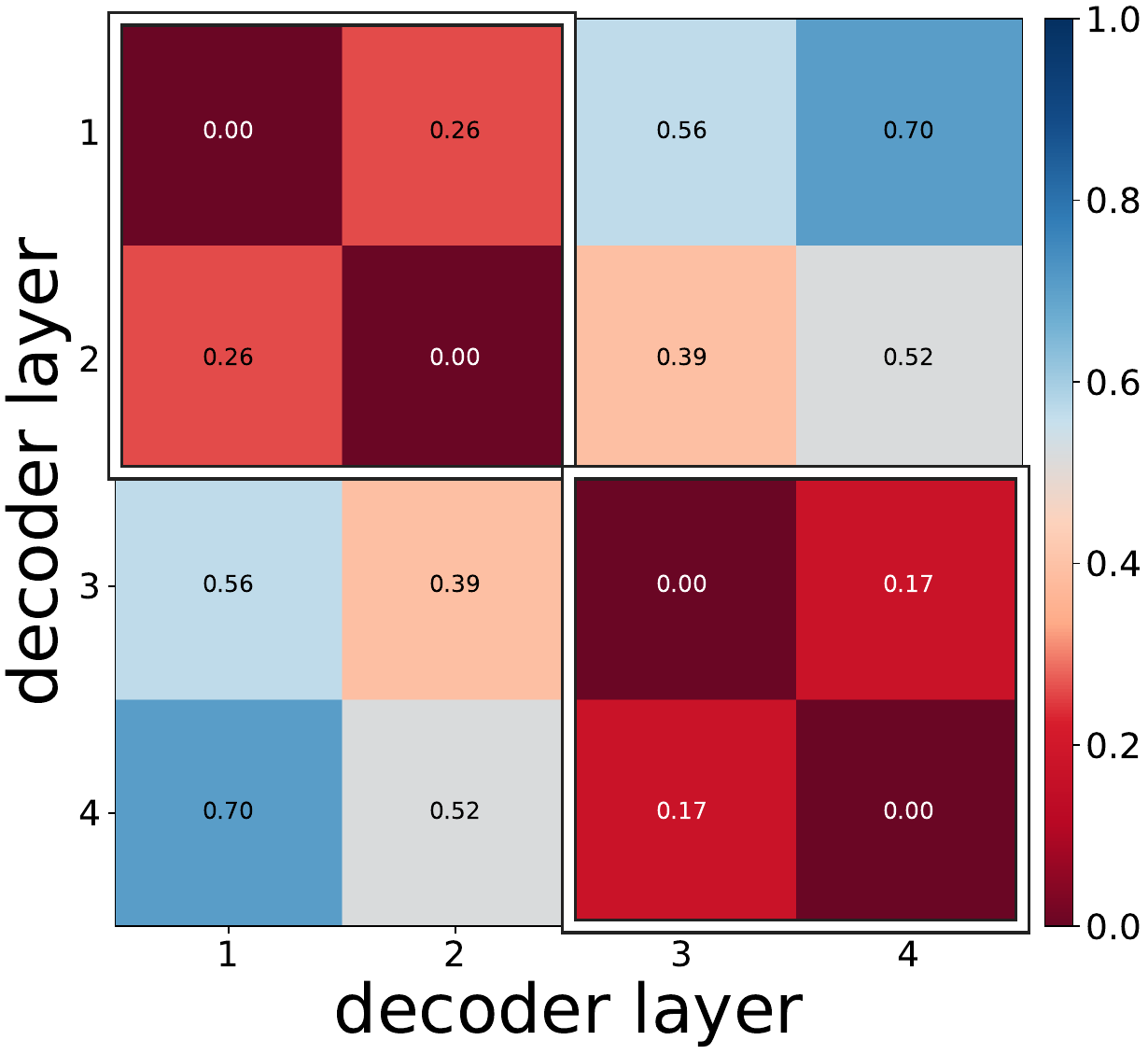}
        \caption{Decoder}
        \label{fig:maxcut_decoder}
    \end{subfigure}
    \vspace{-0.5em}
    \caption{
    \textbf{Grouping via layer-wise representation similarity.}
    Cosine-distance heatmaps of layer representations from (a) the distilled backbone and (b) the post-trained decoder (ECDet-S~\cite{liu2026edgecrafter} as an example). 
    Low-distance regions indicate that pairs of nearby layers have similar representations.
    Boxes mark the selected groups.
}
\end{figure}

We measure the dissimilarity between layers $i$ and $j$ using their mean cosine distance $d_{ij}$ over $N$ images:
\begin{equation}
    d_{ij}
    =
    \frac{1}{N}\sum_{n=1}^{N}
    \left[
    1-\cos(\mathbf z_i^{(n)},\mathbf z_j^{(n)})
    \right],
    \label{eq:activation_distance}
\end{equation}
where $\mathbf z_i^{(n)}$ denotes the average of patch tokens at layer $i$ for the $n$-th input image.
A smaller $d_{ij}$ indicates greater representation similarity between the two layers.
As shown in Fig.~\ref{fig:maxcut_backbone}, the inter-layer similarities of the pretrained ECViT~\cite{liu2026edgecrafter} exhibit clear grouping structure, suggesting that adjacent layers within each group learn similar representations.
Guided by these patterns, we introduce parameter sharing within each group.

Based on these dissimilarities, we determine the sharing groups by minimizing the total within-group dissimilarity.
Unlike Raptor, which uses the last class token for classification, our dense recognition architecture uses both the penultimate and final layer representations to construct multi-scale features.
Therefore, we keep the terminal block as a group with independent parameters, and partition the preceding $L-1$ layers into $K_B-1$ consecutive groups.
Let \(\Pi_{K_B-1}^{L-1}\) denote the set of partitions of the \(L-1\) layers into \(K_B-1\) groups. 
The resulting backbone grouping is
\begin{equation}
    \mathcal P_B =
    \Bigl(
        \operatorname*{arg\,min}_{\mathcal P\in\Pi_{K_B-1}^{L-1}}
        \sum_{\mathcal C\in\mathcal P}
        \sum_{\substack{i,j\in\mathcal C\\ i<j}} d_{ij}
    \Bigr)
    \cup \bigl\{\{L\}\bigr\}.
    \label{eq:contiguous_partition}
\end{equation}
Here, \(\mathcal C\) denotes a group in the candidate partition \(\mathcal P\), and \(\arg\min\) denotes one minimizing partition.

\vspace{0.5em}
\noindent\textbf{Structured Initialization.}
\label{sec:initialization}
Unlike approaches that randomly initialize shared blocks, LoopDVR aims to transform pretrained models to fully leverage their prior knowledge.
This raises a key question: which pretrained weights should each shared block inherit? 
For each group, we compare three strategies: copying the weights of its first layer, copying those of its last layer, or averaging the weights across all layers within the group~\cite{xu2024initializing,samragh2023weight}.

\begin{figure}[t]
    \centering
    \includegraphics[width=\linewidth]{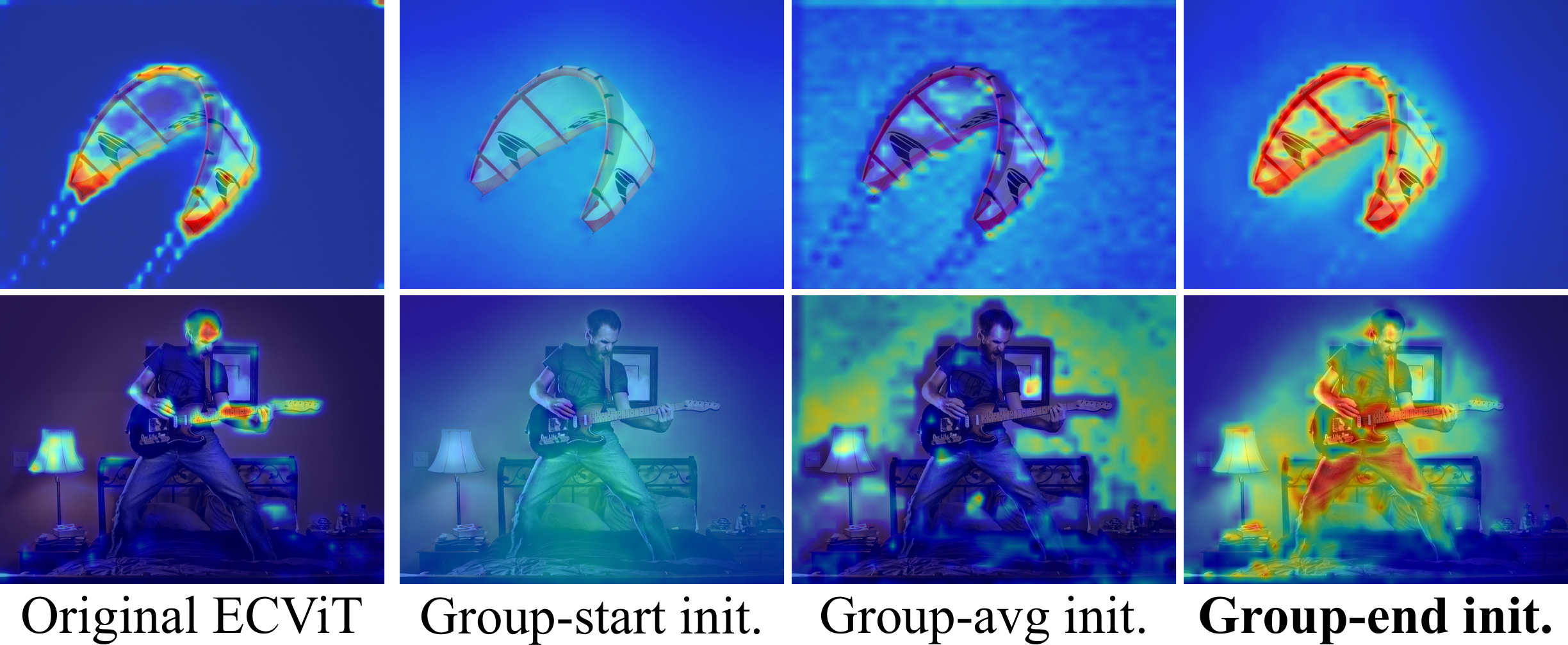}
    \vspace{-1.5em}
    \caption{\textbf{Qualitative comparison of backbone attention at initialization.}
    Attention maps at the final layer of the ECViT and looped backbones initialized with different strategies.
    The looped backbones are visualized before distillation and training.}
    \label{fig:init_attention}
\end{figure}

To examine their effects, we visualize attention maps at the final backbone layer under different initialization strategies, before backbone distillation. 
As shown in Fig.~\ref{fig:init_attention}, group-end initialization produces the most discriminative attention patterns among the three strategies, with attention predominantly focused on target objects.
This attention pattern also most closely resembles that of the original network.
We therefore initialize each shared backbone block with the weights of its group’s terminal layer in the same reference backbone.
Appendix~\ref{sec:supp_init_attention} provides layer-wise comparisons of attention maps, while the ablation in Sec.~\ref{sec:ablation} confirms its advantage after training.

Formally, let the $k$-th backbone group be represented by $\mathcal C_k=[b_k,b_{k+1})$.
We initialize the corresponding shared Transformer block $\mathcal{B}_k$ by copying the parameters of the last reference layer in the group according to
\begin{equation}
    \boldsymbol{\theta}_k
    \leftarrow
    \boldsymbol{\theta}^{\mathrm{ref}}_{b_{k+1}-1}.
    \label{eq:phase_end_initialization}
\end{equation}
Here, $\boldsymbol{\theta}_k$ denotes the parameters of the shared block $\mathcal{B}_k$, and $\boldsymbol{\theta}^{\mathrm{ref}}_{b_{k+1}-1}$ denotes those of the terminal reference layer in group $\mathcal C_k$.
We transfer all structurally compatible parameters, including attention and FFN modules.

\vspace{0.5em}
\noindent\textbf{Loop Adapters.}
\label{sec:adapter}
Strict weight sharing may disrupt the original depth-specific behaviors of the network, while gradients from different layers acting on the same shared parameters can lead to training instability~\cite{zhang2022minivit}.
To address this, we introduce a lightweight Loop Adapter with unique parameters for each layer of the backbone and decoder~\cite{rebuffi2017learning}, enabling depth-specific feature transformations.

As illustrated in Fig.~\ref{fig:main}(b), each application of a  shared Transformer block is followed by a depth-specific adapter that applies a non-linear residual projection, following SReT~\cite{shen2022sliced}. 
Specifically, the intermediate representation $\widetilde{\mathbf X}^{(l)}$ produced by the shared block at backbone layer $l$ is transformed as
\begin{equation}
    \begin{aligned}
        \mathbf X^{(l)}
        &= \mathcal A_l^{B}\!
        \left(\widetilde{\mathbf X}^{(l)}\right)\\
        &= \alpha_l\widetilde{\mathbf X}^{(l)}
        + \beta_l\operatorname{MLP}_l\!
        \left(
            \operatorname{LN}_l\!
            \left(\widetilde{\mathbf X}^{(l)}\right)
        \right).
    \end{aligned}
    \label{eq:adapter_form}
\end{equation}
Here, $\alpha_l$ and $\beta_l$ are learnable scaling coefficients, while $\operatorname{LN}_l$ and $\operatorname{MLP}_l$ have parameters specific to layer $l$.
Different from SReT, we initialize each Loop Adapter as an identity mapping by setting $\alpha_l=1$ and $\beta_l=0$ to preserve the pretrained representations at the beginning of training.
The MLP parameters are randomly initialized.
During training, Loop Adapters are jointly optimized with shared Transformer blocks.
We also try a wide range of depth-conditioning strategies and different designs of the adapters in Sec.~\ref{sec:ablation}.

\subsection{Adaptation to Dense Recognition Tasks}

\label{sec:training}

After obtaining the looped ViT backbone in Sec.~\ref{sec:loop_construction}, we adapt it for dense recognition through backbone distillation and downstream task-specific training.

\vspace{0.5em}
\noindent\textbf{Backbone Distillation.}
Although group-end initialization provides a strong starting point, parameter sharing changes the feature transformation process.
To adapt the looped backbone to this change, we distill visual representations from a pretrained teacher, as shown in Fig.~\ref{fig:main}(c).
We use a DINOv3~\cite{simeoni2025dinov3}, which is adapted for detection following EdgeCrafter~\cite{liu2026edgecrafter}, as the teacher for distillation~\cite{hinton2015distilling,romero2015fitnetshintsdeepnets}.
Specifically, we concatenate the features from the teacher's last two layers and align them with the student's final-layer features:
\begin{equation}
    \mathcal L_{\mathrm{distill}}
    =
    \operatorname{MSE}\left(
    \mathcal \phi(\mathbf X_L^{\mathrm S}),
    \operatorname{Concat}\left(
    \mathbf X_{L_T-1}^{\mathrm T},
    \mathbf X_{L_T}^{\mathrm T}
    \right)
    \right),
\end{equation}
where $\mathcal \phi$ denotes a linear projector, and $\mathbf S$ and $\mathbf T$ denote the student and teacher, respectively.
Shared Transformer blocks, their Loop Adapters and the projector are optimized jointly.

\vspace{0.5em}
\noindent\textbf{Looped Decoder Construction.}
Similar to the backbone, the Transformer decoder exhibits clear grouping structure in inter-layer representation dissimilarity, as shown in Fig.~\ref{fig:maxcut_decoder}.
Accordingly, we extend the same group-wise sharing and Loop Adapter design to the decoder, partitioning its \(T\) refinement layers into \(K_D\) groups.
The prediction heads are also shared along with the Transformer blocks.
To determine the grouping, we compute the dissimilarities using trained, non-looped reference models.
The resulting decoder partition is consistent across different models and tasks (Fig.~\ref{fig:supp_phase_partitions} in the Appendix), allowing us to adopt a unified grouping for all decoders.
Since ECViT~\cite{liu2026edgecrafter} focuses only on optimizing backbones and decoders require task-specific post-training, we train looped decoders from scratch for downstream tasks.

\vspace{0.5em}
\noindent\textbf{Downstream Task-Specific Training.}
As illustrated in Fig.~\ref{fig:main}(d), we integrate the distilled looped ViT with a hybrid encoder and the looped decoder for downstream dense recognition tasks.
We train the resulting models for object detection, instance segmentation, and human pose estimation using DETR-style bipartite matching and task-specific supervision~\cite{carion2020detr}.
For instance segmentation and pose estimation, we initialize the backbone from the trained detector and extend the architecture with a query-based mask head and a pose head, respectively~\cite{maskdino,janampa2026detrpose}.
Note that implementation details are provided in Appendix~\ref{sec:training_config}.

\begin{table*}[!t]
\begingroup
\captionsetup{belowskip=1.5em}
\renewenvironment{table*}[1][]{\begingroup}{\par\endgroup}
\begin{table*}[t]
\captionsetup{hypcap=false}
\caption{\textbf{Object detection results on COCO~\texttt{val2017}~\cite{lin2014microsoft}, grouped by parameter budget.}
Results with $^\dagger$ use Objects365~\cite{shao2019objects365} pretraining. 
\textcolor{gray}{ECDet}~\cite{liu2026edgecrafter} are our reference models for compression. \textbf{Bold results} indicate the best results within each group, excluding ECDet.
\textcolor{orange}{Orange results} indicate that our models outperform the corresponding baseline. 
More details of evaluation settings are in the Appendix~\ref{sec:inference_protocol}.} 
\vspace{-0.5em}
\label{tab:loopdet}
	\centering
    \small
		\begin{autotabular}{12}{l | ccccc | ccc ccc}
        \toprule
        Model & \#Epochs & \#Params. (M) & Peak mem. (MiB) & GFLOPs &
        Lat. (ms) & AP$^{val}$ & AP$^{val}_{50}$ & AP$^{val}_{75}$ &
        AP$^{val}_S$ & AP$^{val}_M$ & AP$^{val}_L$ \\
        \hline

        {DEIMv2}-S~\cite{huang2025deimv2} & 132 & \parambase{9.7} & \parambase{21.1} & 49.5 & 1.56 & \textbf{50.9} & 68.4 & \textbf{55.1} & 31.3 & 55.3 & \textbf{70.2} \\

        YOLO26-S~\cite{yolo26} & 70 & \parambase{9.5} & \parambase{42.2} & 20.9 & 2.59 & 48.6$^\dagger$ & 65.8$^\dagger$ & 52.8$^\dagger$ & 29.5$^\dagger$ & 53.2$^\dagger$ & 65.8$^\dagger$ \\

        RF-DETR-P~\cite{robinson2025rf} & -- & \parambase{8.5} & \parambase{18.8} & 17.7 & 0.75 & 41.6 & 60.2 & 44.1 & 19.2 & 44.7 & 61.7 \\

        \color[gray]{0.6} ECDet-S~\cite{liu2026edgecrafter} & \color[gray]{0.6} 74 & \color[gray]{0.6} \parambase{9.8} & \color[gray]{0.6} \parambase{21.3} & \color[gray]{0.6} 50.0 & \color[gray]{0.6} 1.51 & \color[gray]{0.6} 51.7 & \color[gray]{0.6} 69.4 & \color[gray]{0.6} 55.8 & \color[gray]{0.6} 32.3 & \color[gray]{0.6} 56.4 & \color[gray]{0.6} 70.5 \\

        \rowcolor[gray]{0.9}
        \textbf{LoopDet-S (ours)} & 74 & \paramdrop{7.0}{29} & \paramdrop{15.9}{25} & 53.0 & 1.62 & \textbf{50.9} & \textbf{68.6} & 54.9 & \textbf{31.4} & \textbf{55.9} & 69.8 \\

        \hline

        YOLOv12-M-turbo~\cite{tian2025yolov12} & 600 & \parambase{19.6} & \parambase{51.8} & 63.1 & 2.96 & 52.5 & 69.9 & 57.1 & \textbf{35.2} & 57.8 & 69.7 \\

        DEIMv2-M~\cite{huang2025deimv2} & 102 & \parambase{18.1} & \parambase{37.2} & 84.1 & 2.19 & 53.0 & 70.2 & 57.6 & 34.2 & 57.4 & 71.5 \\

        RT-DETRv4-M~\cite{liao2025rtdetrv4} &  102 & \parambase{19.4} & \parambase{40.3} & 57.0 & 1.53 & 53.5 & 71.1 & 58.1 & 34.9 & 57.7 & 72.1 \\

        \color[gray]{0.6} {ECDet}-M~\cite{liu2026edgecrafter} & \color[gray]{0.6} \color[gray]{0.6} 62 & \color[gray]{0.6} \parambase{19.2} & \color[gray]{0.6} \parambase{39.3} & \color[gray]{0.6} 84.9 & \color[gray]{0.6} 1.92 & \color[gray]{0.6} 54.3 & \color[gray]{0.6} 72.2 & \color[gray]{0.6} 58.7 & \color[gray]{0.6} 35.9 & \color[gray]{0.6} 59.1 & \color[gray]{0.6} 72.7 \\

        \rowcolor[gray]{0.9}
        \textbf{LoopDet-M (ours)}
        & 62 & \paramdrop{13.9}{28} & \paramdrop{29.1}{26} & 90.3 & 2.03 & \textbf{54.0} & \textbf{71.9} & \textbf{58.4} & 34.2 & \textcolor{orange}{\textbf{59.1}} & \textcolor{orange}{\textbf{73.0}} \\

        \hline

        YOLOv12-L-turbo~\cite{tian2025yolov12} & 600 & \parambase{26.5} & \parambase{62.2} & 88.9 & 3.79 & 53.8 & 71.0 & 58.6 & 36.9 & 59.4 & 71.0 \\

        YOLO26-L~\cite{yolo26} & 60 & \parambase{24.8} & \parambase{65.3} & 86.8 & 3.35 & 55.0$^\dagger$ & 72.5$^\dagger$ & 60.0$^\dagger$ & \textbf{38.4}$^\dagger$ & 59.5$^\dagger$ & 71.1$^\dagger$ \\

        \color[gray]{0.6} ECDet-L~\cite{liu2026edgecrafter} & \color[gray]{0.6} \color[gray]{0.6} 50 & \color[gray]{0.6} \parambase{32.7} & \color[gray]{0.6} \parambase{65.7} & \color[gray]{0.6} 147.2 & \color[gray]{0.6} 2.60 & \color[gray]{0.6} 57.0 & \color[gray]{0.6} 75.1 & \color[gray]{0.6} 61.7 & \color[gray]{0.6} 38.7 & \color[gray]{0.6} 62.5 & \color[gray]{0.6} 75.0 \\

        \rowcolor[gray]{0.9}
        \textbf{LoopDet-L (ours)}
        & 50 & \paramdrop{23.4}{28} & \paramdrop{47.3}{28} & 158.9 & 2.82 & \textbf{56.6} & \textbf{74.8} & \textbf{61.4} & 37.4 & \textbf{62.1} & \textcolor{orange}{\textbf{75.1}}\\

		\hline

        DEIMv2-L~\cite{huang2025deimv2}  & 68  & \parambase{32.2} & \parambase{64.9} & 144.5 & 2.69 & 56.0 & 73.5 & 61.1 & 37.6 & 60.9 & 74.9 \\

        RT-DETRv4-L~\cite{liao2025rtdetrv4} & 58  & \parambase{31.0} & \parambase{61.9} & 91.6 & 2.22 & 55.4 & 73.0 & 60.3 & 37.1 & 60.1 & 72.9 \\

        RF-DETR-L~\cite{robinson2025rf} & --  & \parambase{33.8} & \parambase{66.9} & 99.6 & 1.67 & 56.5$^\dagger$ & 75.1$^\dagger$ & 61.3$^\dagger$ & 39.0$^\dagger$ & 61.0$^\dagger$ & 73.9$^\dagger$ \\ %

         \color[gray]{0.6} ECDet-X~\cite{liu2026edgecrafter} & \color[gray]{0.6} \color[gray]{0.6} 50 & \color[gray]{0.6} \parambase{49.0} & \color[gray]{0.6} \parambase{96.4} & \color[gray]{0.6} 204.5 & \color[gray]{0.6} 3.12 & \color[gray]{0.6}  57.9 & \color[gray]{0.6}  76.0 & \color[gray]{0.6}  62.9 & \color[gray]{0.6}  38.7 & \color[gray]{0.6}  63.4 & \color[gray]{0.6}  76.1 \\

        \rowcolor[gray]{0.9}
        \textbf{LoopDet-X (ours)}
        & 50 & \paramdrop{35.1}{28} & \paramdrop{69.8}{28} & 216.2 & 3.33 & \textbf{57.5} & \textbf{75.7} & \textbf{62.4} & \textcolor{orange}{\textbf{40.1}} & \textcolor{orange}{\textbf{63.5}} & \textbf{75.1} \\

		\bottomrule
	\end{autotabular}
\end{table*}

\par\vspace{1em}
\begin{table*}[t]
\captionsetup{hypcap=false}
\caption{\textbf{Instance segmentation results on COCO~\texttt{val2017}, grouped by parameter budget.}
Results with $^\dagger$ denote models trained with additional data beyond COCO.
\textcolor{gray}{ECInsSeg}~\cite{liu2026edgecrafter} are our reference models for compression. \textbf{Bold results} indicate the best results within each group, excluding ECInsSeg.
\textcolor{orange}{Orange results} indicate that our models outperform the corresponding ECInsSeg with much fewer parameters.
}
\vspace{-1em}
\label{tab:loopseg}
	\centering
    \small
		\begin{autotabular}{11}{l | cccc | ccc ccc}
        \toprule
        Model & \#Params. (M) & Peak mem. (MiB) & GFLOPs &
        Lat. (ms) & AP$^{val}$ & AP$^{val}_{50}$ & AP$^{val}_{75}$ &
        AP$^{val}_S$ & AP$^{val}_M$ & AP$^{val}_L$ \\
        \hline

        YOLO-MS-XS-Seg~\cite{yoloms} & \parambase{5.1} & \parambase{46.2} & 26.2 & 39.52 & 36.6 & 56.7 & 39.1 & 15.7 & 40.5 & 56.0\\

        YOLOv12-Seg-S~\cite{tian2025yolov12} & \parambase{9.8} & \parambase{46.4} & 33.4 & 2.88 & 38.6 & 60.8 & 40.7 & 17.0 & 42.1 & 59.2 \\

        \color[gray]{0.6} ECInsSeg-S~\cite{liu2026edgecrafter} & \color[gray]{0.6} \parambase{10.3} & \color[gray]{0.6} \parambase{22.3} & \color[gray]{0.6} 57.0 & \color[gray]{0.6} 1.59 & \color[gray]{0.6} 43.0 & \color[gray]{0.6} 65.7 & \color[gray]{0.6} 46.0 & \color[gray]{0.6} 20.8 & \color[gray]{0.6} 46.3 & \color[gray]{0.6} 65.9 \\

        \rowcolor[gray]{0.9}
        \textbf{LoopSeg-S (ours)} & \paramdrop{7.4}{28} & \paramdrop{16.7}{25} & 60.0 & 1.72 & \textcolor{orange}{\textbf{43.0}} & \textbf{65.3} & \textcolor{orange}{\textbf{46.2}} & \textbf{20.2} & \textcolor{orange}{\textbf{46.5}} & \textcolor{orange}{\textbf{65.9}}\\

        \hline

        YOLO11-Seg-S~\cite{yolo11} & \parambase{10.1} & \parambase{51.2} & 33.6 & 2.64 & 37.8 & 59.9 & 40.4 & 19.8 & 42.7 & 55.2 \\

        YOLO26-Seg-S~\cite{yolo26} & \parambase{10.4} & \parambase{46.1} & 35.0 & 3.07 & 40.2$^\dagger$ & 62.0$^\dagger$ & 43.0$^\dagger$ & 18.8$^\dagger$ & 44.0$^\dagger$ & 61.0$^\dagger$ \\

        \color[gray]{0.6} ECInsSeg-M~\cite{liu2026edgecrafter} & \color[gray]{0.6} \parambase{20.1} & \color[gray]{0.6} \parambase{41.0} & \color[gray]{0.6} 96.1  & \color[gray]{0.6} 2.00 & \color[gray]{0.6} 45.2 & \color[gray]{0.6} 68.2 & \color[gray]{0.6} 48.3 & \color[gray]{0.6} 22.9 & \color[gray]{0.6} 49.0 & \color[gray]{0.6} 68.1 \\

        \rowcolor[gray]{0.9}
        \textbf{LoopSeg-M (ours)} & \paramdrop{14.6}{27} & \paramdrop{30.5}{26} & 101.4 & 2.18 & \textcolor{orange}{\textbf{45.5}} & \textcolor{orange}{\textbf{68.4}} & \textcolor{orange}{\textbf{48.8}} & \textbf{22.4} & \textcolor{orange}{\textbf{49.3}} & \textcolor{orange}{\textbf{68.5}} \\

		\hline

        YOLO11-Seg-L~\cite{yolo11} &  \parambase{27.6} & \parambase{79.2} & 132.9 & 4.20 & 42.9 & 67.0 & 46.2 & 24.6 & 48.7 & 60.6 \\

        YOLOv12-Seg-L~\cite{tian2025yolov12} & \parambase{28.8} & \parambase{70.5} & 134.7 & 4.42 & 43.1 & 67.3 & 45.8 & 22.3 & 47.9 & 67.9 \\

		YOLO26-Seg-M~\cite{yolo26} & \parambase{23.6} & \parambase{63.8} & 122.2 & 3.74 & 44.2$^\dagger$ & 67.4$^\dagger$ & 47.6$^\dagger$ & 23.0$^\dagger$ & 48.7$^\dagger$ & 64.1$^\dagger$ \\

        \color[gray]{0.6} ECInsSeg-L~\cite{liu2026edgecrafter} & \color[gray]{0.6} \parambase{33.6} & \color[gray]{0.6} \parambase{67.5} & \color[gray]{0.6} 158.3 & \color[gray]{0.6} 2.74 & \color[gray]{0.6} 47.1 & \color[gray]{0.6} 70.9 & \color[gray]{0.6} 50.5 & \color[gray]{0.6} 24.8 & \color[gray]{0.6} 51.1 & \color[gray]{0.6} 69.6 \\

        \rowcolor[gray]{0.9}
        \textbf{LoopSeg-L (ours)} & \paramdrop{24.1}{28} & \paramdrop{48.7}{28} & 170.0 & 2.98 & \textcolor{orange}{\textbf{48.1}} & \textcolor{orange}{\textbf{71.7}} & \textcolor{orange}{\textbf{51.7}} & \textcolor{orange}{\textbf{24.9}} & \textcolor{orange}{\textbf{52.2}} & \textcolor{orange}{\textbf{71.0}} \\

        \hline

        RF-DETR-Seg-L~\cite{robinson2025rf} & \parambase{36.2}  & \parambase{70.5} & 125.3 & 2.02 & 47.1$^\dagger$ & 70.5$^\dagger$ & 50.9$^\dagger$ & 28.4$^\dagger$ & 52.1$^\dagger$ & 65.6$^\dagger$ \\

        RF-DETR-Seg-X~\cite{robinson2025rf} & \parambase{38.1}  & \parambase{75.3} & 218.4 & 3.23 & 48.8$^\dagger$ & 72.2$^\dagger$ & \textbf{53.1}$^\dagger$ & \textbf{30.6}$^\dagger$ & \textbf{53.3}$^\dagger$ & 65.9$^\dagger$ \\

        \color[gray]{0.6} ECInsSeg-X~\cite{liu2026edgecrafter}  & \color[gray]{0.6} \parambase{49.9} & \color[gray]{0.6} \parambase{98.2} & \color[gray]{0.6} 215.6 & \color[gray]{0.6} 3.27 & \color[gray]{0.6} 48.4 & \color[gray]{0.6} 72.2 & \color[gray]{0.6} 52.0 & \color[gray]{0.6} 26.3 & \color[gray]{0.6} 52.7 & \color[gray]{0.6} 71.1 \\

        \rowcolor[gray]{0.9}
        \textbf{LoopSeg-X (ours)} & \paramdrop{35.8}{28} & \paramdrop{71.2}{27} & 227.3 & 3.47 & \textcolor{orange}{\textbf{48.9}} & \textcolor{orange}{\textbf{72.7}} & 51.7 & 26.0 & \textcolor{orange}{53.2} & \textcolor{orange}{\textbf{71.4}} \\

        \bottomrule
	\end{autotabular}
\end{table*}

\endgroup
\end{table*}

\section{Experiments}
\label{sec:experiments}

We evaluate LoopDVR on three dense visual recognition tasks using LoopDet for object detection, LoopSeg for instance segmentation, and LoopPose for human pose estimation.
We instantiate these models based on the ECDet, ECInsSeg, and ECPose architectures from EdgeCrafter~\cite{liu2026edgecrafter}, respectively.
Before task-specific training, we distill the looped backbone using unlabeled images from ImageNet-1K~\cite{imagenet15russakovsky} and COCO~\texttt{train2017}~\cite{lin2014microsoft}.
We then train four model scales for each task on COCO~\texttt{train2017} and evaluate them on COCO~\texttt{val2017}.
Training is conducted using $4\times$ NVIDIA RTX 4090 GPUs.

\subsection{Main Result}

\noindent\textbf{Object Detection.}
As summarized in Tab.~\ref{tab:loopdet}, we compare our proposed LoopDet with recent compact object detectors under similar parameter budgets on COCO~\texttt{val2017}, including the corresponding ECDet~\cite{liu2026edgecrafter} as reference baselines.
LoopDet achieves the best accuracy–parameter trade-offs across all four parameter-budget ranges.
Specifically, LoopDet-S achieves 50.9~AP on COCO with only 7M params, highlighting strong performance of LoopDVR under a highly constrained parameter budget.
LoopDet-L achieves 99.3\% of ECDet-L's AP with 28.4\% fewer parameters.
Moreover, LoopDet demonstrates favorable peak GPU memory usage during inference compared with competing detectors.

\vspace{0.5em}
\noindent\textbf{Instance Segmentation.}
As shown in Tab.~\ref{tab:loopseg}, we compare LoopSeg with recent instance segmentation models under comparable parameter budgets.
LoopSeg achieves the highest mask AP across all four parameter-budget ranges.
Notably, all scales of LoopSeg match or outperform their corresponding reference models despite using approximately 28\% fewer parameters.
In particular, LoopSeg-L achieves 48.1 mask AP, improving upon ECInsSeg-L by 1.0 AP with 28\% fewer parameters.
These results demonstrate that LoopDVR improves parameter efficiency while preserving or even enhancing instance segmentation performance.

\begin{table}[t]
\captionsetup{hypcap=false}
\caption{\textbf{Performance comparison of pose estimation models on COCO~\texttt{val2017}, grouped by parameter budget.}
Results with $^\dagger$ denote models trained with additional data beyond COCO.
\textcolor{gray}{ECPose}~\cite{liu2026edgecrafter} are our reference models. \textbf{Bold results} indicate the best results within each group, excluding ECPose.
}
\label{tab:looppose}
\centering
\small
\begin{autotabular}{5}{l cccc}
\toprule
Model & \#Params. (M) & Peak mem. & Lat. (ms) & AP \\
\hline

RTMO-S~\cite{rtmo} & \parambase{9.9} & 93.2\,MiB & 7.82 & 67.7 \\

YOLO11-Pose-S~\cite{yolo11} & \parambase{9.9} & 30.2\,MiB & 3.00 & 58.9 \\

\color[gray]{0.6} ECPose-S~\cite{liu2026edgecrafter} & \color[gray]{0.6} \parambase{9.9} & \color[gray]{0.6} 30.6\,MiB & \color[gray]{0.6} 2.65 & \color[gray]{0.6} 68.9 \\

\rowcolor[gray]{0.9}
\textbf{LoopPose-S (ours)} & \paramdrop{7.7}{23} & 26.3\,MiB & 2.98 & \textbf{68.2} \\

\hline

YOLO26-Pose-S~\cite{yolo26} & \parambase{10.4} & 31.0\,MiB & 2.92 & 62.4 \\

DETRPose-S~\cite{janampa2026detrpose} & \parambase{11.5} & 33.8\,MiB & 2.33 & 67.0$^\dagger$ \\

\color[gray]{0.6} ECPose-M~\cite{liu2026edgecrafter} & \color[gray]{0.6} \parambase{19.8} & \color[gray]{0.6} 49.4\,MiB & \color[gray]{0.6} 3.27 & \color[gray]{0.6} 72.4 \\

\rowcolor[gray]{0.9}
\textbf{LoopPose-M (ours)} & \paramdrop{14.6}{26} & 39.5\,MiB & 3.24 & \textbf{71.8} \\

\hline

RTMO-M~\cite{rtmo} & \parambase{22.6} & 76.9\,MiB & 6.51 & 71.0 \\

YOLO26-Pose-L~\cite{yolo26} & \parambase{25.9} & 60.7\,MiB & 4.24 & 69.6 \\

DETRPose-M~\cite{janampa2026detrpose} & \parambase{20.8} & 51.9\,MiB & 2.71 & 69.4$^\dagger$ \\

\color[gray]{0.6} ECPose-L~\cite{liu2026edgecrafter} & \color[gray]{0.6} \parambase{34.3} & \color[gray]{0.6} 77.8\,MiB & \color[gray]{0.6} 3.78 & \color[gray]{0.6} 73.5 \\

\rowcolor[gray]{0.9}
\textbf{LoopPose-L (ours)} & \paramdrop{24.6}{28} & 58.7\,MiB & 3.98 & \textbf{72.9} \\

\hline

RTMO-L~\cite{rtmo} & \parambase{44.8} & 119.3\,MiB & 7.62 & 72.4 \\

DETRPose-L~\cite{janampa2026detrpose} & \parambase{32.8} & 74.5\,MiB & 3.51 & 72.5$^\dagger$ \\

\color[gray]{0.6} ECPose-X~\cite{liu2026edgecrafter} & \color[gray]{0.6} \parambase{50.6} & \color[gray]{0.6} 108.8\,MiB & \color[gray]{0.6} 4.75 & \color[gray]{0.6} 74.8 \\

\rowcolor[gray]{0.9}
\textbf{LoopPose-X (ours)} & \paramdrop{36.3}{28} & 82.0\,MiB & 4.46 & \textbf{74.3} \\

\bottomrule
\end{autotabular}
\end{table}

\vspace{0.5em}
\noindent\textbf{Human Pose Estimation.}
Tab.~\ref{tab:looppose} compares LoopPose with recent human pose estimation models under comparable parameter budgets, including the corresponding ECPose~\cite{liu2026edgecrafter} as reference baselines.
LoopPose achieves the best accuracy–parameter trade-offs across all four parameter-budget ranges.
Compared with the corresponding ECPose, LoopPose achieves competitive accuracy with fewer parameters and lower memory usage.

\subsection{Analysis}
\label{sec:ablation}

Unless otherwise specified, ablations use the LoopDet-S with $6$ shared Transformer blocks in the backbone and $2$ in the decoder, retaining depths of $12$ and $4$ layers, respectively.

\begin{figure}[t]
    \centering
    \includegraphics[width=0.5\textwidth]{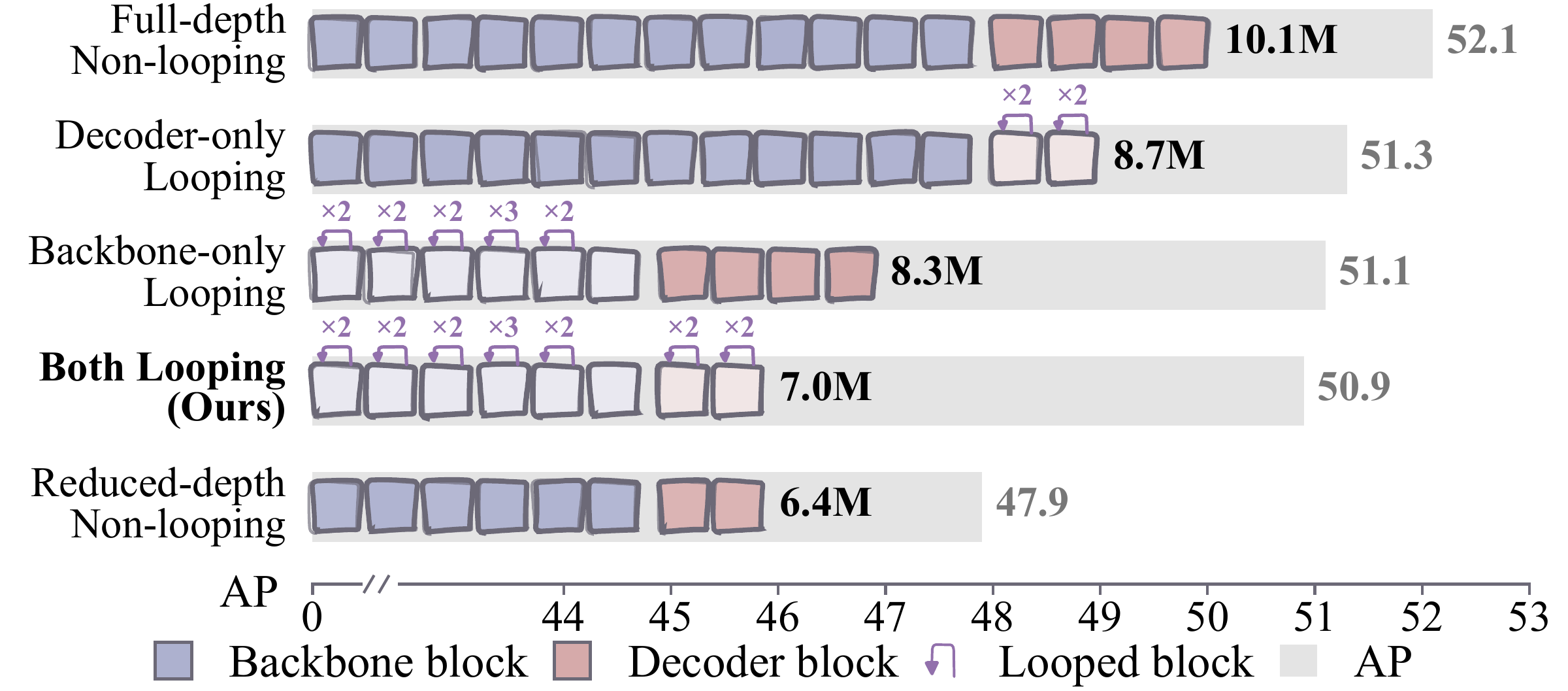}
    \caption{\textbf{Effect of looping.}
    All models follow the same initialization and training protocols, with the Loop Adapters.
    }
    \label{fig:loop_ablation}
\end{figure}

\vspace{0.5em}
\noindent\textbf{Ablation Study on Looping Structure. }
As shown in Fig.~\ref{fig:loop_ablation}, we compare different looping strategies in terms of their parameter–accuracy trade-offs. 
Compared with the reduced-depth non-looped model, our method preserves the full computational depth and improves AP by 3.0 with only 0.6M additional parameters. 
Relative to the full-sized baseline, it reduces the parameter count by 3.2M at the cost of only a 1.2 AP drop, demonstrating favorable parameter efficiency.
Furthermore, looping both the backbone and decoder achieves nearly identical accuracy to backbone-only and decoder-only looping while requiring substantially fewer parameters.
More looping variants are compared in Appendix~\ref{sec:loop_variants}.

\begin{figure}[!t]
    \centering
    \begin{minipage}{\linewidth}
        \centering
        \begin{minipage}[t]{0.49\linewidth}
            \centering
            \includegraphics[width=\linewidth]{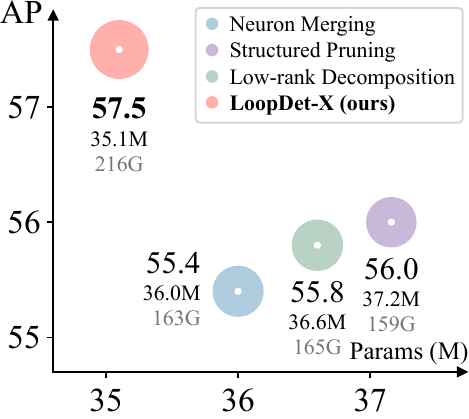}
            \vspace{-1.5em}
            \caption{\textbf{Comparison with other compression methods.}
            Circle areas denote FLOPs.
            All variants are trained with the same protocol after initialization.}
            \label{fig:compression}
        \end{minipage}
        \hfill
        \begin{minipage}[t]{0.48\linewidth}
            \centering
            \includegraphics[width=\linewidth]{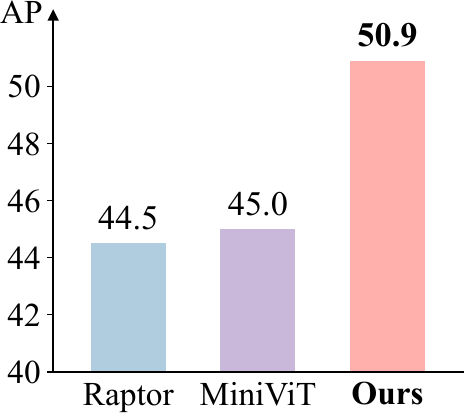}
            \vspace{-1.5em}
            \caption{\textbf{Comparison of distillation strategies.}
            All variants use the same LoopDet-S backbone and teacher, and are trained with the same protocol.}
            \label{fig:distillation}
        \end{minipage}
    \end{minipage}
    \vspace{-0.5em}
\end{figure}

\vspace{0.5em}
\noindent\textbf{Comparison of Parameter Compression Methods.}
To further illustrate the effectiveness of looping, we implement several representative parameter compression methods based on the same baseline (\ie, ECDet-X~\cite{liu2026edgecrafter}), including neuron merging~\cite{theus2024metapruning}, structured pruning~\cite{fang2024isomorphic}, and low-rank decomposition~\cite{FLAR}.
As shown in Fig.~\ref{fig:compression}, our looping strategy achieves the highest AP of 57.5 while using the fewest parameters, indicating a superior parameter–accuracy trade-off.
Implementation details are provided in Appendix~\ref{sec:compression_implement}.

\vspace{0.5em}
\noindent\textbf{Ablation Study on the Independent Terminal Block.}
Tab.~\ref{tab:partition} evaluates the effect of treating the last layer of the backbone as an independent parameter sharing group.
Without reserving the terminal block, representation-based grouping yields uniform grouping and achieves 49.6 AP. 
Reserving one block for the final layer and grouping the preceding layers raises AP to 50.9. 
The 1.3 AP gain supports retaining independent parameters in the final backbone layer.

\begin{table}[h]
\captionsetup{hypcap=false}
\caption{\textbf{Effect of reserving an independent terminal block.}
$^\dagger$ Without terminal block reservation, the result is uniform grouping.}
\label{tab:partition}
    \centering
    \small
    \begin{autotabular}{3}{c c c}
        \toprule
        \makecell{Independent terminal block} &
        Group sizes &
        AP \\
        \hline

         $\times$ & 
        2--2--2--2--2--2$^\dagger$ & 49.6 \\

        \rowcolor[gray]{0.9}
        $\checkmark$ &
        2--2--2--3--2--1& \textbf{50.9} \\
        \bottomrule
    \end{autotabular}
\end{table}

\begin{figure}[t]
    \centering
    \includegraphics[width=1\linewidth]{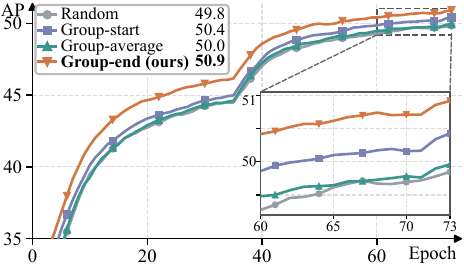}
    \caption{\textbf{Comparison of initialization strategies.}
    All strategies use the fixed LoopDet-S architecture and training protocol.
    }
    \label{fig:init}
\end{figure}

\vspace{0.5em}
\noindent\textbf{Ablation Study on Structured Initialization.}
Fig.~\ref{fig:init} compares different initialization strategies for the shared backbone blocks.
We compare random, group-start, group-average, and group-end initialization, followed by the same distillation for each strategy.
All methods initialized from pre-trained models consistently outperform random initialization throughout the entire training process, demonstrating the critical importance of leveraging pre-trained knowledge.
Among these initialization strategies, group-end initialization achieves the best performance, consistent with the comparison of attention maps in Fig.~\ref{fig:init_attention}.

\vspace{0.5em}
\noindent\textbf{Comparison of Depth Conditioning Methods.}
Prior works have proposed several depth conditioning methods to preserve depth-specific operations of the model.
To further validate the effectiveness of the adapter, we implement existing depth-conditioning methods and other PEFT-style transformation approaches~\cite{houlsby2019parameter} on LoopDet-S.
As shown in Tab.~\ref{tab:PEFT}, the Loop Adapter achieves the highest AP among the tested configurations, improving AP by $1.5$ over the model without conditioning.
The implementation details of comparative methods are provided in Appendix~\ref{sec:conditioning_implement}.

\begin{table}[h]
\captionsetup{hypcap=false}
\caption{\textbf{Comparison of depth conditioning methods.}
We implement weight transformation following MiniViT~\cite{zhang2022minivit} and depth scaling following Raptor~\cite{jacobs2026block}.
}
\label{tab:PEFT}
    \centering
    \small
    \begin{autotabular}{4}{l ccc}
        \toprule
        Method  & \#Params. & GFLOPs & AP \\
        \hline
        w/o conditioning & 5.8M & 50.0 & 49.4 \\
        Depth scaling~\cite{jacobs2026block}~($d_{\mathrm{h}}=64$) & 6.1M & 50.0 & 49.4 \\
        Weight transformation~\cite{zhang2022minivit}  & 5.8M & 51.2 & 49.4\\
        LoRA~\cite{hu2022lora,bae2025relaxed}~(rank=$32$) & 7.4M & 53.7 & 50.1\\
        LN tuning~\cite{ln-tuning} & 5.8M & 50.0 & 49.5 \\
        VPT-deep~\cite{jia2022visual}~($l_p=64$) & 6.0M & 52.8 & 49.4 \\
        \rowcolor[gray]{0.9}
        \textbf{Loop Adapter}~($d_{\mathrm{h}}=192$) & 7.0M & 53.0 & \textbf{50.9} \\
        \bottomrule
    \end{autotabular}
\end{table}

\begin{figure}[t]
    \centering
    \begin{subfigure}[t]{0.49\linewidth}
        \centering
        \includegraphics[width=\linewidth]{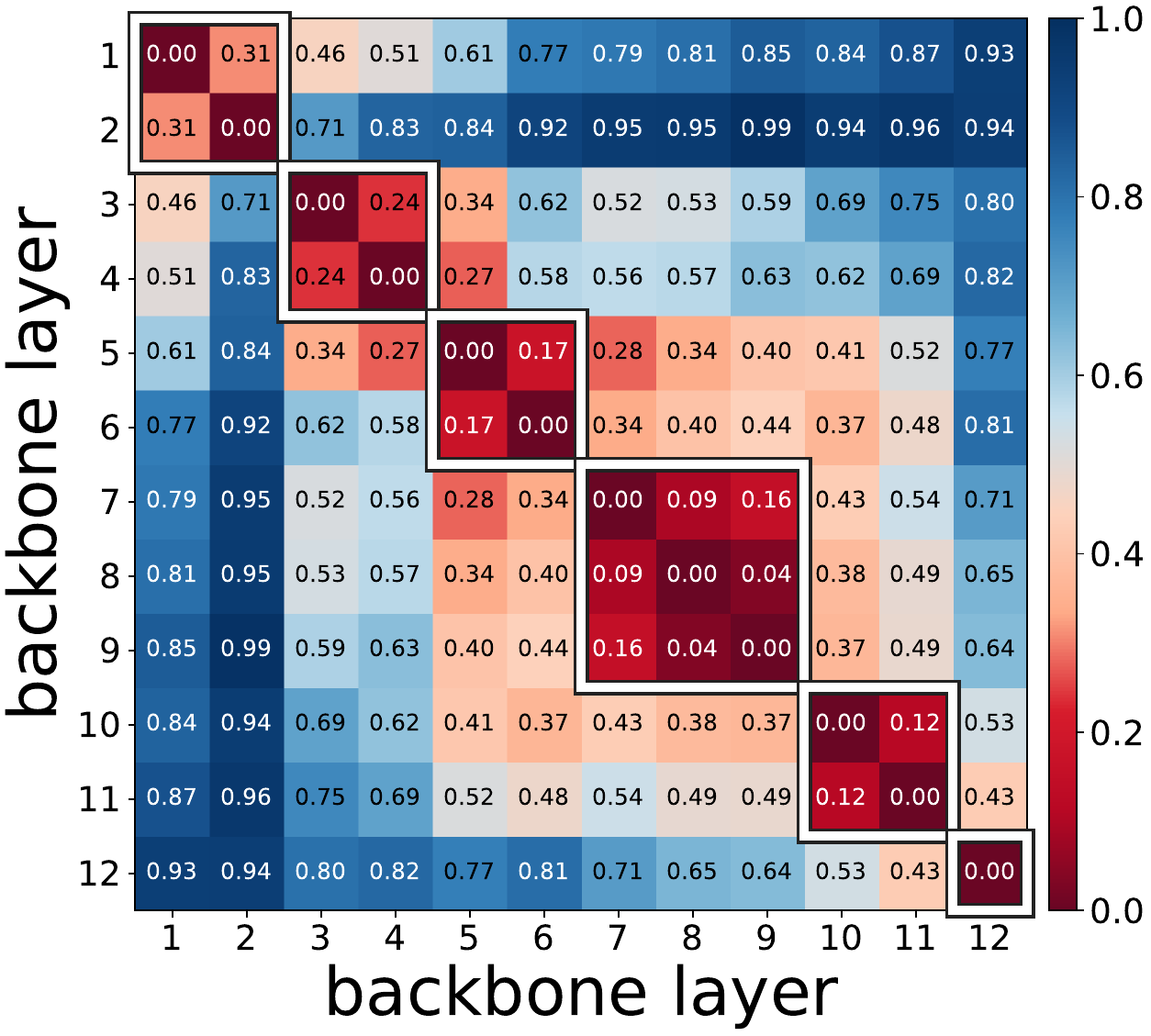}
        \caption{w/o Loop Adapters}
    \end{subfigure}
    \hfill
    \begin{subfigure}[t]{0.49\linewidth}
        \centering
        \includegraphics[width=\linewidth]{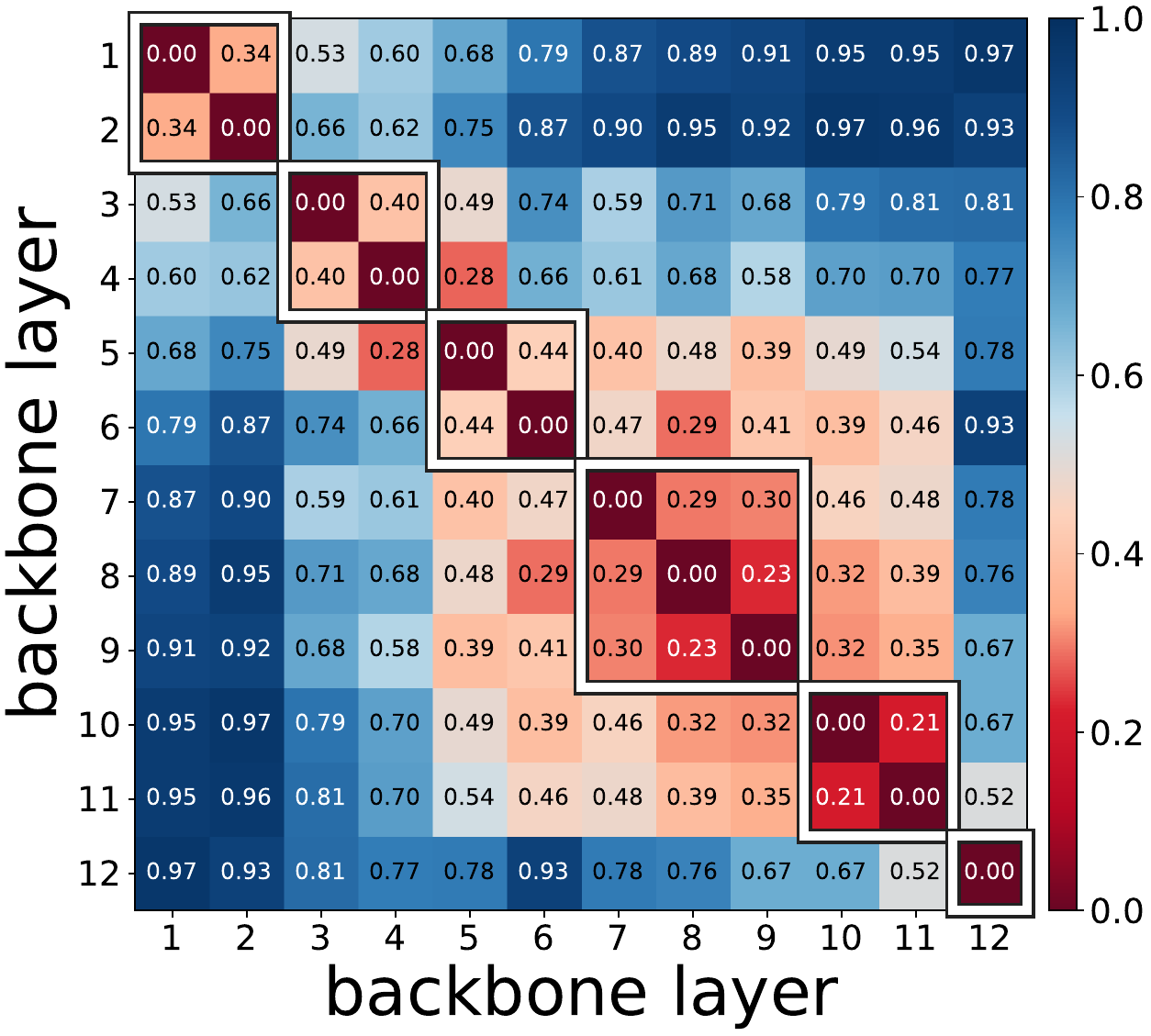}
        \caption{w/ Loop Adapters}
    \end{subfigure}
    \caption{\textbf{Effect of Loop Adapters on layer-wise representations.}
    High-distance regions indicate that pairs of layers learn distinct representations.
    Boxes indicate the selected groups.
}
    \label{fig:adapter_heatmap}
\end{figure}

Fig.~\ref{fig:adapter_heatmap} compares pairwise cosine distances among backbone layers under strict parameter sharing (a) and with the Loop Adapters (b).
The representation similarity between different layers that reuse the same Transformer block significantly decreases when Loop Adapters are used.
This observation indicates that Loop Adapters better preserve depth-specific transformations despite parameter sharing.

\paragraph{Ablation Study on Loop Adapter Design.}
Tab.~\ref{tab:adapter} examines the form and capacity of the Loop Adapter.
Non-linear adapters outperform the linear projection, while increasing the hidden dimension provides further gains with more parameters.
We adopt a two-layer MLP with hidden dimension $d_h=192$ as a trade-off between accuracy and model size.

\begin{table}[h]
\captionsetup{hypcap=false}
\caption{\textbf{Comparison of Loop Adapter designs.}
$d_h$ denotes the hidden dimension of the two-layer MLP in the non-linear adapter.}
\label{tab:adapter}
    \centering
    \small
    \begin{autotabular}{6}{l cccccc}
        \toprule
        Config. &
        AP &
        AP$_S$ &
        AP$_M$ &
        AP$_L$ &
        \#Params.&
        GFLOPs \\
        \hline
        w/o adapters & 49.4 & 30.0 & 53.8 & 68.8 & 5.8M & 50.0 \\
        Linear & 50.2 & 29.8 & 55.2 & 69.4 & 6.4M & 51.5 \\
        $d_h=96$ & 50.5 & 29.8 & 54.9 & 70.1 & 6.4M & 51.5 \\
        
        \rowcolor[gray]{0.9}
        \textbf{$d_h=192$ (ours)} & 50.9 & \textbf{31.4} & 55.9 & 69.8 & 7.0M & 53.0 \\
        $d_h=384$ & \textbf{51.7} & 30.4 & \textbf{56.9} & \textbf{71.3} & 8.1M & 56.0 \\
        \bottomrule
    \end{autotabular}
\end{table}

\vspace{0.5em}
\noindent\textbf{Ablation Study on Distillation Strategies.}
Since distillation strategies developed for other looped Transformers may not be optimal for dense visual recognition, we compare our approach with those of Raptor~\cite{jacobs2026block} and MiniViT~\cite{zhang2022minivit}.
Raptor aligns intermediate representations with the layer-wise feature trajectory of a ViT, while MiniViT applies supervision at the end of each group.
In contrast, our strategy supervises only the student's final-layer patch tokens.
We implement all strategies using the same LoopDet-S backbone and teacher.
As shown in Fig.~\ref{fig:distillation}, our strategy achieves the highest AP, outperforming Raptor and MiniViT by 6.4 and 5.9 AP.

\FloatBarrier

\section{Conclusion}
\label{sec:conclusion}

We presented LoopDVR, a pretrained-model-guided conversion framework that improves parameter efficiency by looping Transformer blocks in both the ViT backbone and query decoder without reducing computational depth.
To enable effective parameter sharing, it combines representation-based grouping, structured initialization, and Loop Adapters. 
LoopDVR achieves the best accuracy--parameter trade-offs across dense visual recognition tasks, substantially reducing parameters while maintaining competitive accuracy.

\ifarxiv
\section*{Acknowledgments}
We thank Yihao Zhen for valuable feedback and assistance in improving the manuscript, and Yunkang Tao for helpful comments on an earlier draft.
\fi

{\small
\bibliographystyle{ieeenat_fullname}
\bibliography{11_references}
}

\clearpage \appendix
\renewcommand{\dblfloatpagefraction}{0.5}
\setcounter{figure}{0}
\renewcommand{\thefigure}{S\arabic{figure}}
\renewcommand{\theHfigure}{supp.\arabic{figure}}
\setcounter{table}{0}
\renewcommand{\thetable}{S\arabic{table}}
\renewcommand{\theHtable}{supp.\arabic{table}}
\setcounter{algorithm}{0}
\renewcommand{\thealgorithm}{S\arabic{algorithm}}
\def\theHalgorithm{supp.\arabic{algorithm}}

\label{sec:appendix}

\section{Implementation Details}

\subsection{Architecture}
\label{sec:Architecture Configurations}

Algorithm~\ref{alg:loopdvr_forward} illustrates the forward pass of the looped backbone.
Each layer executes the Transformer block assigned to its group, followed by a depth-specific Loop Adapter.
The block first transforms the incoming tokens. 
The adapter then combines the block output with an MLP of its normalized tokens, using the learnable coefficients in Eq.~\ref{eq:adapter_form}.
The adapted tokens become the input to the next layer. The final two layer outputs supply the patch features for the multi-scale interface.
The decoder applies its shared block and Loop Adapter in the same order at each layer.

\begin{algorithm}[h]
\caption{Forward pass of the looped backbone}
\label{alg:loopdvr_forward}
\begin{lstlisting}[style=pseudocode]
# x: input patch tokens
# blocks: shared Transformer blocks
# group_sizes: number of layers per group
# adapters: layer-specific Loop Adapters

def forward(x, blocks, group_sizes, adapters):
    outputs = []
    i = 0

    # Apply shared blocks with adapters
    for block, size in zip(blocks, group_sizes):
        for _ in range(size):
            h = block(x)
            a = adapters[i]
            r = a.mlp(a.norm(h))
            x = a.alpha * h + a.beta * r
            outputs.append(x)
            i += 1

    # Return the penultimate and final outputs
    return outputs[-2], outputs[-1]
\end{lstlisting}
\end{algorithm}

\paragraph{Model Configurations.} We instantiate LoopDVR at four model scales based on the corresponding EdgeCrafter configurations~\cite{liu2026edgecrafter}. Table~\ref{tab:loopdet_config} summarizes the default architecture configurations of our models, including the feature dimensions, computational depths, shared blocks budgets, and the Loop Adapters settings. The models inherit the backbone, encoder and decoder widths of the reference architecture, while reducing the number of independently parameterized Transformer blocks through group-wise sharing.

\begin{table}[t]
    \centering
    \small
    \captionsetup{hypcap=false}
    \caption{\textbf{Architecture configurations of LoopDVR variants.}
    The number of layers denotes how many times blocks are executed, while the number of shared blocks denotes how many distinct parameterized blocks are used.
    }
    \label{tab:loopdet_config}
    \begin{autotabular}{5}{lc ccc}
    \toprule
    Setting & S & M & L & X \\
    \midrule
    \multicolumn{5}{l}{\textit{\textbf{Backbone}}} \\
    Embedding dimension & 192 & 256 & 384 & 384 \\
    Attention heads & 3 & 4 & 6 & 6 \\
    FFN expansion ratio & 4 & 4 & 4 & 6 \\
    Backbone layers & 12 & 12 & 12 & 12 \\
    Shared Transformer blocks & 6 & 6 & 6 & 6 \\
    Loop Adapters hidden dimension & 192 & 256 & 384 & 384 \\
    \midrule
    \multicolumn{5}{l}{\textit{\textbf{Hybrid Encoder}}} \\
    hidden dimension & 192 & 256 & 256 & 256 \\
    FFN dimension & 512 & 512 & 1024 & 2048 \\
    \midrule
    \multicolumn{5}{l}{\textit{\textbf{LoopDet \& LoopSeg Decoder}}} \\
    hidden dimension & 192 & 256 & 256 & 256 \\
    FFN dimension & 512 & 1024 & 1024 & 2048 \\
    Object queries & 300 & 300 & 300 & 300 \\
    Decoder layers & 4 & 4 & 4 & 4 \\
    Shared Transformer blocks & 2 & 2 & 2 & 2 \\
    Loop Adapters hidden dimension & 192 & 256 & 256 & 256 \\
    \midrule
    \multicolumn{5}{l}{\textit{\textbf{LoopPose Decoder}}} \\
    hidden dimension & 192 & 256 & 256 & 256 \\
    FFN dimension & 512 & 512 & 1024 & 2048 \\
    Object queries & 60 & 60 & 60 & 60 \\
    Decoder layers & 4 & 4 & 4 & 4 \\
    Shared Transformer blocks & 2 & 2 & 2 & 2 \\
    Loop Adapter hidden dim. & 192 & 256 & 256 & 256 \\
    \bottomrule
    \end{autotabular}
\end{table}

\paragraph{Loop Adapters. }
Each adapter provides a depth-specific transformation while preserving the feature dimension of its host module. Following the nonlinear projection design of SReT~\cite{shen2022sliced}, it applies LayerNorm followed by two linear layers with a GELU activation between them. Adapter parameters are independent for each reuse of the shared block.

\paragraph{Multi-scale Feature Interface. }
The looped backbone uses a feature interface for integration with the hybrid encoder. We average the patch-token features from the final two layers and reshape them into a spatial feature map, excluding the non-spatial token. Bilinear interpolation and the reference channel projections construct the feature pyramid at different scales. The hybrid encoder fuses these features to provide the multi-scale memory consumed by the looped decoder.

\begin{figure*}[t]
    \centering
    \captionsetup[subfigure]{font=footnotesize,skip=2pt}

    \begin{subfigure}[t]{0.24\textwidth}
        \centering
        \includegraphics[width=\linewidth]{figs/backbone.pdf}
        \caption{ECViT-T (for -S models)}
    \end{subfigure}\hfill
    \begin{subfigure}[t]{0.24\textwidth}
        \centering
        \includegraphics[width=\linewidth]{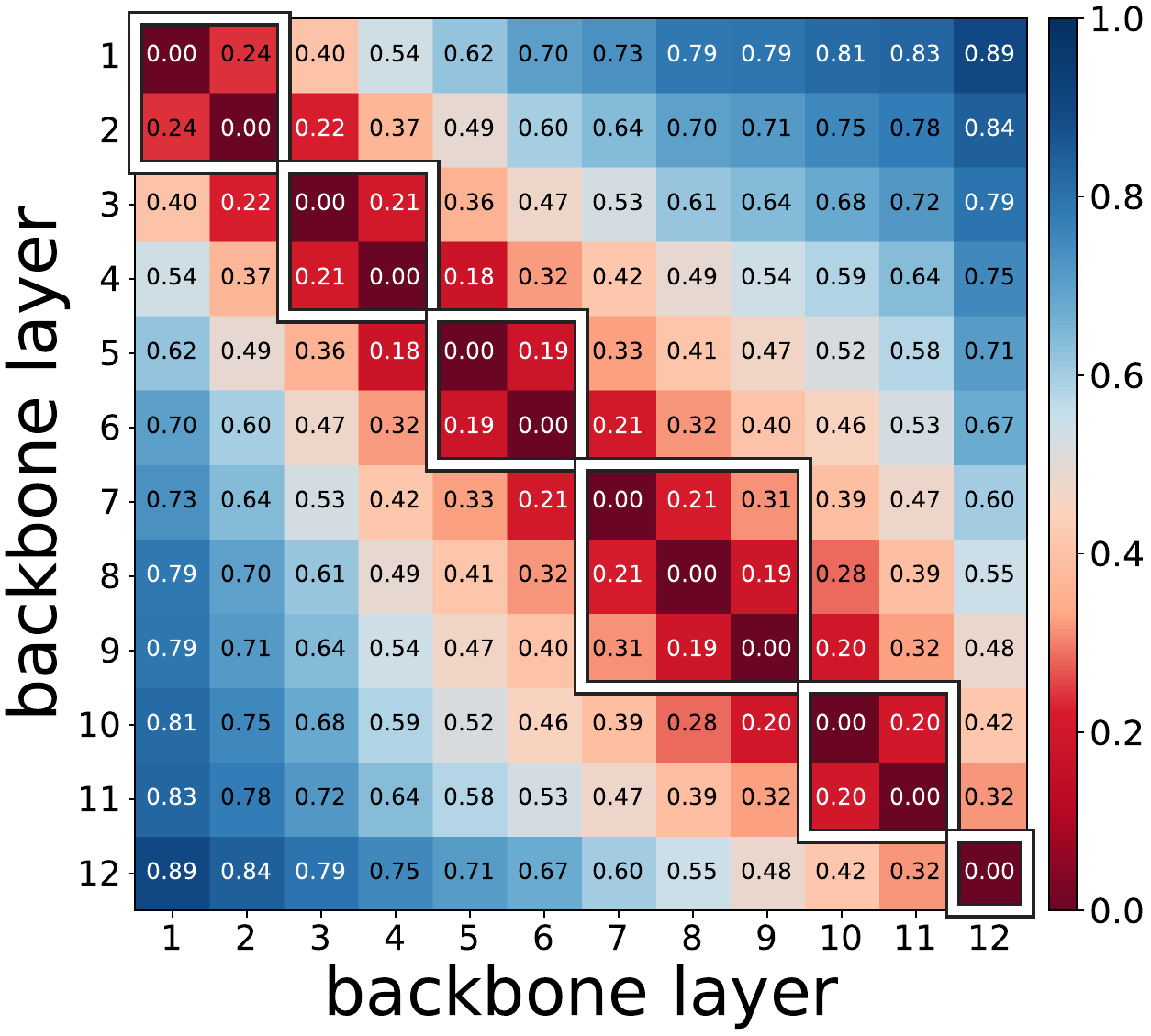}
        \caption{ECViT-T+ (for -M models)}
    \end{subfigure}\hfill
    \begin{subfigure}[t]{0.24\textwidth}
        \centering
        \includegraphics[width=\linewidth]{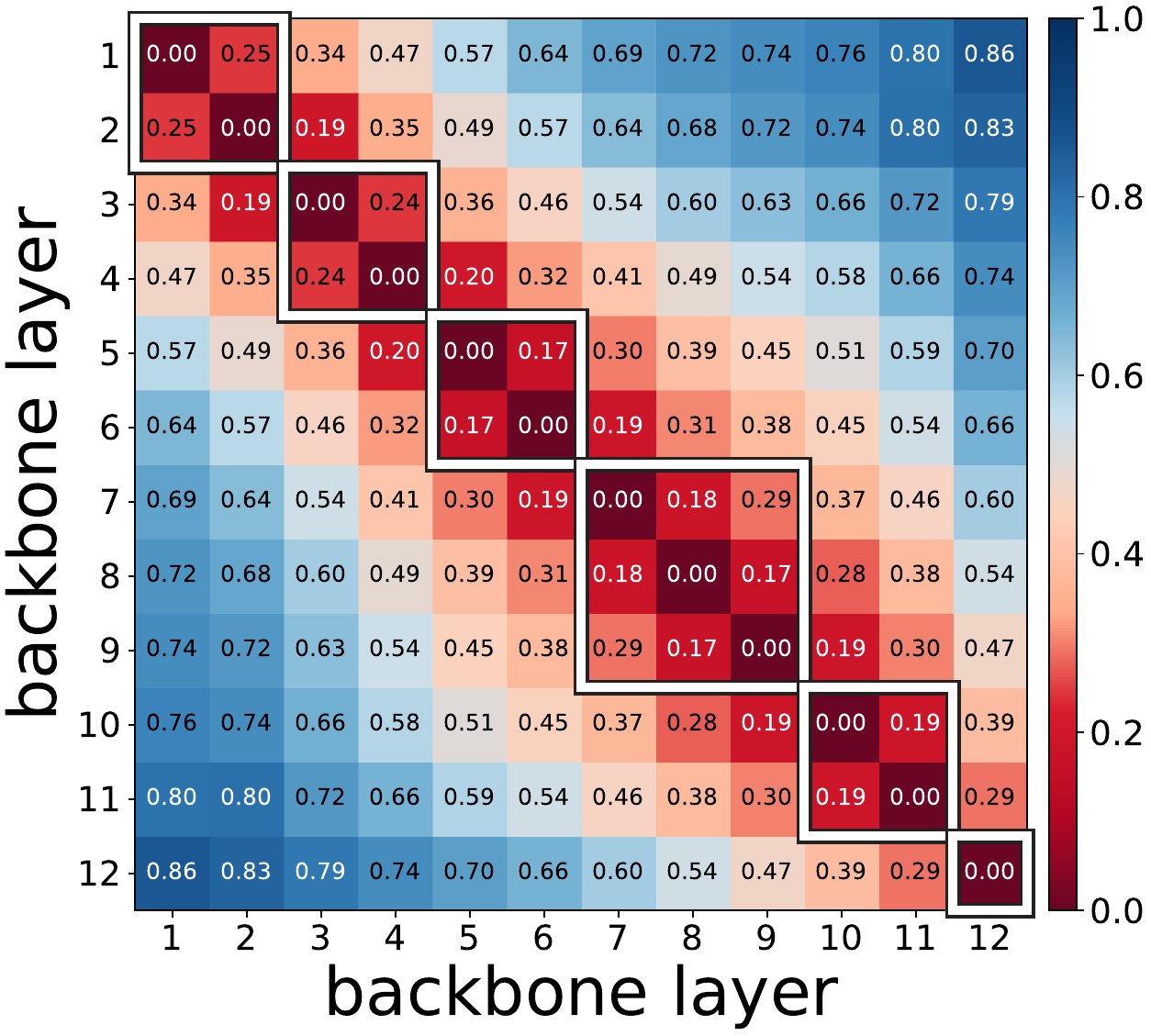}
        \caption{ECViT-S (for -L models)}
    \end{subfigure}\hfill
    \begin{subfigure}[t]{0.24\textwidth}
        \centering
        \includegraphics[width=\linewidth]{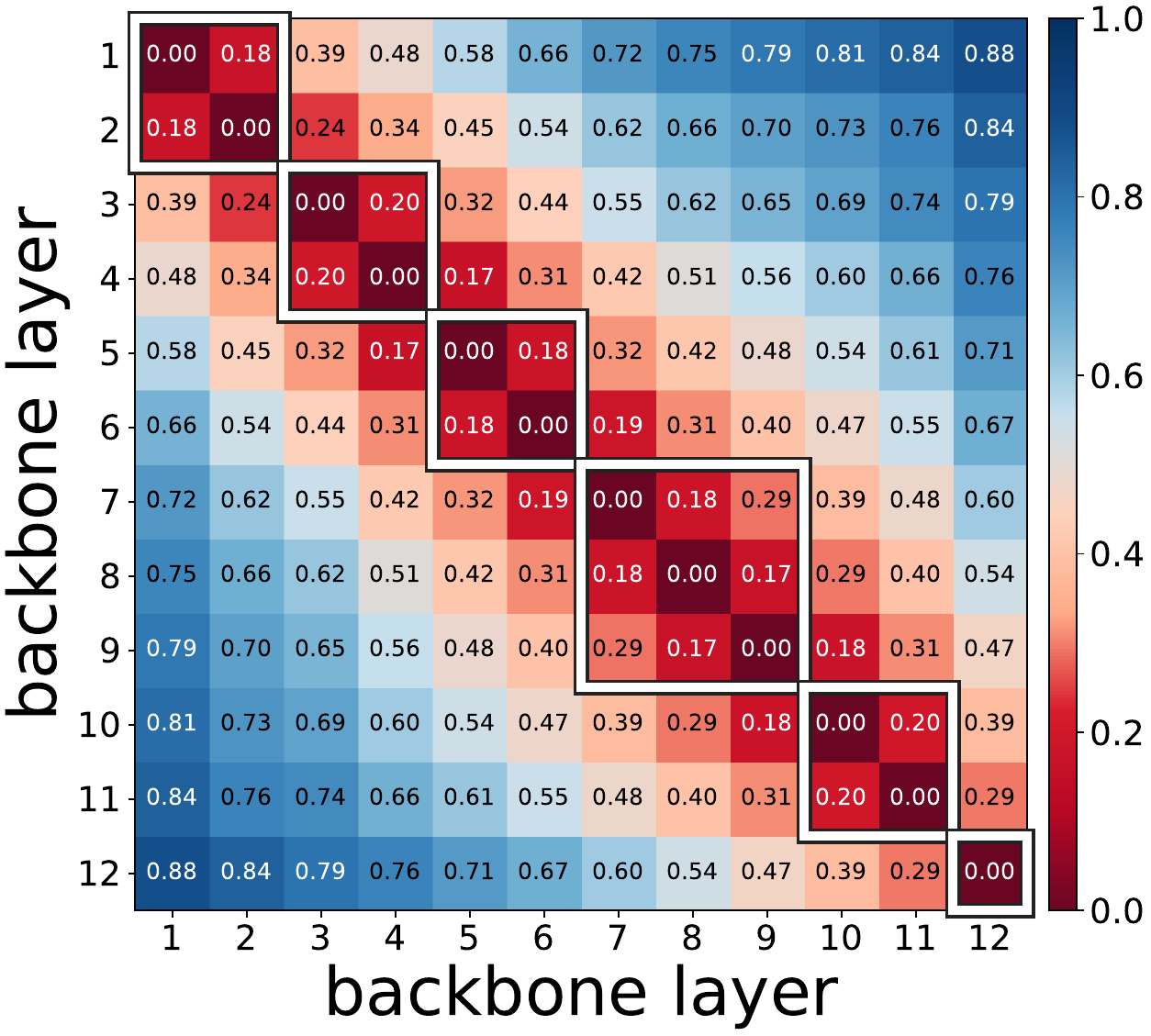}
        \caption{ECViT-S+ (for -X models)}
    \end{subfigure}

    \par\medskip
    \vspace{-0.5em}

    \begin{subfigure}[t]{0.24\textwidth}
        \centering
        \includegraphics[width=\linewidth]{figs/decoder.pdf}
        \caption{ECDet-S Decoder}
    \end{subfigure}\hfill
    \begin{subfigure}[t]{0.24\textwidth}
        \centering
        \includegraphics[width=\linewidth]{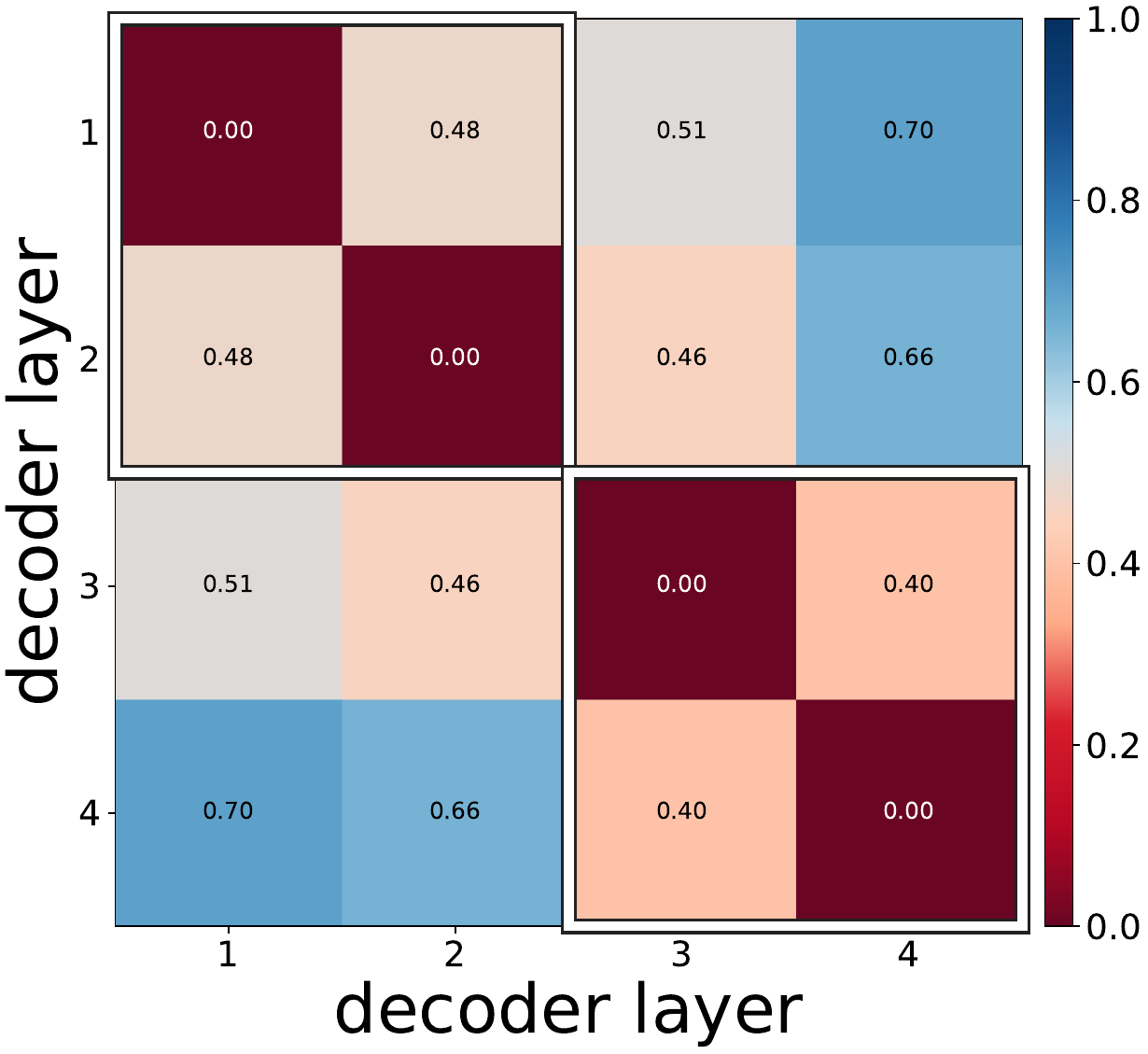}
        \caption{ECDet-M Decoder}
    \end{subfigure}\hfill
    \begin{subfigure}[t]{0.24\textwidth}
        \centering
        \includegraphics[width=\linewidth]{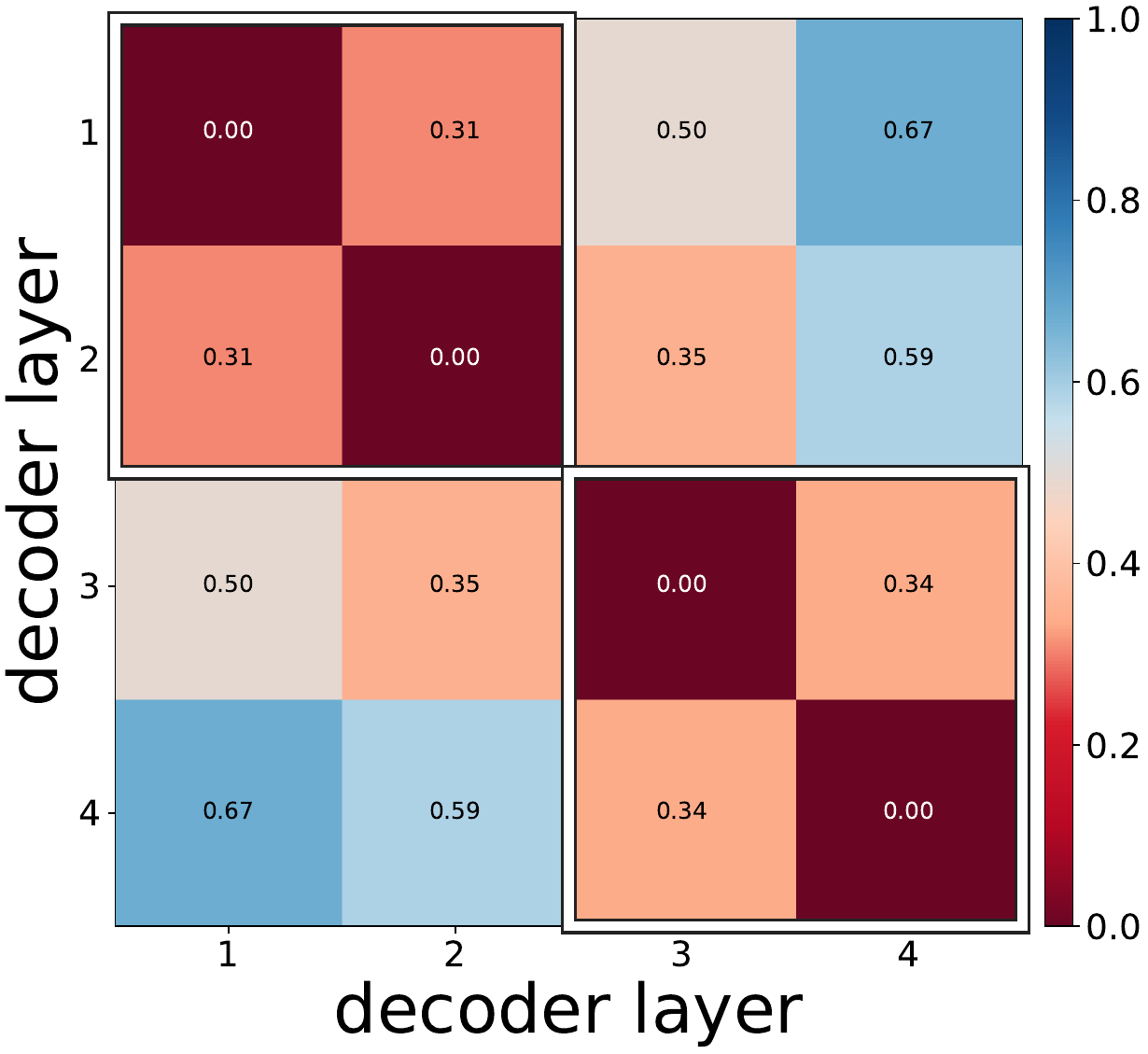}
        \caption{ECDet-L Decoder}
    \end{subfigure}\hfill
    \begin{subfigure}[t]{0.24\textwidth}
        \centering
        \includegraphics[width=\linewidth]{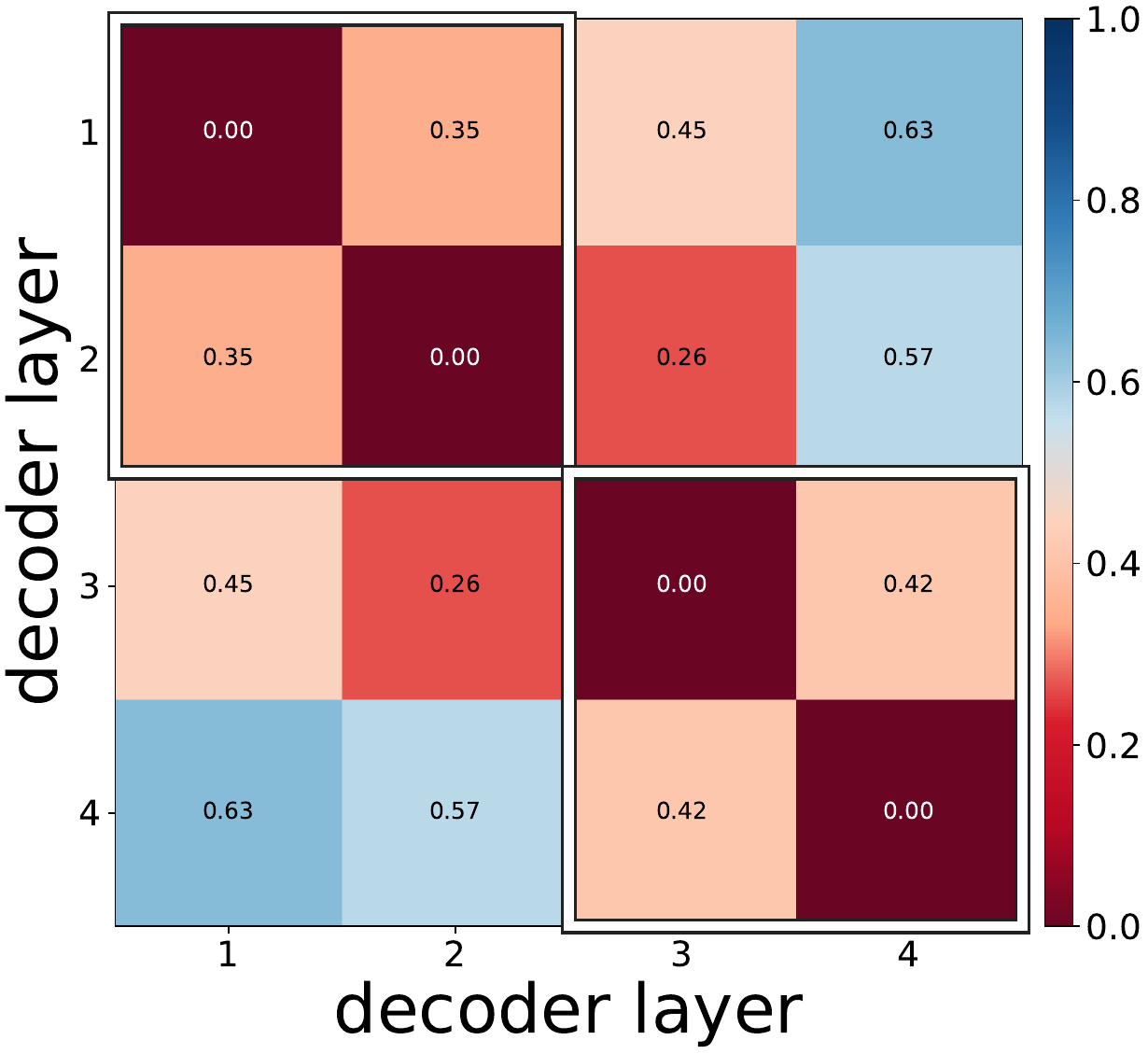}
        \caption{ECDet-X Decoder}
    \end{subfigure}

    \par\medskip
    \vspace{-0.5em}

    \begin{subfigure}[t]{0.24\textwidth}
        \centering
        \includegraphics[width=\linewidth]{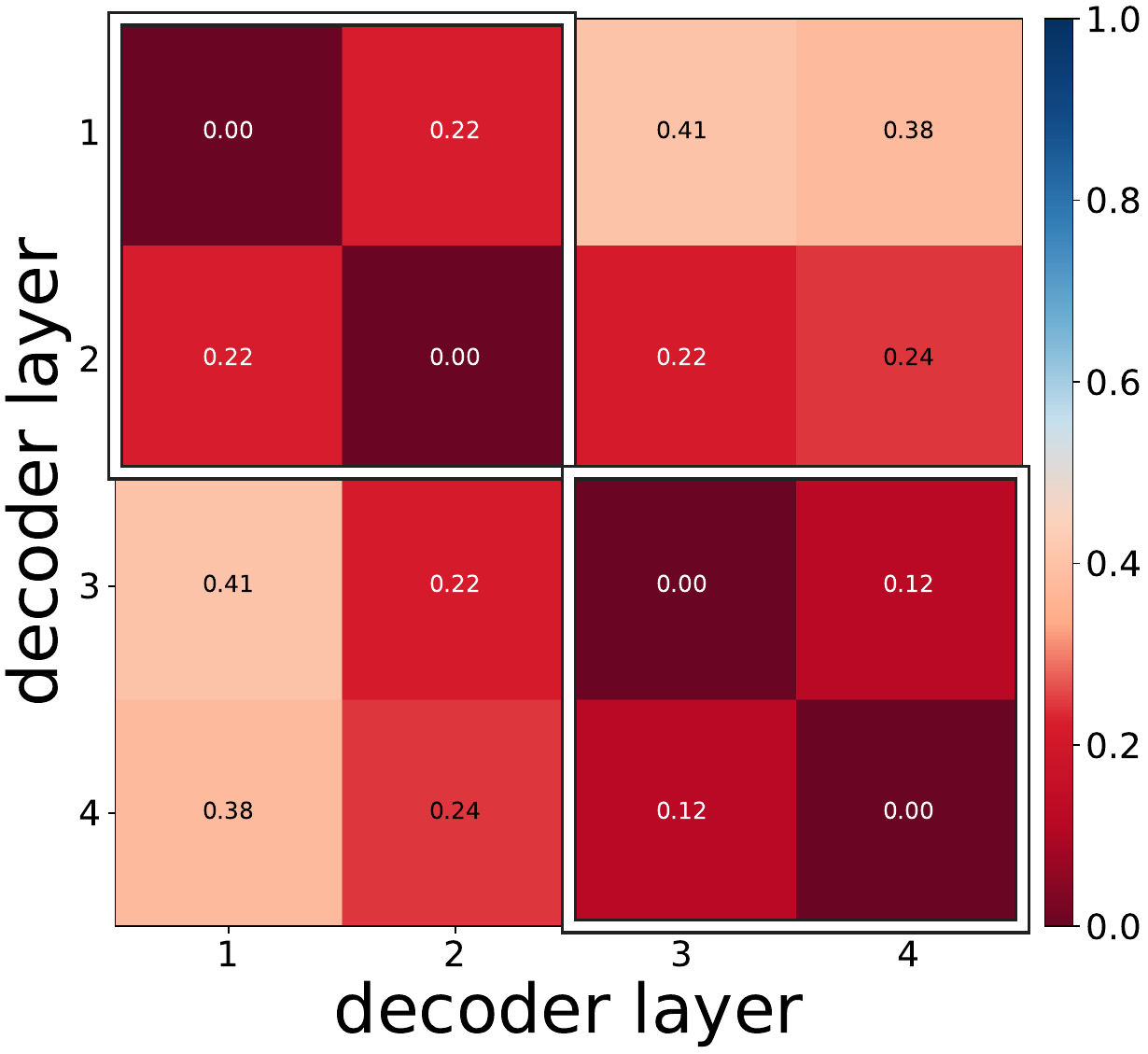}
        \caption{ECInsSeg-S Decoder}
    \end{subfigure}\hfill
    \begin{subfigure}[t]{0.24\textwidth}
        \centering
        \includegraphics[width=\linewidth]{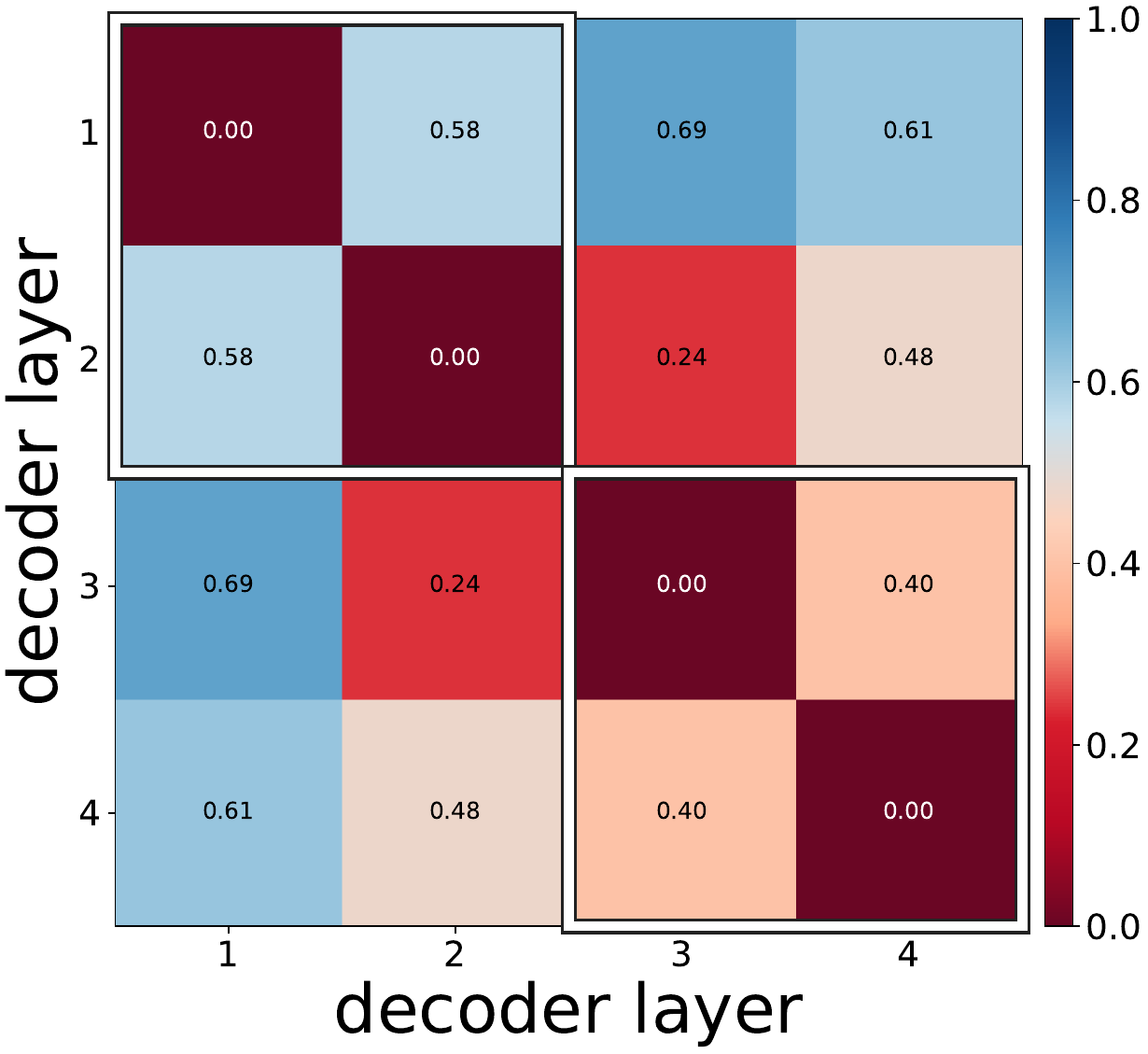}
        \caption{ECInsSeg-M Decoder}
    \end{subfigure}\hfill
    \begin{subfigure}[t]{0.24\textwidth}
        \centering
        \includegraphics[width=\linewidth]{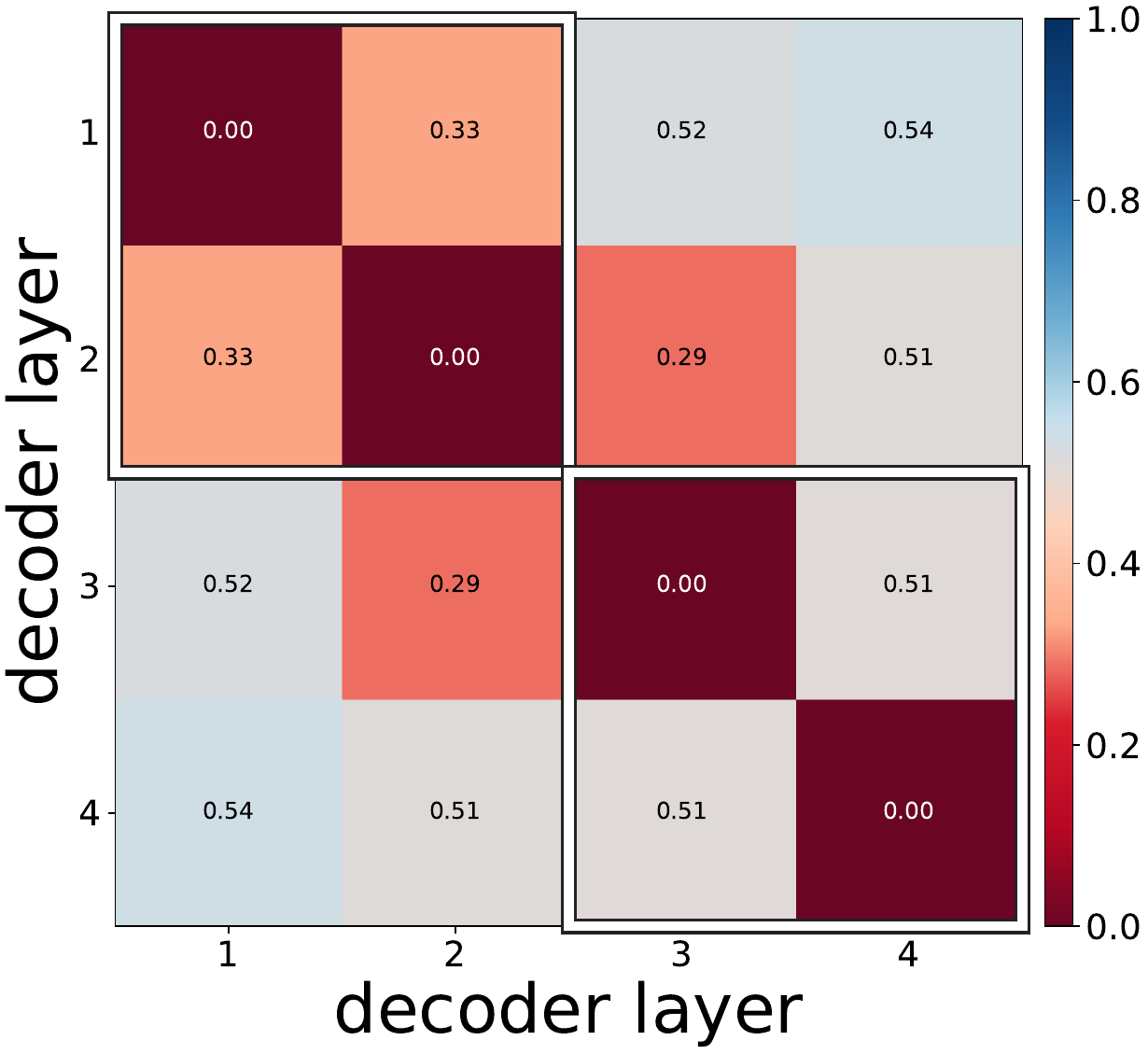}
        \caption{ECInsSeg-L Decoder}
    \end{subfigure}\hfill
    \begin{subfigure}[t]{0.24\textwidth}
        \centering
        \includegraphics[width=\linewidth]{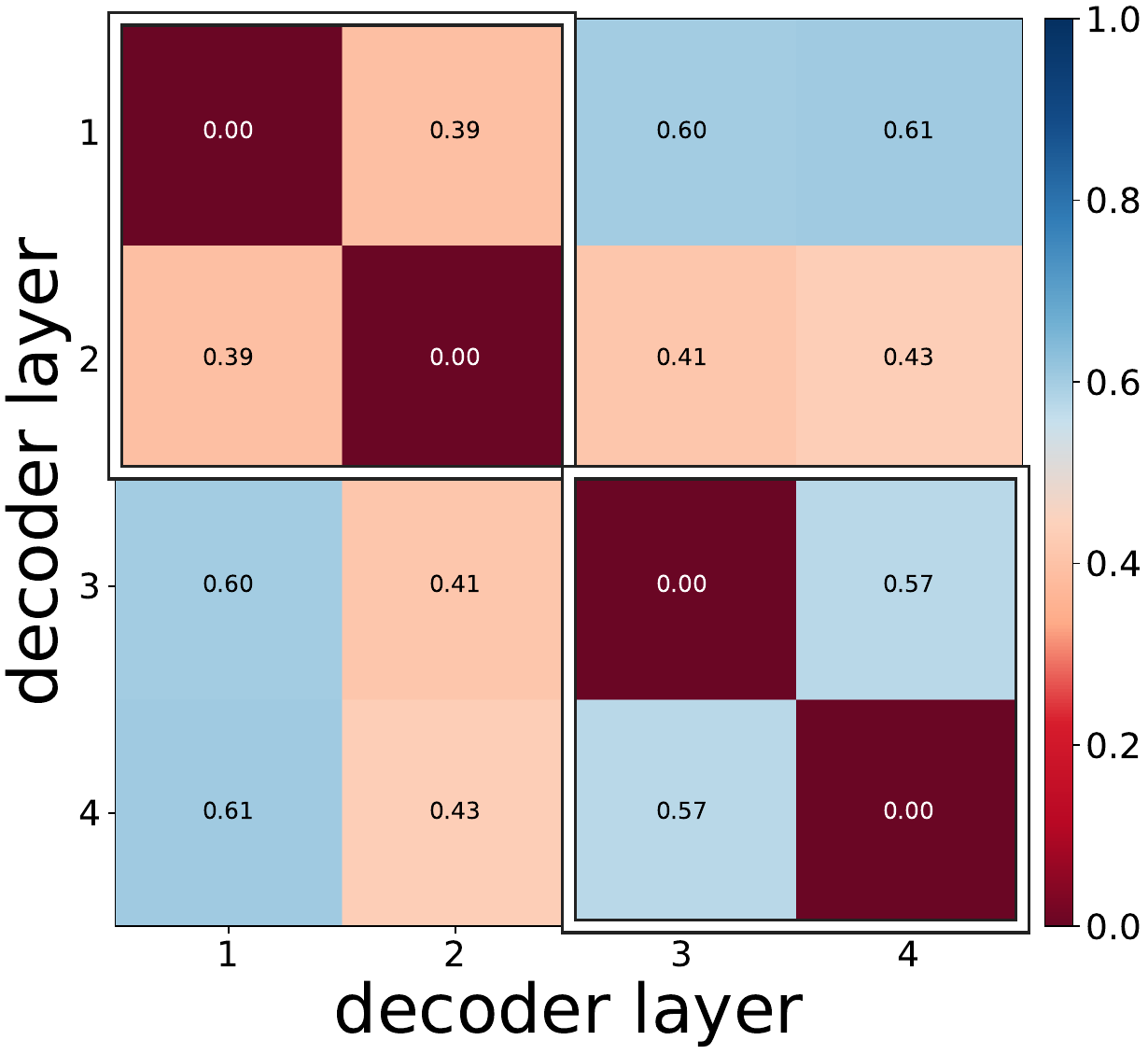}
        \caption{ECInsSeg-X Decoder}
    \end{subfigure}

    \par\medskip
    \vspace{-0.5em}

    \begin{subfigure}[t]{0.24\textwidth}
        \centering
        \includegraphics[width=\linewidth]{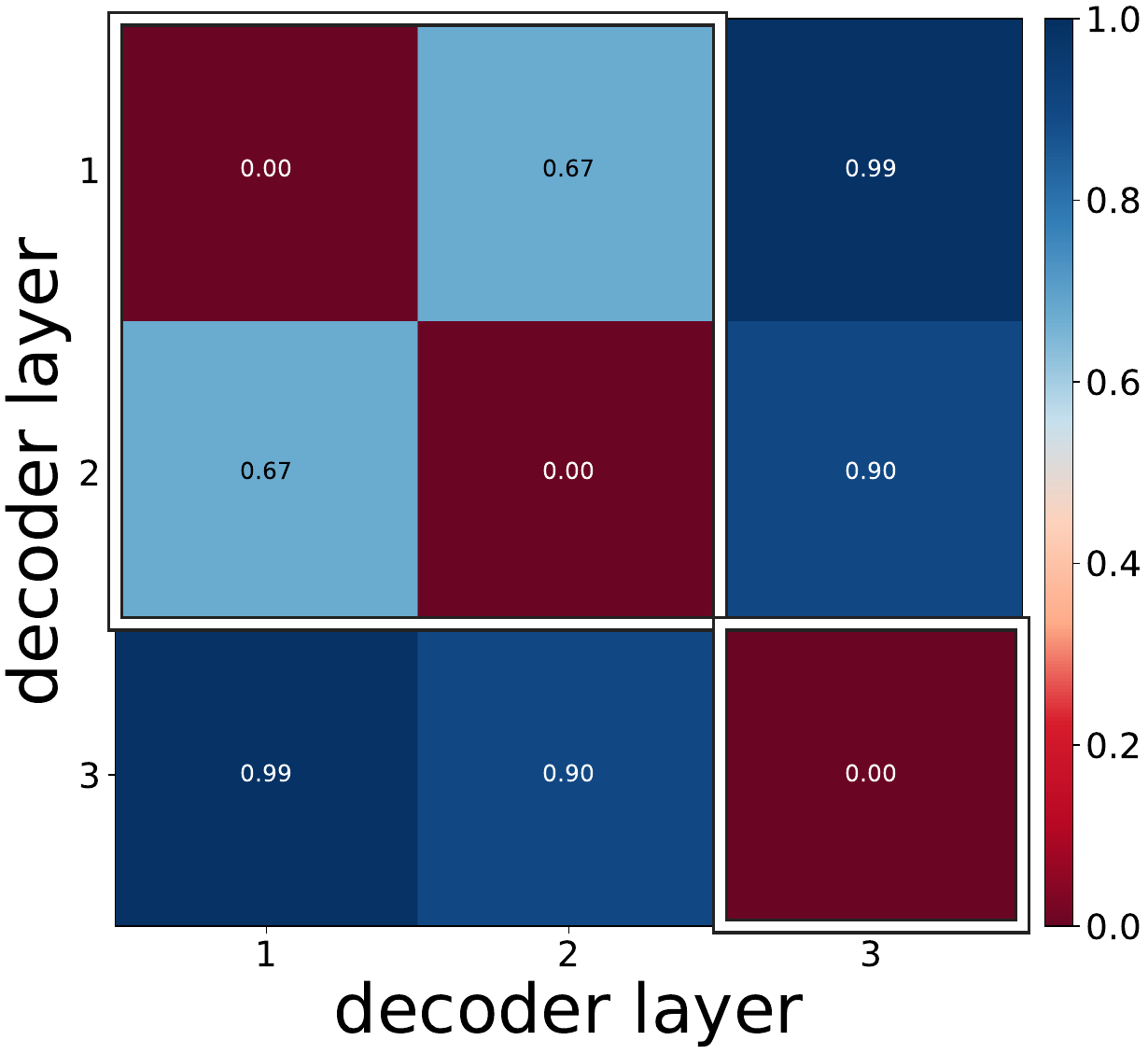}
        \caption{ECPose-S Decoder}
    \end{subfigure}\hfill
    \begin{subfigure}[t]{0.24\textwidth}
        \centering
        \includegraphics[width=\linewidth]{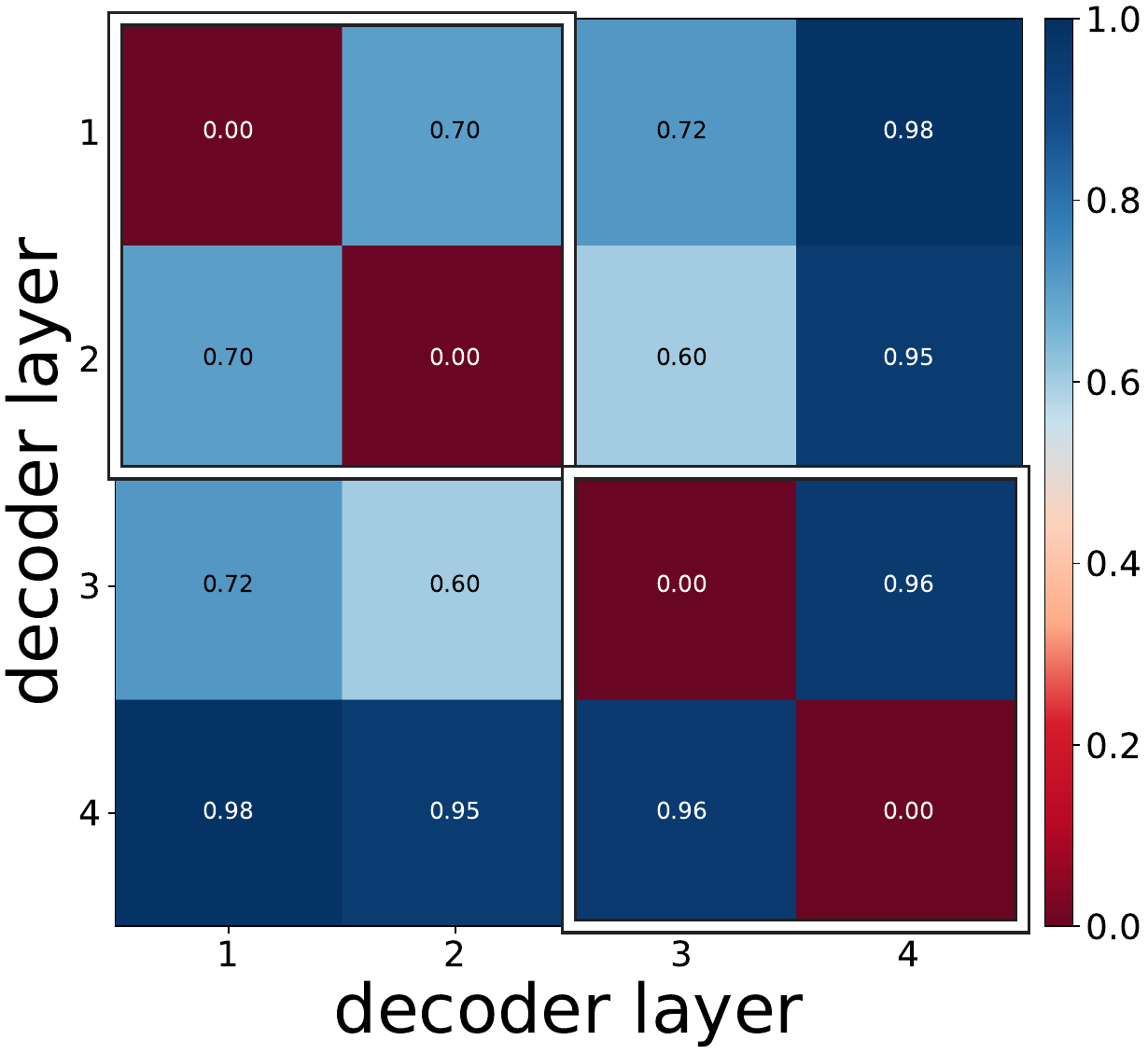}
        \caption{ECPose-M Decoder}
    \end{subfigure}\hfill
    \begin{subfigure}[t]{0.24\textwidth}
        \centering
        \includegraphics[width=\linewidth]{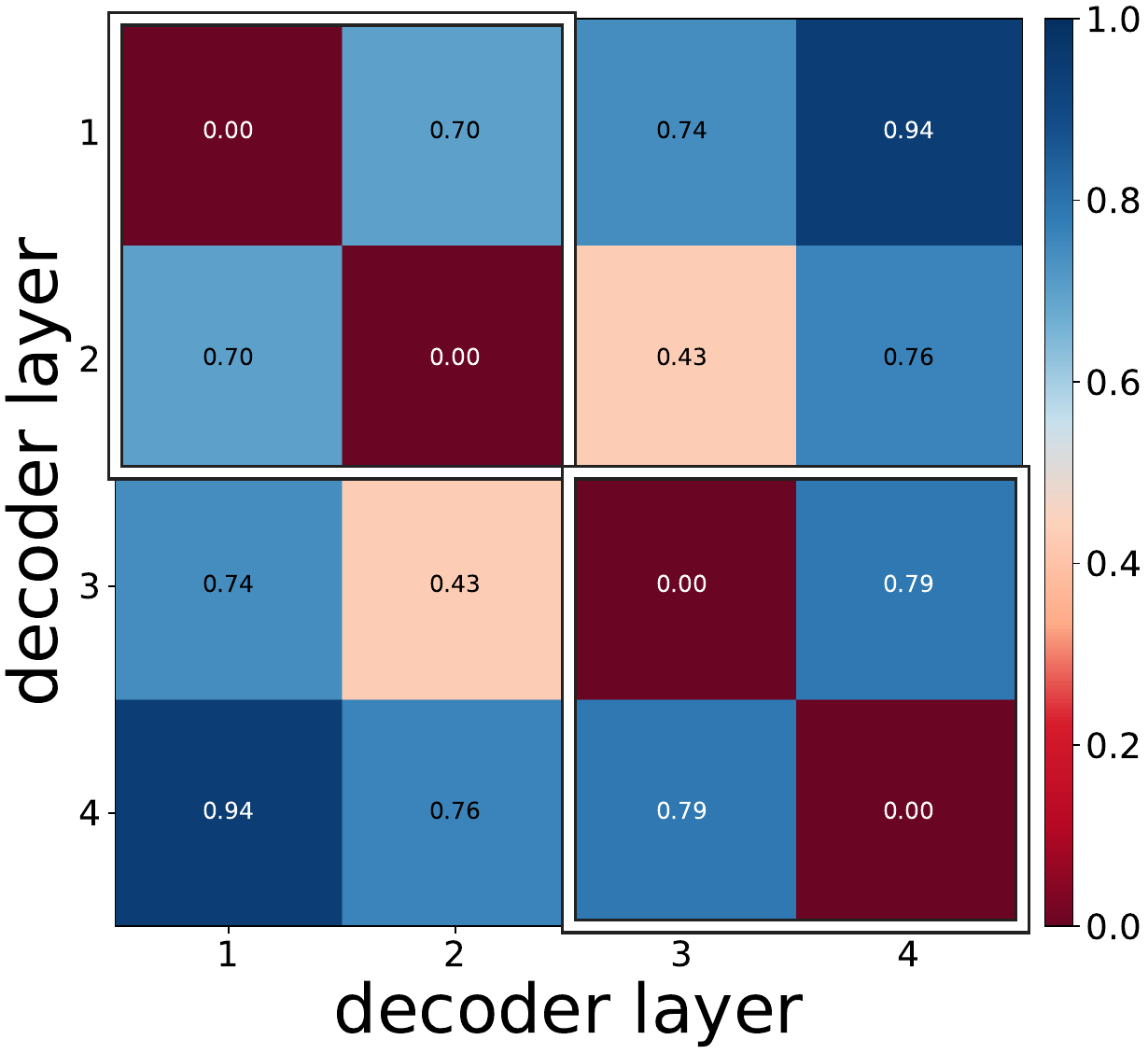}
        \caption{ECPose-L Decoder}
    \end{subfigure}\hfill
    \begin{subfigure}[t]{0.24\textwidth}
        \centering
        \includegraphics[width=\linewidth]{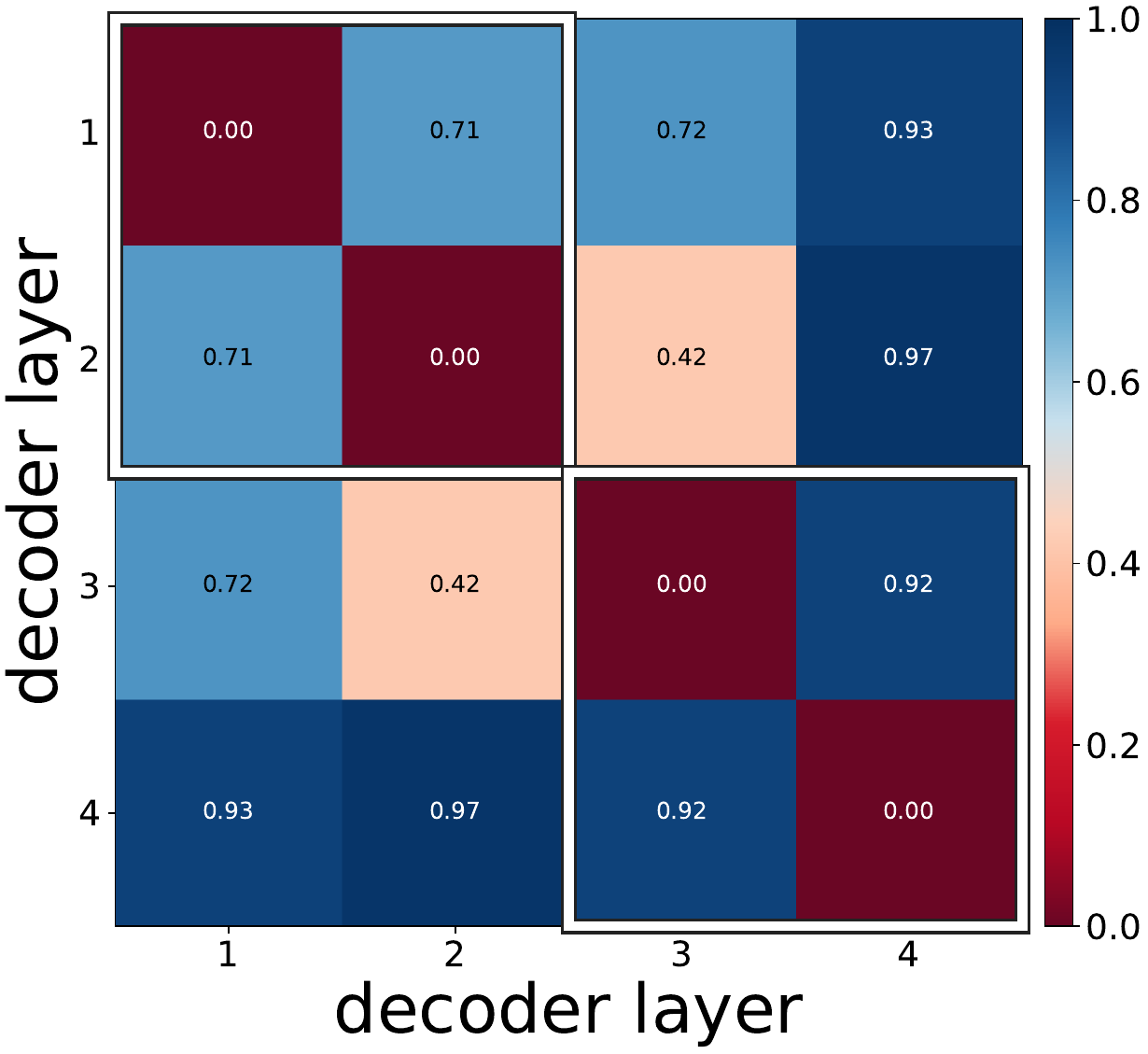}
        \caption{ECPose-X Decoder}
    \end{subfigure}
    \vspace{-0.5em}

    \caption{
    \textbf{Layer-wise representation distances and grouping across reference EdgeCrafter~\cite{liu2026edgecrafter} models.}
    Each entry reports the cosine distance of representations between layers , with smaller values indicating greater similarity. 
    Boxes mark the selected groups.
    }
    \label{fig:supp_phase_partitions}
\end{figure*}

\subsection{Grouping Across Model Variants}

Figure~\ref{fig:supp_phase_partitions} visualizes the layer-wise representation distances and sharing groups of the reference backbones and task-specific decoders.
All four backbone variants contain twelve layers partitioned into six groups, following the same group-size sequence $(2,2,2,3,2,1)$, with the final layer forming an independent group.
The decoder layers are partitioned into two contiguous sharing groups.

Algorithm~\ref{alg:loopdvr_groups} describes the grouping procedure.
For the backbone, we average the patch tokens at each layer within an image and compute pairwise cosine distances averaged across images to obtain $D_B$ (Eq.~\ref{eq:activation_distance}).
For the decoder, we compute $D_D$ from layer-wise query representations.
Given a group spanning layers $[a,b)$, its cost is defined as
$W(D,a,b)=\sum_{a\leq i<j<b}D_{ij}$.
We use dynamic programming to minimize the total within-group distance, combining the cost of each candidate final group with the optimal partition of the preceding layers.
The final backbone layer is reserved as an independent group (Eq.~\ref{eq:contiguous_partition}), whereas all decoder layers are partitioned jointly.
The resulting group sizes determine the number of executions of each shared block in Algorithm~\ref{alg:loopdvr_forward}.

\begin{algorithm}[t]
\caption{Grouping of backbone and decoder layers}
\label{alg:loopdvr_groups}
\begin{lstlisting}[style=pseudocode]
# D: pairwise representation distance matrix
# K: number of sharing groups

def W(D, a, b):
    return sum(D[i, j] for i in range(a, b)
                       for j in range(i + 1, b))

def partition(D, K):
    n = len(D)
    best = {}

    # Initialize single-group partitions
    for b in range(1, n + 1):
        best[1, b] = (W(D, 0, b), [b])

    # Find optimal contiguous partitions
    for k in range(2, K + 1):
        for b in range(k, n + 1):
            choices = []
            for a in range(k - 1, b):
                cost, sizes = best[k - 1, a]
                cost += W(D, a, b)
                new_sizes = sizes + [b - a]
                choices.append((cost, new_sizes))

            best[k, b] = min(
                choices, key=lambda x: x[0]
            )

    return best[K, n][1]

# Reserve the final backbone layer
sizes_B = partition(D_B[:-1, :-1], K_B - 1) + [1]

# Partition all decoder layers
sizes_D = partition(D_D, K_D)
\end{lstlisting}
\end{algorithm}

\subsection{Initialization.} 

\paragraph{Shared Backbone.}
We initialize the shared backbone from the corresponding pretrained ECViT checkpoint. Following the group-end rule in Eq.~\ref{eq:phase_end_initialization}, each shared block receives the structurally compatible parameters of the terminal reference layer in its group. The transferred parameters include the query, key, and value projections, the attention output projection, the feed-forward layers, and the block normalization parameters. The compatible patch-embedding parameters, including the convolutional stem, are also transferred from the reference checkpoint.

\paragraph{Loop Adapters.}
Each Loop Adapter has independent parameters and takes the form
\begin{equation}
    \mathcal{A}_l(\mathbf{x})
    =
    \alpha_l \mathbf{x}
    +
    \beta_l \operatorname{MLP}_l
    \bigl(\operatorname{LN}_l(\mathbf{x})\bigr),
    \label{eq:supp_adapter_initialization}
\end{equation}
where $\alpha_l$ and $\beta_l$ are learnable scalar coefficients.
The MLP parameters are randomly initialized, while the coefficients are initialized to $\alpha_l=1$ and $\beta_l=0$. Consequently, all adapters initially implement identity mapping, $\mathcal{A}_l(\mathbf{x})=\mathbf{x}$.

\subsection{Training Configurations}
\label{sec:training_config}

We train LoopDVR through backbone distillation followed by downstream task-specific training, on 4$\times$ NVIDIA RTX 4090
GPUs. The hyperparameters of downstream task-specific training are summarized in Table~\ref{tab:hyperparameters}.

\paragraph{Teacher configuration.} 
Following EdgeCrafter~\cite{liu2026edgecrafter}, we use the detection-specialized DINOv3~\cite{simeoni2025dinov3} models for backbone distillation. All teachers are obtained by fine-tuning DINOv3 on COCO~\texttt{train2017}~\cite{lin2014microsoft}. 
We used fine-tuned DINOv3-S to supervise the S-scale backbone, while DINOv3-B supervises the M, L, and X backbones. Each teacher remains frozen and operates in evaluation mode throughout distillation.

\paragraph{Backbone distillation. }
We distill the looped backbones on unlabeled images comprising ImageNet-1K~\cite{imagenet15russakovsky} and COCO train2017~\cite{lin2014microsoft}. The teacher and student receive the same augmented images, with random resized cropping, horizontal flipping, color jitter, grayscale conversion, Gaussian blur, and Mixup~\cite{zhang2018mixup}. 
We apply a trainable linear projector to each final-layer student patch token and use an MSE loss to align it with the concatenated features of the corresponding patch token from the teacher’s last two layers. 
The shared backbone blocks, Loop Adapters, and the projector are jointly optimized using layer-wise adaptive rate scaling (LARS)~\cite{LARS}, with linear learning rate warm-up followed by cosine decay.

\begin{table*}[t]
    \centering
    \small
    \captionsetup{hypcap=false}
    \caption{\textbf{Hyperparameters for downstream task-specific training.}}
    \label{tab:hyperparameters}
    \begin{autotabular}{13}{l | cccc | cccc | cccc}
    \toprule
    \multirow{2}{*}{Setting}
        & \multicolumn{4}{c|}{LoopDet}
        & \multicolumn{4}{c|}{LoopSeg}
        & \multicolumn{4}{c}{LoopPose} \\
    \cmidrule(lr){2-5}
    \cmidrule(lr){6-9}
    \cmidrule(lr){10-13}
        & S & M & L & X
        & S & M & L & X
        & S & M & L & X \\
    \midrule
    Optimizer
        & \multicolumn{4}{c|}{AdamW}
        & \multicolumn{4}{c|}{AdamW}
        & \multicolumn{4}{c}{AdamW} \\
    Backbone LR
        & $2.5\mathrm{e}{-5}$ & $1.25\mathrm{e}{-5}$
        & $5\mathrm{e}{-6}$ & $5\mathrm{e}{-6}$
        & $2.5\mathrm{e}{-5}$ & $2.5\mathrm{e}{-5}$
        & $2.5\mathrm{e}{-6}$ & $2.5\mathrm{e}{-6}$
        & $2.5\mathrm{e}{-5}$ & $2.5\mathrm{e}{-5}$
        & $2.5\mathrm{e}{-6}$ & $2.5\mathrm{e}{-6}$ \\
    Base LR
        & $5\mathrm{e}{-4}$ & $5\mathrm{e}{-4}$
        & $5\mathrm{e}{-4}$ & $5\mathrm{e}{-4}$
        & $5\mathrm{e}{-4}$ & $5\mathrm{e}{-4}$
        & $5\mathrm{e}{-4}$ & $5\mathrm{e}{-4}$
        & $1\mathrm{e}{-3}$ & $5\mathrm{e}{-4}$
        & $5\mathrm{e}{-4}$ & $5\mathrm{e}{-4}$ \\
    LR decay
        & 0.5 & 0.5 & 0.5 & 0.5 
        & 0.5 & 0.5 & 0.5 & 0.5 
        & 0.5 & -- & -- & -- \\
    Weight decay
        & $1\mathrm{e}{-4}$ & $1\mathrm{e}{-4}$
        & $1.25\mathrm{e}{-4}$ & $1.25\mathrm{e}{-4}$
        & $1\mathrm{e}{-4}$ & $1\mathrm{e}{-4}$
        & $1.25\mathrm{e}{-4}$ & $1.25\mathrm{e}{-4}$
        & $1\mathrm{e}{-4}$ & $1\mathrm{e}{-4}$
        & $1.25\mathrm{e}{-4}$ & $1.25\mathrm{e}{-4}$ \\
    \midrule
    Epochs
        & $74$ & $62$ & $50$ & $50$
        & $74$ & $74$ & $50$ & $50$
        & $92$ & $92$ & $74$ & $74$ \\
    $\mathrm{Prob}_{\mathrm{mosaic}}$
        & 0.75 & 0.75 & 1.0 & 1.0
        & 0.75 & 0.75 & 1.0 & 1.0
        & 0.5 & 0.5 & 0.5 & 0.5 \\
    $\mathrm{Epochs}_{\mathrm{mosaic}}$
        & 36 & 30 & 24 & 24
        & 36 & 36 & 24 & 24
        & 41 & 41 & 44 & 44 \\
    $\mathrm{Prob}_{\mathrm{mixup}}$
        & 0.75 & 0.75 & 1.0 & 1.0
        & 0.75 & 0.75 & 1.0 & 1.0
        & 0.25 & 0.25 & 0.25 & 0.25 \\
    $\mathrm{Epochs}_{\mathrm{mixup}}$
        & 36 & 30 & 24 & 24
        & 36 & 36 & 24 & 24
        & 86 & 86 & 68 & 68 \\
    $\mathrm{Prob}_{\mathrm{copypaste}}$
        & -- & -- & -- & --
        & -- & -- & -- & --
        & 0.5 & 0.5 & 0.5 & 0.5 \\
    $\mathrm{Epochs}_{\mathrm{copypaste}}$
        & -- & -- & -- & --
        & -- & -- & -- & --
        & 85 & 85 & 67 & 67 \\
    Total batch size
        & 32 & 32 & 32 & 32
        & 32 & 32 & 32 & 32
        & 32 & 32 & 32 & 32 \\
    \midrule
    $\lambda_{\mathrm{cls}}$
        & 1 & 1 & 1 & 1
        & 2 & 2 & 2 & 2
        & 2 & 2 & 2 & 2 \\
    $\lambda_{\ell_1}$
        & 5 & 5 & 5 & 5
        & 1 & 1 & 1 & 1
        & -- & -- & -- & -- \\
    $\lambda_{\mathrm{giou}}$
        & 2 & 2 & 2 & 2
        & 1 & 1 & 1 & 1
        & -- & -- & -- & -- \\
    $\lambda_{\mathrm{ddf}}$
        & 1.5 & 1.5 & 1.5 & 1.5
        & 1.5 & 1.5 & 1.5 & 1.5
        & -- & -- & -- & -- \\
    $\lambda_{\mathrm{fgl}}$
        & 0.15 & 0.15 & 0.15 & 0.15
        & 0.15 & 0.15 & 0.15 & 0.15
        & -- & -- & -- & -- \\
    $\lambda_{\mathrm{kpt}}$
        & -- & -- & -- & --
        & -- & -- & -- & --
        & 10 & 10 & 10 & 10 \\
    $\lambda_{\mathrm{oks}}$
        & -- & -- & -- & --
        & -- & -- & -- & --
        & 4 & 4 & 4 & 4 \\
    \bottomrule
    \end{autotabular}
\end{table*}

\paragraph{Object detection. }
We train the complete LoopDet models end to end on COCO~\texttt{train2017} and evaluate them on COCO~\texttt{val2017}~\cite{lin2014microsoft}. The distilled backbone, hybrid encoder, looped decoder, and prediction heads are jointly optimized using AdamW~\cite{loshchilov2018decoupled} and an exponential moving average (EMA). 
We retain DETR-style bipartite matching~\cite{carion2020detr} and the Decoupled Distillation Focal (DDF) and Fine-Grained Localization (FGL) losses of D-FINE~\cite{peng2025dfine}. The optimizer settings, augmentation schedules, training durations, batch sizes, and loss weights are summarized in Table~\ref{tab:hyperparameters}.

\paragraph{Instance segmentation. }
We initialize the LoopSeg backbone from the corresponding trained LoopDet checkpoint and optimize the complete segmentation model on COCO instance annotations. Bipartite matching includes mask costs, and the objective combines the detection loss terms with binary cross-entropy and Dice losses~\cite{V-Net} on matched masks.

\paragraph{Human pose estimation. }
We initialize the LoopPose backbone from the corresponding trained detector and train the pose model on COCO keypoint annotations. Following DETRPose~\cite{janampa2026detrpose}, person queries are matched to ground-truth instances through bipartite matching. The objective combines classification, $L_1$ regression on visible keypoints, and an object keypoint similarity (OKS) loss~\cite{lin2014microsoft}.

\section{Experimental Setup}
\label{sec:setup}

This section provides additional implementation and training details for some experiments in the main paper.

\subsection{Inference Benchmarking Protocol}
\label{sec:inference_protocol}

Latency and peak memory are measured on an NVIDIA RTX 4090 D GPU with batch size 1 and resolution $640\times640$ under FP16 using PyTorch (v2.11.0+cu128). 
Peak memory reports PyTorch peak allocated memory.
YOLO latency includes NMS.
GFLOPs are computed using PyTorch's \texttt{FlopCounterMode}, including attention matrix multiplications but excluding NMS and DCC keypoint decoding.

\subsection{Implementation of Parameter Compression Methods}
\label{sec:compression_implement}

Fig.~\ref{fig:compression} compares LoopDet-X with three representative parameter compression methods under comparable parameter budgets.
For each compression baseline, we start from the same pretrained ECViT checkpoint and integrate the compressed backbone with the hybrid encoder and decoder.
All configurations follow the same backbone distillation and downstream detection training protocol described in Sec.~\ref{sec:training_config}.

\paragraph{Neuron Merging.}
We adapt Intra-Fusion~\cite{theus2024metapruning}, which reduces the number of neurons by merging their weights rather than directly removing less important neurons.
We apply this method to the backbone FFNs, reducing their hidden dimension from 2304 to 896 while retaining all six attention heads per layer.
Specifically, we first select the target neurons using $L_1$ importance scores.
We then measure the similarity between neurons using the normalized weights of the two FFN projections and compute an optimal transport map based on their pairwise $L_1$ distances.
This map determines how the original neurons are merged into the selected neurons, using uniform marginals and an exact Earth Mover's Distance solver.
The merging is applied consistently to both FFN projections and the first projection's bias, while the output bias remains unchanged.

\paragraph{Structured Pruning.}
We adapt Isomorphic Pruning~\cite{fang2024isomorphic} to ECViT using Torch-Pruning.
Unlike neuron merging, this approach directly removes less important structures, such as attention heads and FFN hidden neurons.
To compare their importance more reliably, pruning candidates are grouped according to their structural types and ranked within each group using normalized $L_1$ importance scores.
We then remove the selected attention heads and FFN neurons, updating the corresponding projections together to maintain consistent dimensions.
Each retained attention head has 64 dimensions.
The resulting backbone contains 60 attention heads and 13,824 FFN hidden neurons across 12 layers, with an average FFN width of 1152.

\paragraph{Low-rank Decomposition.}
We adapt FLAR-SVD~\cite{FLAR}, which reduces parameters by approximating a large weight matrix with the product of two smaller matrices, $W \approx UV$.
The intermediate dimension, known as the rank, determines the degree of compression.
We apply this approach to the backbone's QKV, attention output, and two FFN projections.
Following FLAR-SVD, we collect activation statistics from ImageNet-1K~\cite{imagenet15russakovsky} training images to guide the decomposition and search for suitable ranks for each layer.
The decomposition uses stabilized whitening to account for the input feature distribution, while the rank search targets a backbone parameter retention ratio of 55\%.
Among the 48 candidate linear layers, 35 are factorized and 13 remain unchanged.
Each factorized layer retains its original bias, without introducing additional activations between the two factors.

\subsection{Implementation of Depth Conditioning} 
\label{sec:conditioning_implement}

We apply each conditioning mechanism to both the backbone and decoder in Tab.~\ref{tab:PEFT}, with the hybrid encoder unchanged. All variants retain the same backbone grouping and the two shared blocks executed over four layers. Shared model weights and conditioning parameters are jointly optimized. The unconditioned baseline shares the original normalization parameters within each shared block.

The compared mechanisms introduce depth conditioning through scaling, attention mixing, low-rank updates, normalization, or prompts. 
Depth scaling~\cite{jacobs2026block} uses a multilayer perceptron attached to each shared Transformer block to map the layer index to channel-wise scales. 
The weight-transformation baseline follows MiniViT~\cite{zhang2022minivit}: it uses depth-specific attention head mixing before and after softmax, together with independent LayerNorm parameters; decoder head mixing is confined to self-attention. 
LoRA~\cite{hu2022lora} adds depth-specific low-rank updates to selected linear transformations, excluding the terminal singleton backbone group. 
LayerNorm tuning~\cite{ln-tuning} assigns independent gain and bias parameters to each layer. 
VPT-Deep~\cite{jia2022visual} inserts independent prompts at each layer and discards their outputs afterward; decoder prompts participate only in self-attention.

\begin{figure}
    \centering
    \includegraphics[width=1\linewidth] {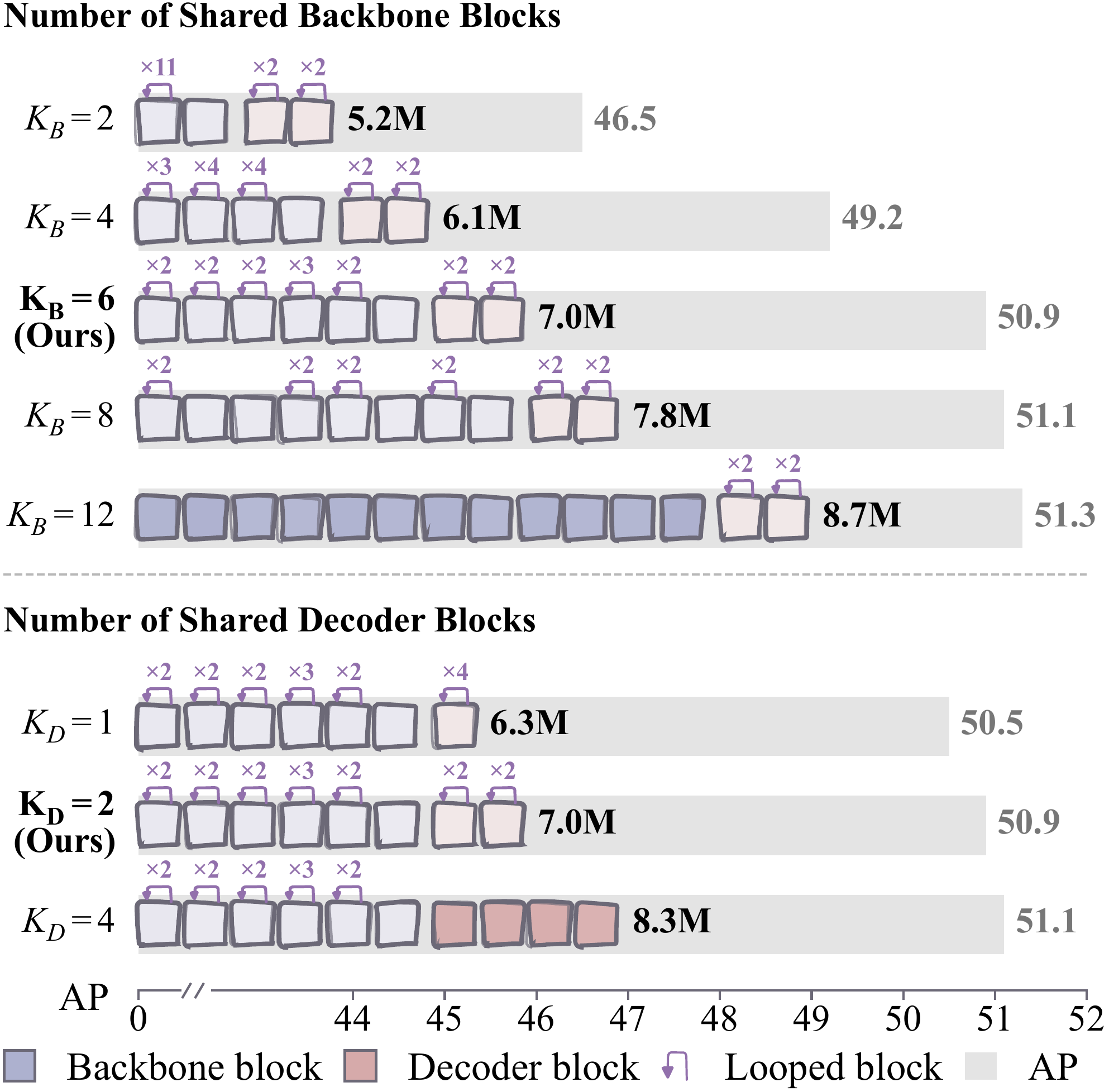}
    \caption{\textbf{Effect of looped blocks count for LoopDet-S.}
    Top: Comparison of different numbers of shared backbone blocks.
    Bottom: Comparison of different numbers of shared decoder blocks.
    All LoopDet-S variants use fixed depths of $L=12$ in the backbone and $T=4$ in the decoder, and follow the same partitioning method and training protocol, using Loop Adapters.
    }
    \label{fig:physical_module_count}
\end{figure}

\section{Additional Experiments}

\subsection{Effect of Looped Blocks Count}
\label{sec:loop_variants}

To investigate how many shared blocks should be retained for a favorable accuracy--parameter trade-off, we vary the number of shared blocks in the backbone and decoder.
We keep their depths fixed at $L=12$ and $T=4$, respectively.
For each configuration, we independently determine the group boundaries using the representation-based grouping strategy described in Sec.~\ref{sec:loop_construction}, while reserving an independent block for the final backbone layer.

In the upper panel of Fig.~\ref{fig:physical_module_count}, detection AP increases with the number of shared backbone blocks, but the gains are smaller beyond $K_B=6$. For example, increasing $K_B$ from $6$ to $8$ adds two shared blocks and $0.8$M parameters for a gain of only $0.2$ AP points. We therefore use $K_B=6$ as a trade-off between accuracy and model size.

In the lower panel of Fig.~\ref{fig:physical_module_count}, detection AP
varies only slightly across the tested decoder configurations.
The decoder with two shared blocks achieves AP comparable to that of the decoder with independent layers while using fewer parameters.

\begin{figure*}[!t]
    \centering
    \includegraphics[width=1\linewidth] {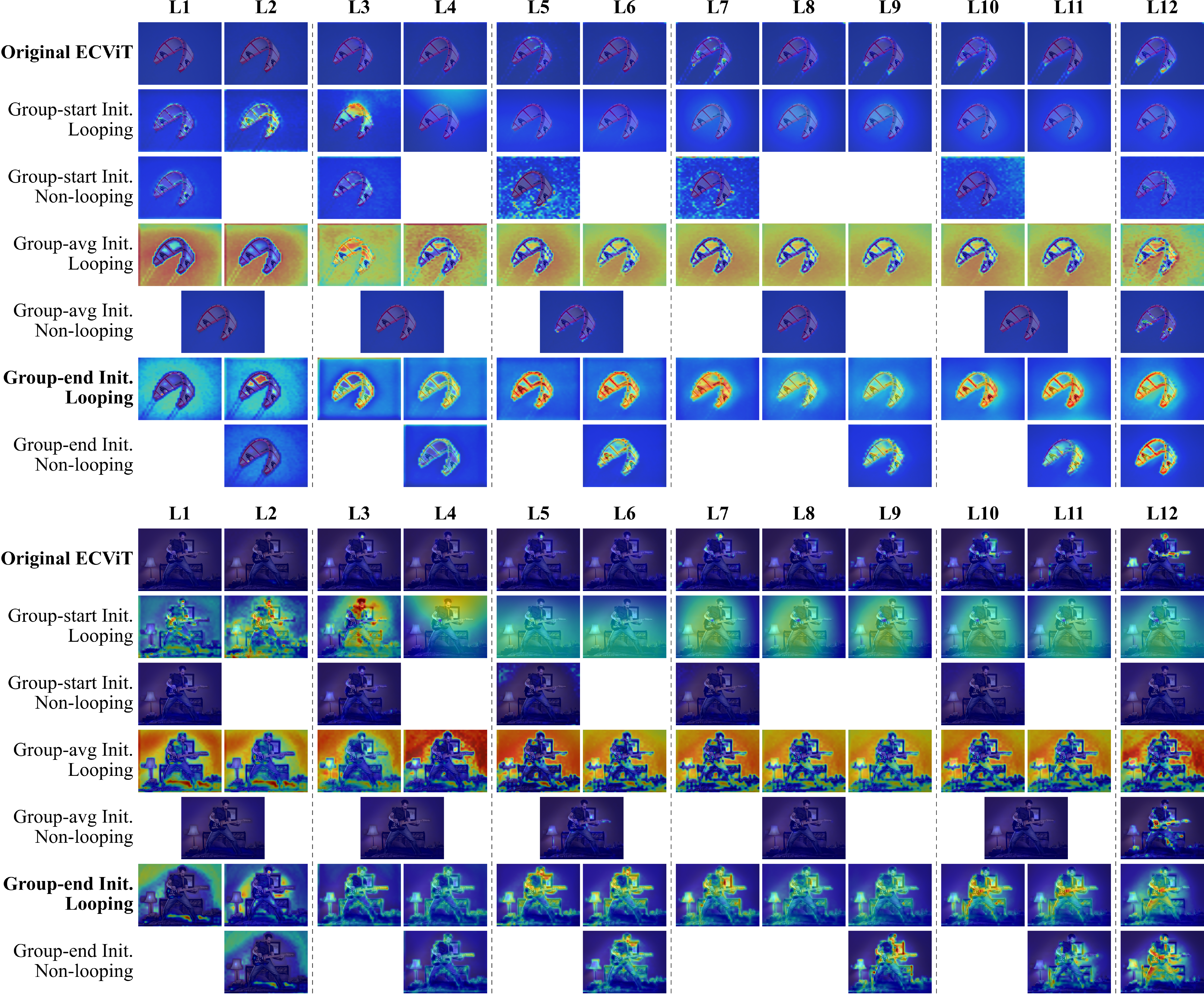}
    \caption{\textbf{Layer-wise attention comparison under different initialization strategies.}
    For each image, the first row shows the distilled ECViT reference, followed by paired looping and non-looping variants using different initialization. The looping variants are visualized immediately after initialization, before distillation. Dashed lines indicate group boundaries. Attention maps are normalized independently for each row, using a shared color scale across all layers within that row.
    }
    \label{fig:supp_init_attention}
\end{figure*}

\subsection{Qualitative Study of Structured Initialization}
\label{sec:supp_init_attention}

Figure~\ref{fig:supp_init_attention} extends the final-layer comparison in Figure~\ref{fig:init_attention} with layer-wise attention visualizations and non-looping controls. 
For each initialization strategy, the looping backbone retains twelve layers, whereas its non-looping counterpart executes one block per group in sequence, yielding six layers. 
All constructed backbones are visualized immediately after initialization, before backbone distillation. 
In the non-looping controls, group-end initialization produces clearer object-related attention structures than group-start initialization, showing that its qualitative advantage is also observable without repeated block execution.

The looping variants reveal how these attention patterns evolve across layers. 
With group-start initialization, early layers exhibit localized responses, while later layers produce spatially smooth maps with little discernible object-related structure, indicating attention collapse.
Group-average initialization instead produces widespread background responses that persist across multiple layers. 
In contrast, group-end initialization maintains clearly identifiable object-related structures through the middle and later layers, including the person and objects’ contour and structure.

These comparisons consistently favor group-end initialization over other strategies under both single and repeated block execution. 
Among the evaluated looping strategies, group-end yields attention patterns closest to those of the original ECViT in the examples shown, suggesting that it provides a better starting point for subsequent backbone distillation by retaining the pretrained model’s object-related attention structure. 
These layer-wise observations reinforce the final-layer analysis in the main paper and are consistent with the initialization ablation in Figure~\ref{fig:init}, where group-end achieves the highest AP.

\subsection{Effect of an Independent Terminal Block.}

To further investigate the effect of reserving an independent terminal block, we extend the uniform grouping ablation in Tab.~\ref{tab:partition}.
Compared with the baseline without an independent terminal block, we keep all preceding sharing groups unchanged and reduce the final block from two executions to one.
As shown in Tab.~\ref{tab:supp_terminal_block}, reserving an independent terminal block achieves 50.6 AP, outperforming the baseline by 1.0 AP despite one fewer layer execution.
Our default configuration further achieves 50.9 AP with 12 layers, improving upon the same baseline by 1.3 AP.
These results highlight the importance of preserving distinct representations between the penultimate and final backbone layers, which facilitates multi-scale feature construction.

\begin{table}[h]
\captionsetup{hypcap=false}
\caption{\textbf{Effect of reserving an independent terminal block.}}
\label{tab:supp_terminal_block}
    \centering
    \small
    \begin{autotabular}{4}{c c c c}
        \toprule
        \makecell{Independent terminal block} & Layers & Group sizes & AP \\
        \hline 
         $\times$ & 12 &
        2--2--2--2--2--2 &  49.6 \\
        $\checkmark$ & 11 &
        2--2--2--2--2--1 &  50.6 \\
        \rowcolor[gray]{0.9}
        $\checkmark$ & 12 &
        2--2--2--3--2--1 & \textbf{50.9} \\
        \bottomrule
    \end{autotabular}
\end{table}

\subsection{Effectiveness of the Looped Architecture.}
\label{sec:supp_detection_from_scratch}

To investigate the advantages of the looped architecture itself, we compare LoopDet-X with ECDet~\cite{liu2026edgecrafter} models trained from scratch. 
All models are randomly initialized and trained on COCO for 50 epochs, without backbone distillation.
We include ECDet-X as the reference model and ECDet-L for comparison under a closer parameter budget.
As shown in Tab.~\ref{tab:supp_detection_from_scratch}, LoopDet-X achieves 39.4 AP with 28.4\% fewer parameters than ECDet-X, trailing by only 0.1 AP.
Compared with ECDet-L, LoopDet-X improves AP by 1.6 with only 2.4M additional parameters.
These results demonstrate that the looped architecture itself provides a favorable accuracy--parameter trade-off, even without pretrained initialization or distillation.

\begin{table}[h]
    \centering
    \small
    \caption{\textbf{Object detection results with training from scratch.} }
    \label{tab:supp_detection_from_scratch}
    \begin{tabular}{lcc}
        \toprule
        Model & \#Params. (M) & AP \\
        \midrule
        ECDet-L~\cite{liu2026edgecrafter} & 32.7 & 37.8 \\
        \color[gray]{0.6} ECDet-X~\cite{liu2026edgecrafter} & \color[gray]{0.6} 49.0 & \color[gray]{0.6} 39.5 \\
        \rowcolor[gray]{0.9}
        \textbf{LoopDet-X (ours)} & 35.1 & 39.4 \\
        \bottomrule
    \end{tabular}
\end{table}

\subsection{Comparison with Existing Looped ViT}

As shown in Tab.~\ref{tab:looped_transformers}, we evaluate the effectiveness of existing looped ViT backbones for dense visual recognition by integrating SReT~\cite{shen2022sliced}, MiniViT~\cite{zhang2022minivit}, and Raptor~\cite{jacobs2026block} into the LoopDet-X detection framework. Each backbone retains its native sharing structure and is initialized from its released pretrained checkpoint. 
Among the evaluated backbones, our looped backbone yields the highest detection AP with the fewest total model parameters.

For a fair comparison, we adapt each pre-trained backbone to the LoopDet-X framework through backbone-specific feature interfaces while preserving its original sharing structure. SReT-S provides outputs from all three stages, while Mini-Swin-S provides outputs from stages 2–4 before subsequent patch merging, yielding feature pyramids at strides 8, 16, and 32. For Raptor ($k=2$), we combine its released recurrent blocks with the patch embedding, special tokens, positional embeddings, and final normalization of DINOv2 ViT-B/14~\cite{oquab2023dinov2}. After 12 layers, the final features are normalized, stripped of special tokens, and bilinearly resized to the three spatial resolutions required by the detector. 
Each feature level is then projected to the hybrid encoder. 
All three backbones share the same LoopDet-X hybrid encoder and looped decoder architecture, including decoder adapters, and are jointly trained with the detector on COCO. Raptor uses $644\times644$ inputs to accommodate its patch size, while the other backbones use $640\times640$ inputs.

\begin{table}[h]
\captionsetup{hypcap=false}
\caption{\textbf{Comparison with looped Transformer backbones on COCO~\texttt{val2017}}.
All backbones use released pretrained weights, and are trained with the same LoopDet-X Hybrid Encoder and decoder.
More implementation details are in the supplementary.
}
\label{tab:looped_transformers}
    \centering
    \small
    \begin{autotabular}{5}{l cc cc}
        \toprule
        \multirow{2}{*}{Backbone}
        & \multicolumn{2}{c}{\#Params. (M)}
        & \multirow{2}{*}{GFLOPs}
        & \multirow{2}{*}{AP} \\
        \cmidrule(lr){2-3}
        & Backbone & Full model & & \\
        \midrule
        SReT-S~\cite{shen2022sliced} & 20.4 & 36.8 & 166.0 & 52.9 \\
        Mini-Swin-S~\cite{zhang2022minivit} & 25.7 & 42.2 & 210.1 & 55.3 \\
        Raptor ($k=2$)~\cite{jacobs2026block} & 20.5 & 36.8 & 710.1 & 53.7 \\
        \rowcolor[gray]{0.9}
        \textbf{LoopDet-X (ours)} & 18.6 & 35.1 & 216.2 & \textbf{57.5} \\
        \bottomrule
    \end{autotabular}
\end{table}

\subsection{Generalization to Other DVR Models}

\begin{table*}[!t]
\captionsetup{hypcap=false}
\caption{\textbf{Object Detection Results on COCO~\texttt{val2017}~\cite{lin2014microsoft}, grouped by parameter budget.}
Results with $^\dagger$ use Objects365 detection pretraining~\cite{shao2019objects365}. 
\textcolor{gray}{ECDet}~\cite{liu2026edgecrafter} is our reference baseline for distillation.
\textcolor{orange}{Orange Results} indicate that our models outperform the corresponding ECDet baselines with much fewer parameters. More details of the evaluation settings are in the supplementary materials.} 
\label{tab:loopdet_full}
	\centering
    \vspace{-1em}
    \small
		\begin{autotabular}{12}{l | ccccc | ccc ccc}
        \toprule
        Model & \#Epochs & \#Params. (M) & Peak mem. (MiB) & GFLOPs &
        Lat. (ms) & AP$^{val}$ & AP$^{val}_{50}$ & AP$^{val}_{75}$ &
        AP$^{val}_S$ & AP$^{val}_M$ & AP$^{val}_L$ \\
        \hline

        YOLOv12-S-turbo~\cite{tian2025yolov12} & 600 & \parambase{9.1} & \parambase{44.9} & 21.4 & 2.53 & 47.6 & 64.5 & 51.5 & 28.3 & 52.7 & 65.9 \\

        {DEIMv2}-S~\cite{huang2025deimv2} & 132 & \parambase{9.7} & \parambase{21.1} & 49.5 & 1.56 & \textbf{50.9} & 68.4 & \textbf{55.1} & 31.3 & 55.3 & \textbf{70.2} \\

        YOLO26-S~\cite{yolo26} & 70 & \parambase{9.5} & \parambase{42.2} & 20.9 & 2.59 & 48.6$^\dagger$ & 65.8$^\dagger$ & 52.8$^\dagger$ & 29.5$^\dagger$ & 53.2$^\dagger$ & 65.8$^\dagger$ \\

        RF-DETR-P~\cite{robinson2025rf} & -- & \parambase{8.5} & \parambase{18.8} & 17.7 & 0.75 & 41.6 & 60.2 & 44.1 & 19.2 & 44.7 & 61.7 \\

        \color[gray]{0.6} ECDet-S~\cite{liu2026edgecrafter} & \color[gray]{0.6} 74 & \color[gray]{0.6} \parambase{9.8} & \color[gray]{0.6} \parambase{21.3} & \color[gray]{0.6} 50.0 & \color[gray]{0.6} 1.51 & \color[gray]{0.6} 51.7 & \color[gray]{0.6} 69.4 & \color[gray]{0.6} 55.8 & \color[gray]{0.6} 32.3 & \color[gray]{0.6} 56.4 & \color[gray]{0.6} 70.5 \\

        \rowcolor[gray]{0.9}
        \textbf{LoopDet-S (ours)} & 74 & \paramdrop{7.0}{29} & \paramdrop{15.9}{25} & 53.0 & 1.62 & \textbf{50.9} & \textbf{68.6} & 54.9 & \textbf{31.4} & \textbf{55.9} & 69.8 \\

        \hline

        D-FINE-S~\cite{peng2025dfine} & 124 & \parambase{10.2} & \parambase{22.3} & 25.5 & 1.02 & 48.5 & 65.6 & 52.6 & 29.1 & 52.2 & 65.4 \\

        LW-DETR-S~\cite{chen2024lw} & 60 & \parambase{14.6} & \parambase{30.7} & 31.6 & 1.34 & 48.0$^\dagger$ & 66.9$^\dagger$ & 51.7$^\dagger$ & 26.8$^\dagger$ & 52.5$^\dagger$ & 65.5$^\dagger$ \\

        YOLOv12-M-turbo~\cite{tian2025yolov12} & 600 & \parambase{19.6} & \parambase{51.8} & 63.1 & 2.96 & 52.5 & 69.9 & 57.1 & \textbf{35.2} & 57.8 & 69.7 \\

        DEIM-S~\cite{huang2025deim} & 132 & \parambase{10.2} & \parambase{22.3} & 25.5 & 1.03 & 49.0 & 65.9 & 53.1 & 30.4 & 52.6 & 65.7 \\
        DEIM-M~\cite{huang2025deim} & 102 & \parambase{19.2} & \parambase{39.9} & 57.0 & 1.56 & 52.7 & 70.0 & 57.3 & 35.3 & 56.7 & 69.5 \\

        DEIMv2-M~\cite{huang2025deimv2} & 102 & \parambase{18.1} & \parambase{37.2} & 84.1 & 2.19 & 53.0 & 70.2 & 57.6 & 34.2 & 57.4 & 71.5 \\

        RT-DETRv4-M~\cite{liao2025rtdetrv4} &  102 & \parambase{19.4} & \parambase{40.3} & 57.0 & 1.53 & 53.5 & 71.1 & 58.1 & 34.9 & 57.7 & 72.1 \\

        \color[gray]{0.6} {ECDet}-M~\cite{liu2026edgecrafter} & \color[gray]{0.6} \color[gray]{0.6} 62 & \color[gray]{0.6} \parambase{19.2} & \color[gray]{0.6} \parambase{39.3} & \color[gray]{0.6} 84.9 & \color[gray]{0.6} 1.92 & \color[gray]{0.6} 54.3 & \color[gray]{0.6} 72.2 & \color[gray]{0.6} 58.7 & \color[gray]{0.6} 35.9 & \color[gray]{0.6} 59.1 & \color[gray]{0.6} 72.7 \\

        \rowcolor[gray]{0.9}
        \textbf{LoopDet-M (ours)}
        & 62 & \paramdrop{13.9}{28} & \paramdrop{29.1}{26} & 90.3 & 2.03 & \textbf{54.0} & \textbf{71.9} & \textbf{58.4} & 34.2 & \textcolor{orange}{\textbf{59.1}} & \textcolor{orange}{\textbf{73.0}} \\

        \hline

        LW-DETR-M~\cite{chen2024lw} & 60 & \parambase{28.2} & \parambase{57.5} & 83.7 & 1.73 & 52.6$^\dagger$ & 69.9$^\dagger$ & 56.7$^\dagger$ & 32.6$^\dagger$ & 57.7$^\dagger$ & 70.7$^\dagger$ \\

        YOLOv12-L-turbo~\cite{tian2025yolov12} & 600 & \parambase{26.5} & \parambase{62.2} & 88.9 & 3.79 & 53.8 & 71.0 & 58.6 & 36.9 & 59.4 & 71.0 \\

        YOLO26-M~\cite{yolo26} & 80 & \parambase{20.4} & \parambase{59.3} & 68.4 & 2.86 & 53.1$^\dagger$ & 70.7$^\dagger$ & 57.7$^\dagger$ & 36.7$^\dagger$ & 57.8$^\dagger$ & 68.9$^\dagger$ \\

        YOLO26-L~\cite{yolo26} & 60 & \parambase{24.8} & \parambase{65.3} & 86.8 & 3.35 & 55.0$^\dagger$ & 72.5$^\dagger$ & 60.0$^\dagger$ & \textbf{38.4}$^\dagger$ & 59.5$^\dagger$ & 71.1$^\dagger$ \\

        \color[gray]{0.6} ECDet-L~\cite{liu2026edgecrafter} & \color[gray]{0.6} \color[gray]{0.6} 50 & \color[gray]{0.6} \parambase{32.7} & \color[gray]{0.6} \parambase{65.7} & \color[gray]{0.6} 147.2 & \color[gray]{0.6} 2.60 & \color[gray]{0.6} 57.0 & \color[gray]{0.6} 75.1 & \color[gray]{0.6} 61.7 & \color[gray]{0.6} 38.7 & \color[gray]{0.6} 62.5 & \color[gray]{0.6} 75.0 \\

        \rowcolor[gray]{0.9}
        \textbf{LoopDet-L (ours)}
        & 50 & \paramdrop{23.4}{28} & \paramdrop{47.3}{28} & 158.9 & 2.82 & \textbf{56.6} & \textbf{74.8} & \textbf{61.4} & 37.4 & \textbf{62.1} & \textcolor{orange}{\textbf{75.1}}\\

		\hline

        D-FINE-L~\cite{peng2025dfine} & 74 & \parambase{30.7} & \parambase{61.5} & 91.6 & 2.28 & 54.0 & 71.6 & 58.4 & 36.5 & 58.0 & 71.9 \\

        DEIM-L~\cite{huang2025deim} & 58 & \parambase{30.8} & \parambase{61.5} & 91.6 & 2.32 & 54.7 & 72.4 & 59.4 & 36.9 & 59.6 & 71.8 \\

        DEIMv2-L~\cite{huang2025deimv2}  & 68  & \parambase{32.2} & \parambase{64.9} & 144.5 & 2.69 & 56.0 & 73.5 & 61.1 & 37.6 & 60.9 & 74.9 \\

        RT-DETRv4-L~\cite{liao2025rtdetrv4} & 58  & \parambase{31.0} & \parambase{61.9} & 91.6 & 2.22 & 55.4 & 73.0 & 60.3 & 37.1 & 60.1 & 72.9 \\

        RF-DETR-S~\cite{robinson2025rf} & -- & \parambase{32.3} & \parambase{64.1} & 98.4 & 1.62 & 54.8$^\dagger$ & 73.9$^\dagger$ & 59.5$^\dagger$ & 35.9$^\dagger$ & 59.8$^\dagger$ & 72.8$^\dagger$ \\
        
        RF-DETR-M~\cite{robinson2025rf} & -- & \parambase{33.8} & \parambase{66.9} & 99.6 & 1.67 & 55.2$^\dagger$ & 74.2$^\dagger$ & 60.1$^\dagger$ & 36.5$^\dagger$ & 60.1$^\dagger$ & 73.6$^\dagger$ \\

        RF-DETR-L~\cite{robinson2025rf} & --  & \parambase{33.8} & \parambase{66.9} & 99.6 & 1.67 & 56.5$^\dagger$ & 75.1$^\dagger$ & 61.3$^\dagger$ & 39.0$^\dagger$ & 61.0$^\dagger$ & 73.9$^\dagger$ \\ %

         \color[gray]{0.6} ECDet-X~\cite{liu2026edgecrafter} & \color[gray]{0.6} \color[gray]{0.6} 50 & \color[gray]{0.6} \parambase{49.0} & \color[gray]{0.6} \parambase{96.4} & \color[gray]{0.6} 204.5 & \color[gray]{0.6} 3.12 & \color[gray]{0.6}  57.9 & \color[gray]{0.6}  76.0 & \color[gray]{0.6}  62.9 & \color[gray]{0.6}  38.7 & \color[gray]{0.6}  63.4 & \color[gray]{0.6}  76.1 \\

        \rowcolor[gray]{0.9}
        \textbf{LoopDet-X (ours)}
        & 50 & \paramdrop{35.1}{28} & \paramdrop{69.8}{28} & 216.2 & 3.33 & \textbf{57.5} & \textbf{75.7} & \textbf{62.4} & \textcolor{orange}{\textbf{40.1}} & \textcolor{orange}{\textbf{63.5}} & \textbf{75.1} \\

		\bottomrule
	\end{autotabular}
\end{table*}

\begin{figure}
    \centering
    \includegraphics[width=1\linewidth] {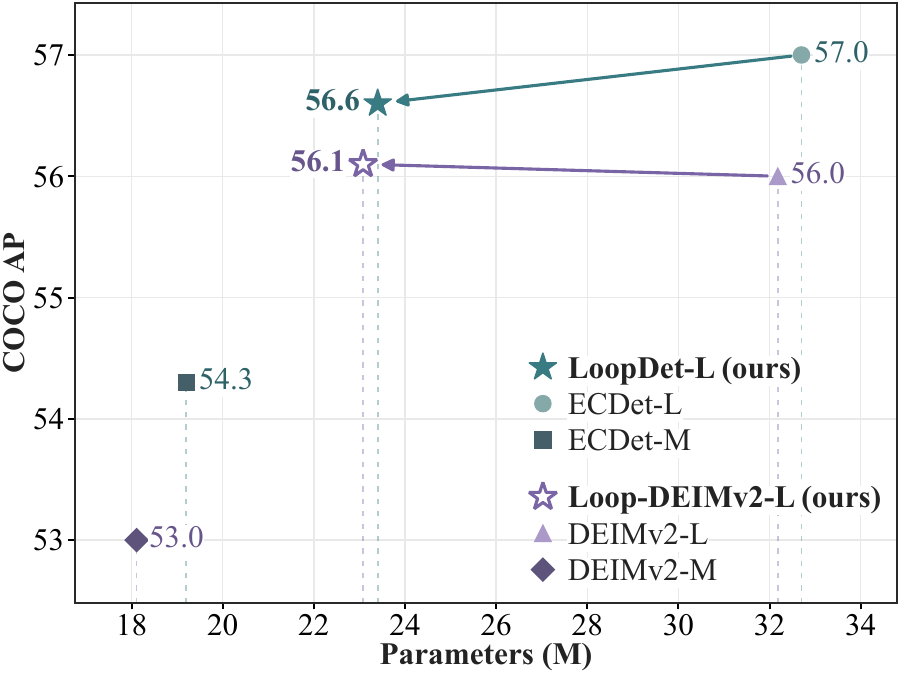}
    \caption{\textbf{Generalization of LoopDVR across pretrained detector families.}
    }
    \label{fig:loop_comparison}
\end{figure}

To further evaluate the generalizability of LoopDVR, we apply the same conversion pipeline to another Transformer-based dense visual recognition architecture.
Specifically, we select DEIMv2~\cite{huang2025deimv2} as an additional evaluation model.
Following the same conversion procedure, we identify sharing groups from pretrained representations and construct a looped backbone and decoder, reducing their independent blocks from 12 to 6 and from 4 to 2, respectively.
The converted model then undergoes backbone distillation and downstream detection training on COCO~\cite{lin2014microsoft}.

As shown in Fig.~\ref{fig:loop_comparison}, Loop-DEIMv2-L achieves 56.1 AP with 23.1M parameters, reducing the parameter count by 28.3\% while matching the original DEIMv2-L's 56.0 AP.
Compared with DEIMv2-M, it achieves 3.1 higher AP with an intermediate parameter budget.
Together with the results on ECDet, these findings demonstrate the generalizability of LoopDVR across different pretrained dense visual recognition architectures.

\begin{table*}[p]
    \centering
    \small
    \captionsetup{skip=5pt}
    \begingroup
\captionsetup{hypcap=false}
\caption{\textbf{Instance Segmentation Results on COCO~\texttt{val2017}, grouped by parameter budget.}
Latency and peak memory are measured at a resolution of $640\times640$, except for RF-DETR~\cite{robinson2025rf} using their native input resolutions.
Results with $^\dagger$ denote models trained with additional data beyond COCO.
\textcolor{gray}{ECInsSeg}~\cite{liu2026edgecrafter} is our reference baseline, although it has more parameters than ours.
\textcolor{orange}{Orange Results} indicate that our models outperform the corresponding ECInsSeg with much fewer parameters.
}
\label{tab:loopseg_full}
	\centering
    \small
		\begin{autotabular}{11}{l | cccc | ccc ccc}
        \toprule
        Model & \#Params. (M) & Peak mem. (MiB) & GFLOPs &
        Lat. (ms) & AP$^{val}$ & AP$^{val}_{50}$ & AP$^{val}_{75}$ &
        AP$^{val}_S$ & AP$^{val}_M$ & AP$^{val}_L$ \\
        \hline

        YOLO-MS-XS-Seg~\cite{yoloms} & \parambase{5.1} & \parambase{46.2} & 26.2 & 39.52 & 36.6 & 56.7 & 39.1 & 15.7 & 40.5 & 56.0\\

        YOLOv12-Seg-S~\cite{tian2025yolov12} & \parambase{9.8} & \parambase{46.4} & 33.4 & 2.88 & 38.6 & 60.8 & 40.7 & 17.0 & 42.1 & 59.2 \\

        \color[gray]{0.6} ECInsSeg-S~\cite{liu2026edgecrafter} & \color[gray]{0.6} \parambase{10.3} & \color[gray]{0.6} \parambase{22.3} & \color[gray]{0.6} 57.0 & \color[gray]{0.6} 1.59 & \color[gray]{0.6} 43.0 & \color[gray]{0.6} 65.7 & \color[gray]{0.6} 46.0 & \color[gray]{0.6} 20.8 & \color[gray]{0.6} 46.3 & \color[gray]{0.6} 65.9 \\

        \rowcolor[gray]{0.9} 
        \textbf{LoopSeg-S (ours)} & \paramdrop{7.4}{28} & \paramdrop{16.7}{25} & 60.0 & 1.72 & \textcolor{orange}{\textbf{43.0}} & \textbf{65.3} & \textcolor{orange}{\textbf{46.2}} & \textbf{20.2} & \textcolor{orange}{\textbf{46.5}} & \textcolor{orange}{\textbf{65.9}}\\

        \hline

        YOLOv11-Seg-S~\cite{yolo11} & \parambase{10.1} & \parambase{51.2} & 33.6 & 2.64 & 37.8 & 59.9 & 40.4 & 19.8 & 42.7 & 55.2 \\

        YOLO26-Seg-S~\cite{yolo26} & \parambase{10.4} & \parambase{46.1} & 35.0 & 3.07 & 40.2$^\dagger$ & 62.0$^\dagger$ & 43.0$^\dagger$ & 18.8$^\dagger$ & 44.0$^\dagger$ & 61.0$^\dagger$ \\

        \color[gray]{0.6} ECInsSeg-M~\cite{liu2026edgecrafter} & \color[gray]{0.6} \parambase{20.1} & \color[gray]{0.6} \parambase{41.0} & \color[gray]{0.6} 96.1  & \color[gray]{0.6} 2.00 & \color[gray]{0.6} 45.2 & \color[gray]{0.6} 68.2 & \color[gray]{0.6} 48.3 & \color[gray]{0.6} 22.9 & \color[gray]{0.6} 49.0 & \color[gray]{0.6} 68.1 \\

        \rowcolor[gray]{0.9} 
        \textbf{LoopSeg-M (ours)} & \paramdrop{14.6}{27} & \paramdrop{30.5}{26} & 101.4 & 2.18 & \textcolor{orange}{\textbf{45.5}} & \textcolor{orange}{\textbf{68.4}} & \textcolor{orange}{\textbf{48.8}} & \textbf{22.4} & \textcolor{orange}{\textbf{49.3}} & \textcolor{orange}{\textbf{68.5}} \\

		\hline

        YOLOv11-Seg-M~\cite{yolo11} & \parambase{22.4} & \parambase{113.8} & 71.8 & 3.11 & 41.5 & 65.1 & 44.5 & 23.0 & 47.2 & 59.1 \\

        YOLOv11-Seg-L~\cite{yolo11} &  \parambase{27.6} & \parambase{79.2} & 132.9 & 4.20 & 42.9 & 67.0 & 46.2 & 24.6 & 48.7 & 60.6 \\

        YOLOv12-Seg-M~\cite{tian2025yolov12} & \parambase{21.9} & \parambase{60.9} & 108.8 & 3.56 & 42.2 & 65.8 & 45.0 & 21.8 & 46.4 & 63.2 \\

        YOLOv12-Seg-L~\cite{tian2025yolov12} & \parambase{28.8} & \parambase{70.5} & 134.7 & 4.42 & 43.1 & 67.3 & 45.8 & 22.3 & 47.9 & 67.9 \\

		YOLO26-Seg-M~\cite{yolo26} & \parambase{23.6} & \parambase{63.8} & 122.2 & 3.74 & 44.2$^\dagger$ & 67.4$^\dagger$ & 47.6$^\dagger$ & 23.0$^\dagger$ & 48.7$^\dagger$ & 64.1$^\dagger$ \\

        \color[gray]{0.6} ECInsSeg-L~\cite{liu2026edgecrafter} & \color[gray]{0.6} \parambase{33.6} & \color[gray]{0.6} \parambase{67.5} & \color[gray]{0.6} 158.3 & \color[gray]{0.6} 2.74 & \color[gray]{0.6} 47.1 & \color[gray]{0.6} 70.9 & \color[gray]{0.6} 50.5 & \color[gray]{0.6} 24.8 & \color[gray]{0.6} 51.1 & \color[gray]{0.6} 69.6 \\

        \rowcolor[gray]{0.9} 
        \textbf{LoopSeg-L (ours)} & \paramdrop{24.1}{28} & \paramdrop{48.7}{28} & 170.0 & 2.98 & \textcolor{orange}{\textbf{48.1}} & \textcolor{orange}{\textbf{71.7}} & \textcolor{orange}{\textbf{51.7}} & \textcolor{orange}{\textbf{24.9}} & \textcolor{orange}{\textbf{52.2}} & \textcolor{orange}{\textbf{71.0}} \\

        \hline

        RF-DETR-Seg-M~\cite{robinson2025rf} & \parambase{35.7} & \parambase{69.9} & 88.2 & 1.77 & 45.3$^\dagger$ & 68.4$^\dagger$ & 48.8$^\dagger$ & 25.5$^\dagger$ & 50.4$^\dagger$ & 65.3$^\dagger$ \\

        RF-DETR-Seg-L~\cite{robinson2025rf} & \parambase{36.2}  & \parambase{70.5} & 125.3 & 2.02 & 47.1$^\dagger$ & 70.5$^\dagger$ & 50.9$^\dagger$ & 28.4$^\dagger$ & 52.1$^\dagger$ & 65.6$^\dagger$ \\  

        RF-DETR-Seg-X~\cite{robinson2025rf} & \parambase{38.1}  & \parambase{75.3} & 218.4 & 3.23 & 48.8$^\dagger$ & 72.2$^\dagger$ & \textbf{53.1}$^\dagger$ & \textbf{30.6}$^\dagger$ & \textbf{53.3}$^\dagger$ & 65.9$^\dagger$ \\ 

        \color[gray]{0.6} ECInsSeg-X~\cite{liu2026edgecrafter}  & \color[gray]{0.6} \parambase{49.9} & \color[gray]{0.6} \parambase{98.2} & \color[gray]{0.6} 215.6 & \color[gray]{0.6} 3.27 & \color[gray]{0.6} 48.4 & \color[gray]{0.6} 72.2 & \color[gray]{0.6} 52.0 & \color[gray]{0.6} 26.3 & \color[gray]{0.6} 52.7 & \color[gray]{0.6} 71.1 \\

        \rowcolor[gray]{0.9} 
        \textbf{LoopSeg-X (ours)} & \paramdrop{35.8}{28} & \paramdrop{71.2}{27} & 227.3 & 3.47 & \textcolor{orange}{\textbf{48.9}} & \textcolor{orange}{\textbf{72.7}} & 51.7 & 26.0 & \textcolor{orange}{53.2} & \textcolor{orange}{\textbf{71.4}} \\

        \bottomrule
	\end{autotabular}
\endgroup

    \par\medskip
    \begingroup
\captionsetup{hypcap=false}
\caption{\textbf{Performance comparison of pose estimation models on COCO~\texttt{val2017}, grouped by parameter budget.}
RTMO~\cite{rtmo} latency includes NMS and DCC.
Results with $^\dagger$ denote models trained with additional data beyond the COCO. 
\textcolor{gray}{ECPose}~\cite{liu2026edgecrafter} is our reference baseline, although it has more parameters than ours.
}
\label{tab:looppose_full}
	\centering
    \small
		\begin{autotabular}{11}{l | cccc | ccc ccc}
        \toprule
        Model & \#Params. (M) & Peak mem. (MiB) & GFLOPs &
        Lat. (ms) & AP$^{val}$ & AP$^{val}_{50}$ & AP$^{val}_{75}$ & AP$^{val}_M$ & AP$^{val}_L$ & AR$^{val}$ \\
        
        \hline

        RTMO-T~\cite{rtmo} & \parambase{6.8} & \parambase{91.4} & 15.8 & 10.68 & 63.4$^\dagger$ & 84.7$^\dagger$ & 68.9$^\dagger$ & 58.1$^\dagger$ & 71.8$^\dagger$ & 67.3$^\dagger$ \\

        RTMO-S~\cite{rtmo} & \parambase{9.9} & \parambase{93.2} & 23.8 & 7.82 & 67.7 & 87.8 & 73.6 & \textbf{61.4} & 77.3 & 71.5 \\

        YOLO11-Pose-S~\cite{yolo11} & \parambase{9.9} & \parambase{30.2} & 23.2 & 3.00  & 58.9 & 86.3 & 64.8 & 54.0 & 68.0 & 66.1 \\
        
        \color[gray]{0.6} ECPose-S~\cite{liu2026edgecrafter} & \color[gray]{0.6} \parambase{9.9} & \color[gray]{0.6} \parambase{30.6} & \color[gray]{0.6} 54.2 & \color[gray]{0.6} 2.65 & \color[gray]{0.6} 68.9  & \color[gray]{0.6} 89.1  & \color[gray]{0.6} 75.2 & \color[gray]{0.6} 60.7  & \color[gray]{0.6} 81.1  & \color[gray]{0.6} 74.6  \\

        \rowcolor[gray]{0.9} 
        \textbf{LoopPose-S (ours)} & \paramdrop{7.7}{23} & \paramdrop{26.3}{14} & 59.6 & 2.98 & \textbf{68.2} & \textbf{88.8} & \textbf{74.0} & 59.8 & \textbf{80.6} & \textbf{73.9} \\

        \hline

        YOLO26-Pose-S~\cite{yolo26}  & \parambase{10.4} & \parambase{31.0} & 24.1 & 2.92 & 62.4 & 85.5 & 68.4 & 55.6 & 73.1 & 68.6 \\

        DETRPose-S~\cite{janampa2026detrpose} & \parambase{11.5} & \parambase{33.8} & 33.4 & 2.33 & 67.0$^\dagger$ & 87.6$^\dagger$ & 72.8$^\dagger$ & 60.2$^\dagger$ & 77.4$^\dagger$ & 73.5$^\dagger$ \\

        \color[gray]{0.6} ECPose-M~\cite{liu2026edgecrafter} & \color[gray]{0.6} \parambase{19.8} & \color[gray]{0.6} \parambase{49.4} & \color[gray]{0.6} 94.7 & \color[gray]{0.6} 3.27 & \color[gray]{0.6} 72.4 & \color[gray]{0.6} 90.9 & \color[gray]{0.6} 78.6 & \color[gray]{0.6} 65.2 & \color[gray]{0.6} 83.6 & \color[gray]{0.6} 78.2 \\

        \rowcolor[gray]{0.9} 
        \textbf{LoopPose-M (ours)} & \paramdrop{14.6}{26} & \paramdrop{39.5}{20} & 100.8 & 3.24 & \textbf{71.8} & \textbf{90.7} & \textbf{78.2} & \textbf{64.1} & \textbf{83.4} & \textbf{77.4} \\

        \hline

        RTMO-M~\cite{rtmo} & \parambase{22.6} & \parambase{76.9} & 61.6 & 6.51 & 71.0 & 89.4 & 77.8 & 65.3 & 79.8 & 74.7 \\

        YOLO11-Pose-M~\cite{yolo11} &  \parambase{20.9} & \parambase{51.1} & 71.5 & 2.97 & 64.9 & 89.4 & 72.4 & 62.2 & 71.6 & 72.2 \\

        YOLO26-Pose-M~\cite{yolo26} & \parambase{21.5} & \parambase{52.3} & 73.3 & 3.49 & 68.3 & 88.8 & 75.1 & 62.9 & 77.4 & 74.3 \\

        YOLO26-Pose-L~\cite{yolo26} & \parambase{25.9} & \parambase{60.7} & 91.7 & 4.24 & 69.6 & 89.7 & 76.2 & 64.6 & 78.4 & 75.7 \\

        DETRPose-M~\cite{janampa2026detrpose} & \parambase{20.8} & \parambase{51.9} & 67.7 & 2.71 & 69.4$^\dagger$ & 89.2$^\dagger$ & 75.4$^\dagger$ & 63.2$^\dagger$ & 79.0$^\dagger$ & 75.5$^\dagger$ \\

        \color[gray]{0.6} ECPose-L~\cite{liu2026edgecrafter} & \color[gray]{0.6} \parambase{34.3} & \color[gray]{0.6} \parambase{77.8} & \color[gray]{0.6} 159.2 & \color[gray]{0.6} 3.78 & \color[gray]{0.6} 73.5  & \color[gray]{0.6} 91.7  & \color[gray]{0.6} 79.9  & \color[gray]{0.6} 66.4  & \color[gray]{0.6} 84.4  & \color[gray]{0.6} 78.8  \\

        \rowcolor[gray]{0.9} 
        \textbf{LoopPose-L (ours)} & \paramdrop{24.6}{28} & \paramdrop{58.7}{25} & 171.7 & 3.98 & \textbf{72.9} & \textbf{91.3} & \textbf{79.1} & \textbf{65.4} & \textbf{84.4} & \textbf{78.3} \\

        \hline

        RTMO-L~\cite{rtmo} & \parambase{44.8} & \parambase{119.3} & 130.8 & 7.62 & 72.4 & 89.9 & 78.8 & 67.1 & 81.0 & 76.2 \\

        DETRPose-L~\cite{janampa2026detrpose} & \parambase{32.8} & \parambase{74.5} & 107.7 & 3.51 & 72.5$^\dagger$ & 90.6$^\dagger$ & 79.0$^\dagger$ & 66.3$^\dagger$ & 82.2$^\dagger$ & 78.7$^\dagger$\\

        \color[gray]{0.6} ECPose-X~\cite{liu2026edgecrafter} & \color[gray]{0.6} \parambase{50.6} & \color[gray]{0.6} \parambase{108.8} & \color[gray]{0.6} 219.8 & \color[gray]{0.6} 4.75 & \color[gray]{0.6} 74.8 & \color[gray]{0.6} 92.2 & \color[gray]{0.6} 81.5 & \color[gray]{0.6} 68.0 & \color[gray]{0.6} 85.4 & \color[gray]{0.6} 80.1 \\

        \rowcolor[gray]{0.9} 
        \textbf{LoopPose-X (ours)} & \paramdrop{36.3}{28} & \paramdrop{82.0}{25} & 232.2 & 4.46 & \textbf{74.3} & \textbf{91.9} & \textbf{80.9} & \textbf{67.5} & \textbf{85.0} & \textbf{79.6} \\

        \bottomrule
	\end{autotabular}
\endgroup

\end{table*}

\section{Complete COCO Results}

The main paper presents representative comparisons grouped by parameter budget. This section extends Tab. 1–3 with additional baseline methods and model scales for object detection, instance segmentation, and human pose estimation on COCO~\texttt{val2017}~\cite{lin2014microsoft}. 
The expanded tables report task-specific accuracy alongside model size and inference efficiency.

\paragraph{Full Object Detection Comparison}

The full detection comparison in Tab.~\ref{tab:loopdet_full} includes a broader range of CNN-based and Transformer-based detectors. Across the S, M, L, and X scales, LoopDet maintains accuracy close to the corresponding ECDet references while reducing parameter counts and peak inference memory.

\paragraph{Full Instance Segmentation Comparison}

The full segmentation comparison in Tab.~\ref{tab:loopseg_full} reports mask AP and its detailed breakdown across additional model families and scales. LoopSeg matches or exceeds the corresponding ECInsSeg references at all four scales with fewer parameters, demonstrating that group-wise parameter sharing also transfers effectively to instance segmentation.

\paragraph{Full Human Pose Estimation Comparison}

The full pose comparison in Tab.~\ref{tab:looppose_full} extends the main-paper results with additional pose estimators and model scales. LoopPose retains competitive keypoint accuracy with substantially fewer parameters than the corresponding ECPose references. Together with the detection and segmentation results, these comparisons demonstrate the applicability of group-wise looping across three dense recognition tasks.

\begin{table*}[p]
\centering
\begingroup

\small
\setlength{\tabcolsep}{4pt}
\renewcommand{\arraystretch}{1.08}
\captionsetup{hypcap=false,skip=5pt}

\caption{\textbf{Parameter-efficiency and complexity of components in backbones.}}
\label{tab:shared_complexity}

\begin{autotabular}{9}{l*{8}{c}}
\toprule
\multirow{2}{*}{Backbone}
& \multicolumn{2}{c}{Transformer Blocks}
& \multicolumn{2}{c}{Loop Adapters}
& \multicolumn{2}{c}{Others}
& \multicolumn{2}{c}{\textbf{Total}} \\

\cmidrule(lr){2-3}
\cmidrule(lr){4-5}
\cmidrule(lr){6-7}
\cmidrule(lr){8-9}

& Params (M) & GFLOPs
& Params (M) & GFLOPs
& Params (M) & GFLOPs
& Params (M) & GFLOPs \\
\midrule

EdgeCrafter-S~\cite{liu2026edgecrafter}
& \parambase{5.3} & 40.6
& -- & --
& 0.3 & 1.8
& \parambase{5.6} & 42.4 \\

\rowcolor{black!5}
\textbf{LoopDVR-S (ours)}
& \paramdrop{2.7}{50} & 40.6
& 0.9 & 2.8
& 0.3 & 1.8
& \paramdrop{\textbf{3.8}}{32} & 45.2 \\

\addlinespace[2pt]

EdgeCrafter-M~\cite{liu2026edgecrafter}
& \parambase{9.5} & 61.7
& -- & --
& 0.5 & 3.1
& \parambase{9.9} & 64.8 \\

\rowcolor{black!5}
\textbf{LoopDVR-M (ours)}
& \paramdrop{4.7}{50} & 61.7
& 1.6 & 5.0
& 0.5 & 3.1
& \paramdrop{\textbf{6.8}}{32} & 69.8 \\

\addlinespace[2pt]

EdgeCrafter-L~\cite{liu2026edgecrafter}
& \parambase{21.3} & 115.2
& -- & --
& 1.2 & 8.3
& \parambase{22.5} & 123.5 \\

\rowcolor{black!5}
\textbf{LoopDVR-L (ours)}
& \paramdrop{10.7}{50} & 115.2
& 3.6 & 11.3
& 1.2 & 8.3
& \paramdrop{\textbf{15.4}}{32} & 134.9 \\

\addlinespace[2pt]

EdgeCrafter-X~\cite{liu2026edgecrafter}
& \parambase{28.4} & 137.9
& -- & --
& 1.2 & 8.3
& \parambase{29.5} & 146.2 \\

\rowcolor{black!5}
\textbf{LoopDVR-X (ours)}
& \paramdrop{14.2}{50} & 137.9
& 3.6 & 11.3
& 1.2 & 8.3
& \paramdrop{\textbf{18.9}}{36} & 157.5 \\

\bottomrule
\end{autotabular}

\par\medskip
\vspace{1em}

\caption{\textbf{Parameter-efficiency and complexity of components in decoders.}}
\label{tab:detseg_decoder_complexity}

\begin{autotabular}{11}{l*{10}{c}}
\toprule
\multirow{2}{*}{Decoder}
& \multicolumn{2}{c}{Transformer Blocks}
& \multicolumn{2}{c}{Prediction Heads}
& \multicolumn{2}{c}{Loop Adapters}
& \multicolumn{2}{c}{Others}
& \multicolumn{2}{c}{\textbf{Total}} \\

\cmidrule(lr){2-3}
\cmidrule(lr){4-5}
\cmidrule(lr){6-7}
\cmidrule(lr){8-9}
\cmidrule(lr){10-11}

& Params (M) & GFLOPs
& Params (M) & GFLOPs
& Params (M) & GFLOPs
& Params (M) & GFLOPs
& Params (M) & GFLOPs \\
\midrule

\multicolumn{11}{l}{\textit{\textbf{Object Detection}}}\\
\addlinespace[1pt]

ECDet-S~\cite{liu2026edgecrafter}
& \parambase{2.2} & 1.6
& \parambase{0.6} & 0.6
& -- & --
& 0.1 & 0.0
& \parambase{2.9} & 2.2 \\

\rowcolor{black!5}
\textbf{LoopDet-S (ours)}
& \paramdrop{1.1}{50} & 1.6
& \paramdrop{0.3}{49} & 0.6
& 0.3 & 0.2
& 0.1 & 0.0
& \paramdrop{\textbf{1.8}}{38} & 2.4 \\

\addlinespace[2pt]

ECDet-M~\cite{liu2026edgecrafter}
& \parambase{4.5} & 3.1
& \parambase{1.0} & 0.9
& -- & --
& 0.2 & 0.1
& \parambase{5.6} & 4.1 \\

\rowcolor{black!5}
\textbf{LoopDet-M (ours)}
& \paramdrop{2.3}{50} & 3.1
& \paramdrop{0.5}{49} & 0.9
& 0.5 & 0.3
& 0.2 & 0.1
& \paramdrop{\textbf{3.4}}{39} & 4.4 \\

\addlinespace[2pt]

ECDet-L~\cite{liu2026edgecrafter}
& \parambase{4.5} & 3.1
& \parambase{1.0} & 0.9
& -- & --
& 0.2 & 0.1
& \parambase{5.6} & 4.1 \\

\rowcolor{black!5}
\textbf{LoopDet-L (ours)}
& \paramdrop{2.3}{50} & 3.1
& \paramdrop{0.5}{49} & 0.9
& 0.5 & 0.3
& 0.2 & 0.1
& \paramdrop{\textbf{3.4}}{39} & 4.4 \\

\addlinespace[2pt]

ECDet-X~\cite{liu2026edgecrafter}
& \parambase{6.6} & 4.3
& \parambase{1.0} & 0.9
& -- & --
& 0.2 & 0.1
& \parambase{7.7} & 5.3 \\

\rowcolor{black!5}
\textbf{LoopDet-X (ours)}
& \paramdrop{3.3}{50} & 4.3
& \paramdrop{0.5}{49} & 0.9
& 0.5 & 0.3
& 0.2 & 0.1
& \paramdrop{\textbf{4.5}}{42} & 5.6 \\

\addlinespace[2pt]
\midrule

\multicolumn{11}{l}{\textit{\textbf{Instance Segmentation}}}\\
\addlinespace[1pt]

ECInsSeg-S~\cite{liu2026edgecrafter}
& \parambase{2.2} & 1.6
& \parambase{1.1} & 7.6
& -- & --
& 0.1 & 0.0
& \parambase{3.4} & 9.2 \\

\rowcolor{black!5}
\textbf{LoopSeg-S (ours)}
& \paramdrop{1.1}{50} & 1.6
& \paramdrop{0.7}{36} & 7.6
& 0.3 & 0.2
& 0.1 & 0.0
& \paramdrop{\textbf{2.2}}{35} & 9.4 \\

\addlinespace[2pt]

ECInsSeg-M~\cite{liu2026edgecrafter}
& \parambase{4.5} & 3.1
& \parambase{1.9} & 12.0
& -- & --
& 0.2 & 0.1
& \parambase{6.6} & 15.2 \\

\rowcolor{black!5}
\textbf{LoopSeg-M (ours)}
& \paramdrop{2.3}{50} & 3.1
& \paramdrop{1.2}{36} & 12.0
& 0.5 & 0.3
& 0.2 & 0.1
& \paramdrop{\textbf{4.2}}{37} & 15.5 \\

\addlinespace[2pt]

ECInsSeg-L~\cite{liu2026edgecrafter}
& \parambase{4.5} & 3.1
& \parambase{1.9} & 12.0
& -- & --
& 0.2 & 0.1
& \parambase{6.6} & 15.2 \\

\rowcolor{black!5}
\textbf{LoopSeg-L (ours)}
& \paramdrop{2.3}{50} & 3.1
& \paramdrop{1.2}{36} & 12.0
& 0.5 & 0.3
& 0.2 & 0.1
& \paramdrop{\textbf{4.2}}{37} & 15.5 \\

\addlinespace[2pt]

ECInsSeg-X~\cite{liu2026edgecrafter}
& \parambase{6.6} & 4.3
& \parambase{1.9} & 12.0
& -- & --
& 0.2 & 0.1
& \parambase{8.7} & 16.4 \\

\rowcolor{black!5}
\textbf{LoopSeg-X (ours)}
& \paramdrop{3.3}{50} & 4.3
& \paramdrop{1.2}{36} & 12.0
& 0.5 & 0.3
& 0.2 & 0.1
& \paramdrop{\textbf{5.2}}{40} & 16.8 \\

\addlinespace[2pt]
\midrule

\multicolumn{11}{l}{\textit{\textbf{Human Pose Estimation}}}\\
\addlinespace[1pt]

ECPose-S~\cite{liu2026edgecrafter}
& \parambase{2.1} & 3.6
& \parambase{0.7} & 0.7
& -- & --
& 0.1 & 1.1
& \parambase{3.0} & 5.4 \\

\rowcolor{black!5}
\textbf{LoopPose-S (ours)}
& \paramdrop{1.4}{33} & 4.8
& \paramdrop{0.7}{12} & 0.9
& 0.3 & 0.6
& 0.1 & 1.2
& \paramdrop{\textbf{2.5}}{16} & 7.6 \\

\addlinespace[2pt]

ECPose-M~\cite{liu2026edgecrafter}
& \parambase{4.5} & 7.5
& \parambase{1.4} & 1.6
& -- & --
& 0.2 & 2.2
& \parambase{6.2} & 11.3 \\

\rowcolor{black!5}
\textbf{LoopPose-M (ours)}
& \paramdrop{2.3}{50} & 7.5
& \paramdrop{1.1}{21} & 1.6
& 0.5 & 1.1
& 0.2 & 2.2
& \paramdrop{\textbf{4.2}}{33} & 12.4 \\

\addlinespace[2pt]

ECPose-L~\cite{liu2026edgecrafter}
& \parambase{5.6} & 9.8
& \parambase{1.4} & 1.6
& -- & --
& 0.2 & 2.2
& \parambase{7.2} & 13.5 \\

\rowcolor{black!5}
\textbf{LoopPose-L (ours)}
& \paramdrop{2.8}{50} & 9.8
& \paramdrop{1.1}{21} & 1.6
& 0.5 & 1.1
& 0.2 & 2.2
& \paramdrop{\textbf{4.7}}{35} & 14.6 \\

\addlinespace[2pt]

ECPose-X~\cite{liu2026edgecrafter}
& \parambase{7.7} & 14.3
& \parambase{1.4} & 1.6
& -- & --
& 0.2 & 2.2
& \parambase{9.3} & 18.0 \\

\rowcolor{black!5}
\textbf{LoopPose-X (ours)}
& \paramdrop{3.8}{50} & 14.3
& \paramdrop{1.1}{21} & 1.6
& 0.5 & 1.1
& 0.2 & 2.2
& \paramdrop{\textbf{5.7}}{39} & 19.2 \\

\bottomrule
\end{autotabular}

\endgroup
\end{table*}

\begin{figure*}[!t]
    \centering
    \vspace{-2em}
    \includegraphics[width=0.9\linewidth] {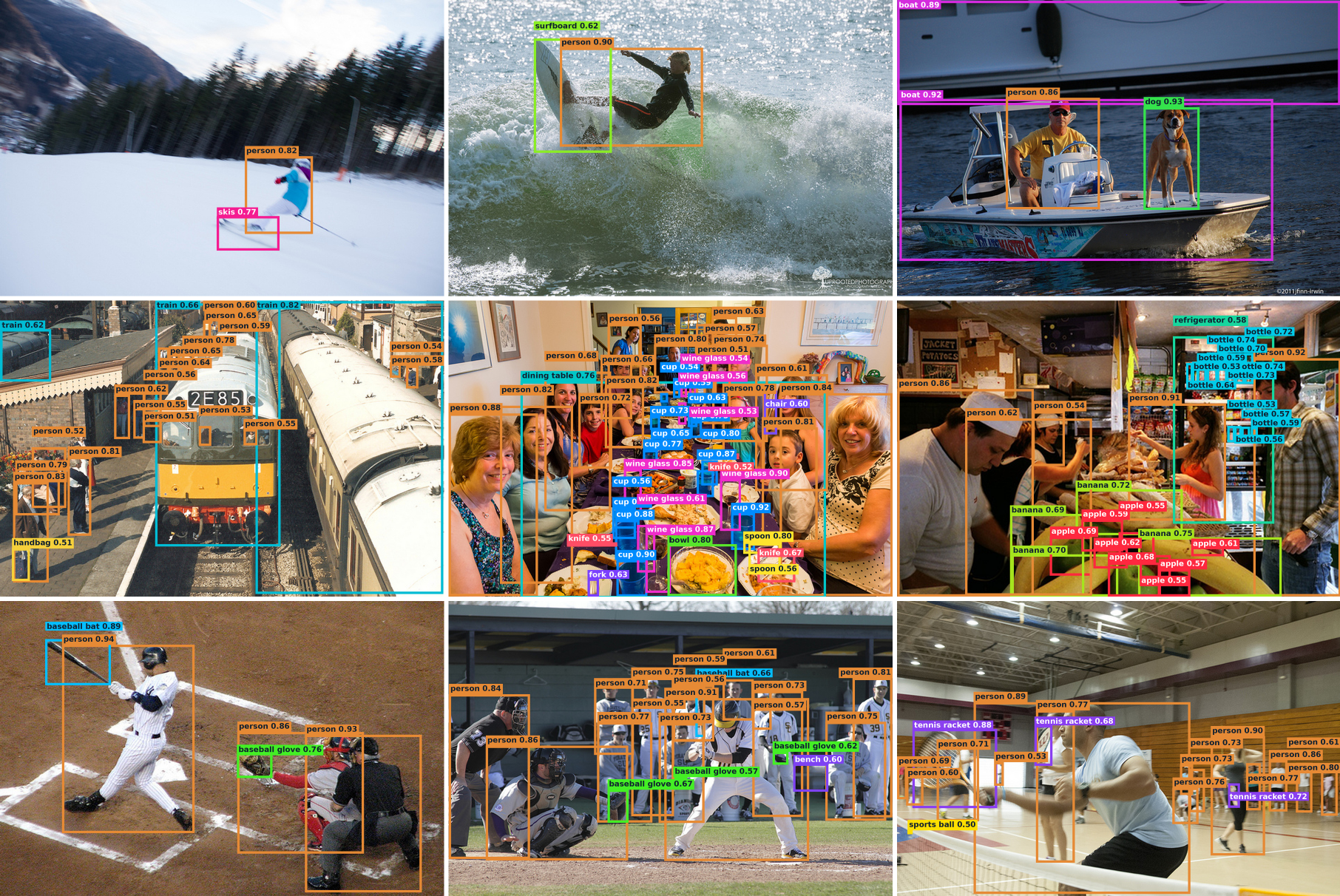}
    \caption{Qualitative object detection results of LoopDet-X (score threshold=0.5).}
    \label{fig:det_vis}
\end{figure*}

\section{Efficiency Analysis}

We compare the parameter and computational costs of the backbone and decoder. We profile models at $640\times640$ resolution with a batch size of one and single-precision arithmetic using PyTorch's \texttt{FlopCounterMode}.

\paragraph{Parameter Efficiency.} 

Group-wise sharing approximately halves the parameters of the backbone Transformer blocks across all four model scales. As shown in Tab.~\ref{tab:shared_complexity}, Loop Adapters introduce additional parameters, while the remaining components retain their original parameter counts.

Sharing also reduces decoder parameters across the three dense recognition tasks, as summarized in Tab.~\ref{tab:detseg_decoder_complexity}. Detection and instance segmentation use the same decoder configuration, with approximately half the Transformer block parameters of their non-looped counterparts. 
Prediction-head parameters are also reduced through sharing.

\paragraph{Computational Cost.} Parameter sharing retains repeated computation at each layer, so the reduction in stored weights does not translate into a proportional reduction in forward operations. The counted costs of the backbone Transformer blocks remain unchanged. Loop Adapters account for most of the additional GFLOPs.

\section{Qualitative Analysis}

We present qualitative results for object detection, instance segmentation, and human pose estimation.
The COCO examples include occlusion, severe truncation, motion blur, defocus, object rotation, and crowded scenes.

\subsection{Object Detection Results}

Figure~\ref{fig:det_vis} illustrates selected LoopDet-X predictions across challenging scenes.
Despite its compact parameter budget, LoopDet-X effectively localizes small, blurry, and partially visible objects, even under substantial visual degradation.

\begin{figure*}[!t]
    \centering
    \vspace{-2em}
    \includegraphics[width=0.9\linewidth]{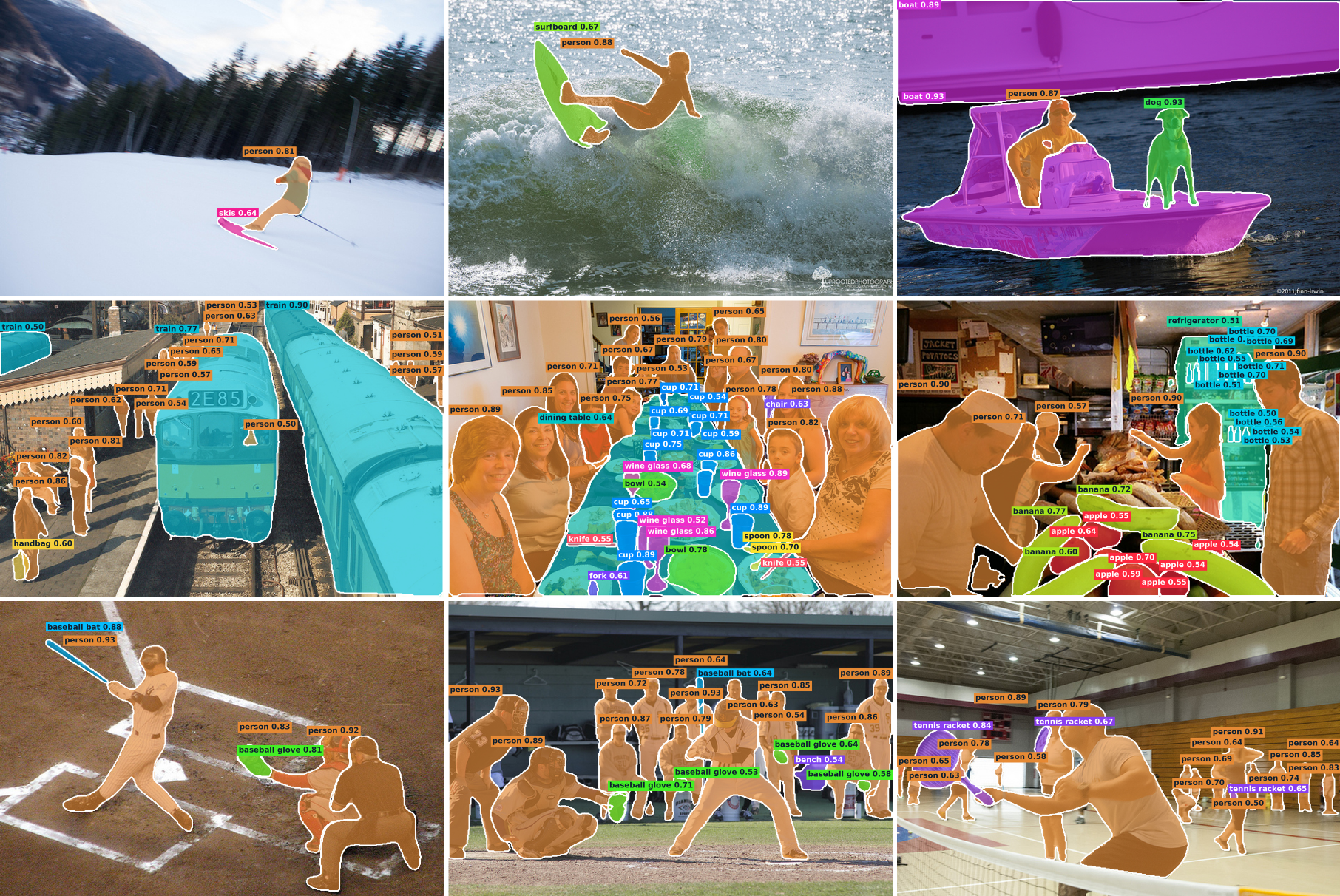}
    \vspace{-0.5em}
    \caption{Qualitative instance segmentation results of LoopSeg-X (score threshold=0.5).}
    \label{fig:seg_vis}
\end{figure*}

\begin{figure*}[!t]
    \centering
    \vspace{-2em}
    \includegraphics[width=0.9\linewidth]{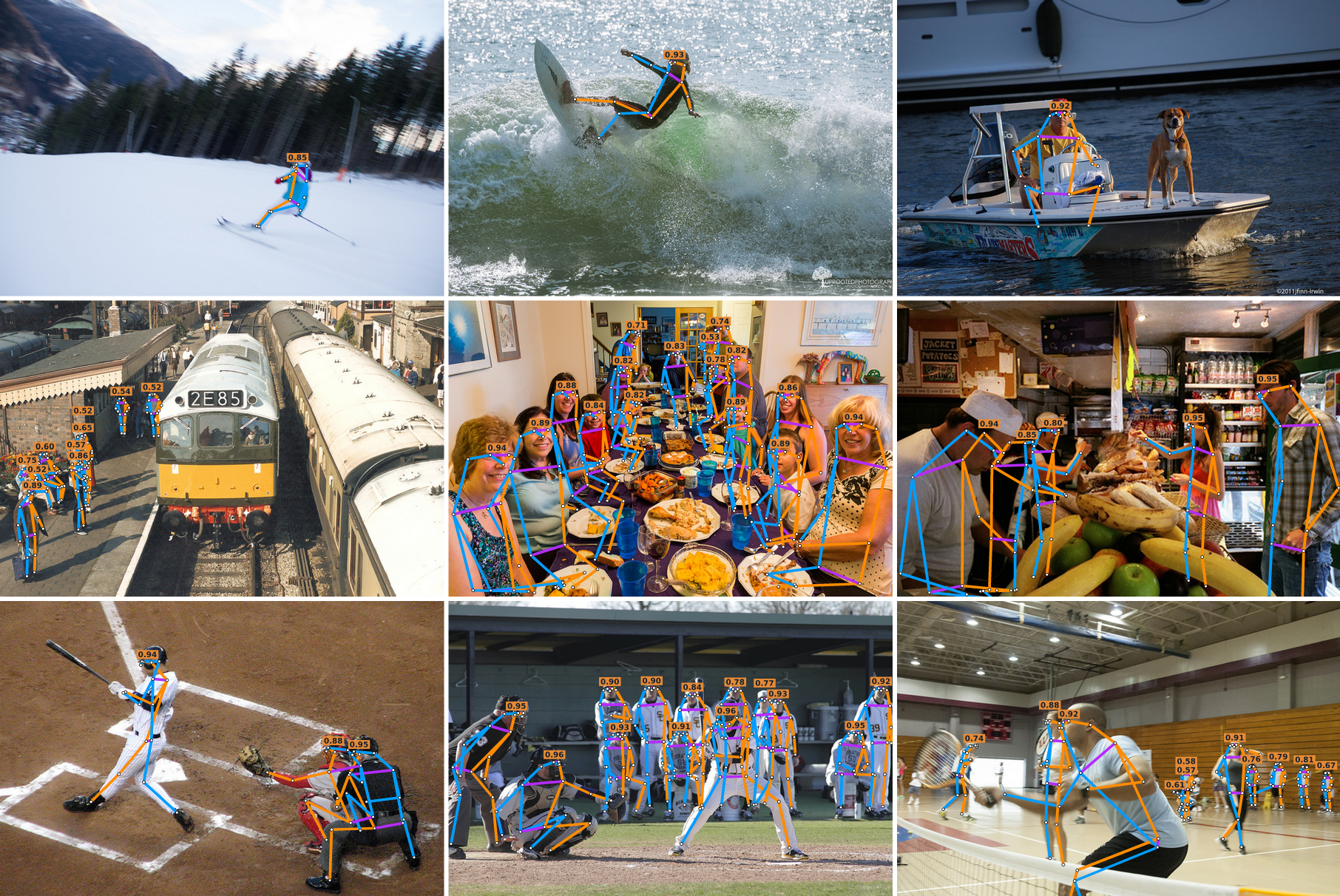}
    \vspace{-0.5em}
    \caption{Qualitative human pose estimation results of LoopPose-X (score threshold=0.5).}
    \label{fig:pose_vis}
\end{figure*}

\subsection{Instance Segmentation Results}

Figure~\ref{fig:seg_vis} presents LoopSeg-X predictions on the same challenging images.
The predicted masks closely follow visible object boundaries and distinguish adjacent or overlapping instances, demonstrating fine-grained spatial recognition.

\subsection{Human Pose Estimation Results}

Figure~\ref{fig:pose_vis} shows LoopPose-X predictions under occlusion, crowding, and motion blur.
The model produces coherent keypoint configurations for multiple individuals, including those with partially visible bodies and complex poses.

\end{document}